\documentclass[10pt,twocolumn,letterpaper]{article}

\usepackage{iccv}

\usepackage{bbding}
\usepackage{times}
\usepackage{epsfig}
\usepackage{graphicx}
\usepackage{amsmath}
\usepackage{amssymb}

\usepackage{appendix}
\usepackage{graphicx}
\usepackage{booktabs}
\usepackage{threeparttable}
\usepackage{multirow}
\usepackage{bm}
\usepackage{nameref}
\usepackage{makecell}
\usepackage{enumerate}
\usepackage{enumitem}
	
\usepackage{color, colortbl}
\usepackage{tabularray}

\usepackage[dvipsnames]{xcolor}
\usepackage{pifont}

\usepackage{cite}
\usepackage{amsmath,amssymb,amsfonts}
\usepackage{algorithmic}
\usepackage{graphicx}
\usepackage{textcomp}

\usepackage{url}
\usepackage{hyperref}
\hypersetup{
	final,
    pagebackref=true,
	breaklinks=true,
	colorlinks,
	linkcolor={Mahogany},
	citecolor={YellowGreen},
	urlcolor={CadetBlue}
}
\usepackage[capitalize]{cleveref}
\Crefname{section}{Section}{Sections}
\Crefname{table}{Table}{Tables}
\Crefname{figure}{Figure}{Figures}
\usepackage[accsupp]{axessibility}  % Improves PDF readability for those with disabilities.

\newcommand{\hq}[2]{\textcolor{black}{#1}}

\definecolor{pink}{rgb}{0.9,0.,0.5}

\iccvfinalcopy 

\def\iccvPaperID{9683} % *** Enter the ICCV Paper ID here

\begin{document}

%%%%%%%%% TITLE
\title{\vspace{-1mm}\Large Multi-view Self-supervised Disentanglement for General Image Denoising\vspace{-3mm}}

\author{Hao Chen$^{*,1}$ \hspace{4mm} Chenyuan Qu$^{*,1}$ \hspace{4mm} Yu Zhang$^{2}$ \hspace{4mm}
 Chen Chen$^{3}$ \hspace{4mm} Jianbo Jiao$^{1}$ \hspace{4mm} \\[2mm]
 $^{1}$University of Birmingham\hspace{4mm} 
  $^{2}$Shanghai Jiao Tong University\hspace{4mm} 
  $^{3}$University of Central Florida \\[2mm]
 Project page: \href{https://chqwer2.github.io/MeD/}{https://chqwer2.github.io/MeD/}
}

\maketitle

% Remove page # from the first page of camera-ready.

%%%%%%%%% ABSTRACT

\newcommand{\G}{\mathit G_\theta^\mathcal {X} }
\newcommand{\D}{\mathit D_\psi^\mathcal {X}   }
\newcommand{\E}{\mathit E_\rho^\mathcal {N}   }
\newcommand{\N}{\mathit F_\phi^\mathcal {N}   }
\newcommand{\R}{\mathit R_\delta^{\mathcal Y} }

\newcommand{\ch}[1]{\textcolor{red}{[#1]}}

% \newcommand{\c}{\iffalse [#1]\fi }

%%%%%%%%% ABSTRACT
\begin{abstract}

With its significant performance improvements, the deep learning paradigm has become a standard tool for modern image denoisers. While promising performance has been shown on seen noise distributions, existing approaches often suffer from generalisation to unseen noise types or general and real noise. It is understandable as the model is designed to learn paired mapping (\eg from a noisy image to its clean version).
In this paper, we instead propose to learn to disentangle the noisy image, under the intuitive assumption that different corrupted versions of the same clean image share a common latent space. A self-supervised learning framework is proposed to achieve the goal, without looking at the latent clean image. By taking two different corrupted versions of the same image as input, the proposed \textbf{M}ulti-view S\textbf{e}lf-supervised \textbf{D}isentanglement (MeD) approach learns to disentangle the latent clean features from the corruptions and recover the clean image consequently.
Extensive experimental analysis on both synthetic and real noise shows the superiority of the proposed method over prior self-supervised approaches, especially on unseen novel noise types. On real noise, the proposed method even outperforms its supervised counterparts by over \textbf{3\ dB}.

\let\thefootnote\relax\footnotetext{$^*$Equal contribution.} 
% \let\thefootnote\relax\footnotetext{\hspace{0.18cm}Communication email: hao.chen.cs@gmail.com} %put yout contact details on the project page.
% \Envelop

\end{abstract}

%-----------------------------------Image F1---------------------------------------%
\begin{figure}[t]
\newcommand\M{\includegraphics[width=0.22 \textwidth]} 

%  左下右上
\newcommand\Ma{\includegraphics[width=0.22 \textwidth, trim=20 270 742 0,clip]}  
 \newcommand\Mb{\includegraphics[width=0.22 \textwidth, trim=260 270 502 0,clip]}  
\newcommand\Mc{\includegraphics[width=0.22 \textwidth, trim=497 270 265 0,clip]}  
\newcommand\Md{\includegraphics[width=0.22 \textwidth, trim=745 270 17 0,clip]}  

	\begin{center}
 \resizebox{1.0\columnwidth}{!}{
		\begin{tabular}{c@{\extracolsep{0.2em}}c@{\extracolsep{0.2em}}c@{\extracolsep{0.2em}}c@{\extracolsep{0.2em}}c@{\extracolsep{0.2em}}c@{\extracolsep{0.2em}}c@{\extracolsep{0.2em}}c}

         \Ma{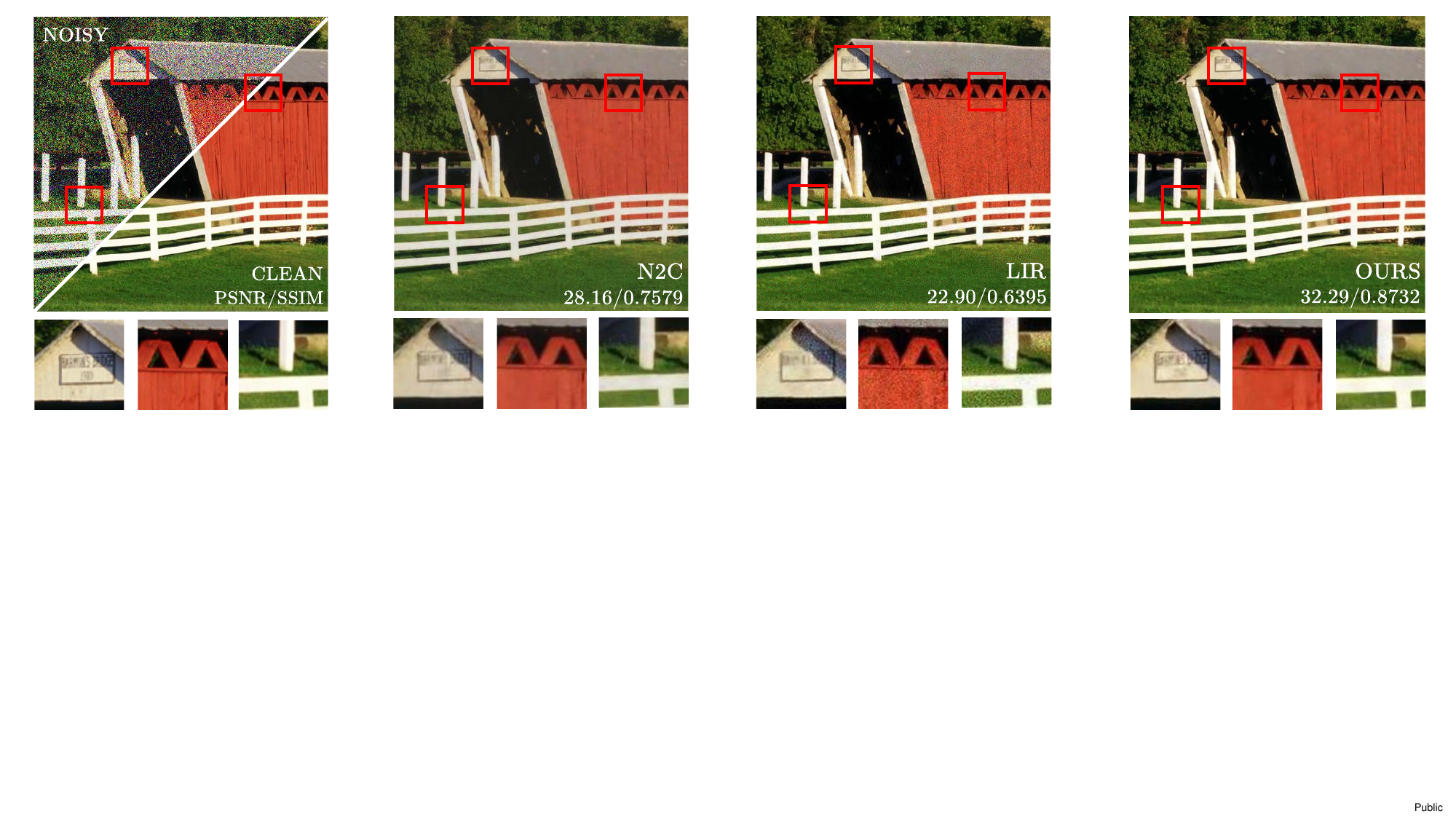}~ \qquad  &\Mb{sections/images/F1.pdf} \\
 		(a) Noisy/ Clean ~ \qquad   & (b) N2C \cite{liu2021Swin}, \textit{Supervised} \\

 		\Mc{sections/images/F1.pdf}~\qquad 		&\Md{sections/images/F1.pdf} \\
 		(c) LIR~\cite{du2020learning}, \textit{Self-supervised}~ \qquad  & (d) Ours, \textit{Self-supervised} \\

 	\end{tabular}
  }

		\vskip 0.25cm
		\caption{ 
		\textbf{Denoising performance on unseen Speckle Noise with $\bm{\hat v}=50$.} The models were trained with Gaussian noise $\sigma\in[5, 50]$. 
		(a) The noisy and clean images, with ground-truth clean patches shown below. 
		(b) \textit{Noise-to-Clean (N2C)}~\cite{liu2021Swin} is trained with clean images. 
		(c) \textit{LIR}~\cite{du2020learning} is self-supervised but needs unpaired clean images as training data. 
		(d) Our approach is fully self-supervised, training with only the noisy input data. 
		}
 \vspace{-6mm}
		\label{fig:overview}
	\end{center}
	% \vskip -0.8cm % 0.4
\end{figure}

%-----------------------------------Image---------------------------------------%

\section{Introduction}
Image restoration is a critical sub-field of computer vision, exploring the reconstruction of image signals from corrupted observations. 
Examples of such ill-posed low-level image restoration problems include image denoising \cite{dncnn, ulyanov2018deep, lehtinen2018noise2noise, neshatavar2022cvf, pang2021recorrupted,linr,zhou2020awgn}, super-resolution~\cite{dong2014learning,lim2017enhanced,agustsson2017ntire,zhang2018image,wang2021unsupervised}, and JPEG artefact removal \cite{dong2015compression, guo2016building, wang2016d3}, to name a few. Usually, a mapping function dedicated to the training data distribution is learned between the corrupted and clean images to address the problem.
While many image restoration systems perform well when evaluated over the same corruption distribution that they have seen, they are often required to be deployed in settings where the environment is unknown and off the training distribution. 
These settings, such as medical imaging, computational lithography, and remote sensing, require image restoration methods that can handle complex and unknown corruptions. Moreover, in many real-world image-denoising tasks, ground truth images are unavailable, introducing additional challenges. 
% This limitation makes supervised methods suboptimal and necessitates the development of blind image restoration methods. 

% \noindent \textbf
% \vspace{-3mm}
\paragraph{Limitations of existing methods:} 
Current low-level corruption removal tasks aim to address the inquiry of ``what is the clean image provided a corrupted observation?"
However, the ill-posed nature of this problem formulation poses a significant challenge in obtaining a unique resolution \cite{bereziat2011solving}.

To mitigate this limitation, researchers often introduce additional information, either explicitly or implicitly. For example, in \cite{NEURIPS2019_2119b8d4}, \textit{Laine et al.} explicitly use the prior knowledge of noise as complementary input, generating a new invertible image model. Alternatively, Learning Invariant Representation (LIR) \cite{du2020learning} implicitly enforces the interpretability in the feature space to guarantee the admissibility of the output. However, these additional forms of information may not always be practical in real-world scenarios or may not result in satisfactory performance.

\paragraph{Main idea and problem formulation:}

Our motivation for tackling this ill-posed nature stems from the solution in the 3D reconstruction of utilising multiple views to provide a unique estimation of the real scene \cite{abdelhamed2018high}. 
Building on this motivation, we propose a training scheme that is explicitly built on multi-corrupted views and perform \textbf{M}ulti-view s\textbf{e}lf-supervised \textbf{D}isentanglement, abbreviated as \textit{MeD}.

Under this new multi-view setting, we 
reformulate the task problem as \textbf{``what is the shared latent information across these views?"} instead of the conventional ``what is the clean image?" By doing so, MeD can effectively leverage the scene coherence of multi-view data and capture underlying common parts without requiring access to the clean image. This makes it more practical and scalable in real-world scenarios.
An example of the proposed method with comparison to prior works is shown in Figure~\ref{fig:overview}, indicating its effectiveness over the state-of-the-art.

Specifically, given any scene image $x^k \sim \mathcal{X}, k\in \mathbb N$ sampled uniformly from a clean image set $\mathcal{X}$, MeD produces two contaminated views:
\begin{equation}
    y_1^k \triangleq \mathcal{T}_1(x^k),\ 
    y_2^k \triangleq \mathcal{T}_2(x^k),
\end{equation}
\noindent forming two independent corrupted image sets $\{\mathcal Y_1\}$, $\{\mathcal Y_2\}$, where $y_1^k\in \mathcal Y_1,\,y_2^k\in \mathcal Y_2$. 
The $\mathcal{T}_1$ and $\mathcal{T}_2$ represent two random independent image degradation operations. 

We parameterise our scene feature encoder $\G$ and decoder $\D$ with $\theta$ and $\psi$. Considering the image pair $\{y_1^k,\, y_2^k\}_{k\in \mathbb N}$, the core of the presented method can be summarised as:
\begin{gather}
	 \G(y_1^k)  \triangleq z_x^{k, i}  \triangleq \G(y_2^k),\\
	\hat x^k\triangleq \D(z_x^{k, i}), 
	\vspace{-0.3pt}
\end{gather}
\noindent where $z_x^{k, i}$ represents the shared scene latent between $y_1^k$ and $y_2^k$ with $i$ referring to the input image index of $y_i$. A clean image estimator $\D$ forms an all-deterministic reverse mapping from $z_x^{k, i}$ to reconstruct an estimated clean image $\hat x^k$. 
Similarly, the noise latent $u_\eta^{k, i}$ is factorised from a corrupted view with a corruption encoder $\E$. Afterwords, the resulting corruption is reconstructed from $u_\eta^{k, i}$ through the use of a corruption decoder, represented by $\N$. 

% factorized by $\rho$  
The \textit{disentanglement} is then performed between $\{z_x^{k, i},\, u_\eta^{k, j}\}_{i \neq j}$ on a cross compose decoder $R_\delta^{\mathcal Y}$ with parameter $\delta$, which can be formulated as:
\begin{gather}
	\hat y_1^k \triangleq R_\delta^{\mathcal Y} (z_{x}^{k, 2}, u_{\eta}^{k, 1}).
	\vspace{-0.3pt}
 \label{eq.4}
\end{gather}

It should be noted that Equation (\ref{eq.4}) is performed over latent features $u$ and $z$ from different views. When assuming that $z_x^{k}$ remains constant across views, the reconstructed view $\hat y_1^k$ is determined by the $u_{\eta}^{k, 1}$. 

\paragraph{Contributions.} 
The contributions of our work are summarised as follows: 
\begin{itemize} %[noitemsep,leftmargin=*] 

\item We propose a new problem formulation to address the ill-posed problem of image denoising using only noisy examples, in a different paradigm than prior works. 
\item We introduce a disentangled representation learning framework that leverages multiple corrupted views to learn the shared scene latent, by exploiting the coherence across views of the same scene and separating noise and scene in the latent space. 
%Our work
\item Extensive experimental analysis validates the effectiveness of the proposed \textit{MeD}, outperforming existing methods with more robust performance to unknown noise distributions, even better than its supervised counterparts.
\end{itemize}
\begin{figure*}%[!t]
  \centering
  % 左下右上
  \includegraphics[width=1.0\linewidth]{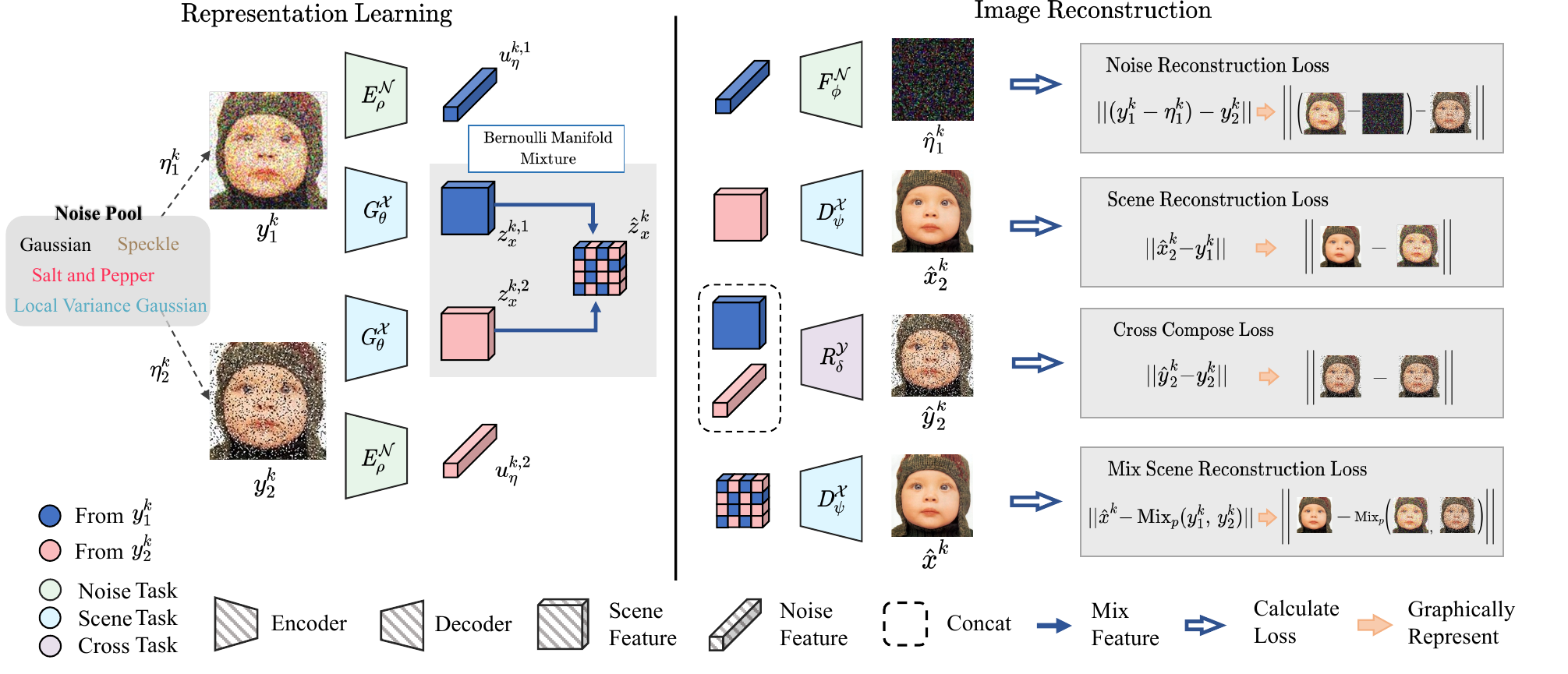} 
  % trim=5 110 20 10,clip]{img/arc.pdf} 
  %{img/architecture.pdf} 
  % \caption{\textbf{Method Overview. }The left part presents the architecture of \textbf{M}ulti-view s\textbf{E}lf-supervised \textbf{DI}sent\textbf{A}nglement (MEDIA). The colored \textit{cubes} and \textit{cuboids} represents scene and distortion features, and the color indicates the image source. The \textit{MEDIA} then permutes features and reconstructs images accordingly. The right part illustrates the self-supervised loss of each output: $\hat \eta_1^k$, $\hat x_2^k$, $y_2^k$ and $\hat x^k$, more details can ref to Sec. \ref{sec:meth}. Noted that, we didn't include all the possible forward paths within this figure, e.g. the reconstruction of ($\hat \eta_2$, $\hat x_1$ and $y_1$). Since they are differ from ($\hat \eta_1$, $\hat x_2$ and $y_2$) only by feature source index. 
  % }

   \caption{\textbf{Method Overview. }
   This figure illustrates the main steps of our proposed method, MeD, which first generates \textit{scene features} (cubes) and \textit{distortion features} (cuboids). The colour of them indicates their image source. In the right section, the features are rearranged and utilised for the four forward paths, from top to bottom, which are the reconstructions of noise ($\hat \eta_1^k$), scene ($\hat x_2^k$), input image ($y_2^k$) and shared scene ($\hat x^k$). It is noteworthy that $\hat y_2^k$ is reconstructed using $z^{k,1}_x$ from $y_1^k$ and $u^{k,2}_\eta$ from $y_2^k$ for feature disentanglement.  Additionally, the reconstruction of $\hat x^k$ relies on mixed scene features to facilitate learning of invariant scene latent. Moreover, the reconstruction paths for ($\hat \eta_2$, $\hat x_1$, and $y_1$) are not depicted here, as they differ from the given paths only in their sub-indexes.
    % all the possible forward paths within this figure, e.g. the reconstruction of . Since they are differ from ($\hat \eta_1$, $\hat x_2$ and $y_2$) only
  }
  
  \label{fig:arch}
\end{figure*}

% Since they are mirror images of the features of
%-------------------------- Architecture ------------------

\section{Related Work}
% Recently, deep networks have shown their superiority in performing image restoration which emerged as a preferable choice compared to conventional restoration approaches. 
% \noindent\textbf{\textbf{Single-view image restoration:}}
\paragraph{Single-view image restoration:}
In~\cite{dong2014learning}, \textit{Dong et al.} were the first to employ a deep network in super-resolution. 
Later, a range of single view-based models 
expanded the idea of supervised deep learning to handle image restoration tasks, such as deblurring \cite{kupyn2019deblurgan},  
 JPEG artefacts \cite{guo2016building}, inpainting \cite{li2020recurrent, yu2019free} and denoising \cite{dncnn, lehtinen2018noise2noise, pang2021recorrupted}.
Recently, it is receiving increasing interest in relaxing the prerequisite of supervised learning with corrupted/ clean image pairs. In the context of image denoising, the ``\textit{corrupted/clean}" pair denotes a corrupted input image and its corresponding clean image for calculating the loss. To tackle the issue of the lack of clean data, several methods have been proposed, such as the Noise2Noise (N2N) method \cite{lehtinen2018noise2noise} and Recorrupted-to-Recorrupted (R2R) \cite{pang2021recorrupted}, which train deep networks on pairs of noisy images. 
Noise2Void (N2V) \cite{krull2019noise2void}, Noise2Self (N2S) \cite{batson2019noise2self}, and the method proposed by \textit{Laine et al.} \cite{NEURIPS2019_2119b8d4} are based on the blind-spot strategy that discards some pixels in the input and predicts them using the remaining. 
In the field of single-image denoising, some methods such as DIP \cite{ulyanov2018deep} and S2S \cite{quan2020self2self}, have achieved remarkable denoising results using only one noisy image.

These methods, however, often inevitably compromise image quality to noise reduction, resulting in over-smoothed output. This trade-off is further exacerbated under domain shifts when dealing with unknown noise distributions.

% \noindent\textbf{\textbf{Restoration based on multi-view:}}
\paragraph{Restoration based on multiple views:}
Existing multi-view variants of image restoration methods mainly focus on sequential data such as video or burst images. 
For example, \textit{Tico} \cite{tico2008multi} builds a paradigm that separates the unique and common features within the multi-frame to produce a denoised estimate. \textit{Liu et al.} \cite{liu2020learning} models degrading elements as foreground and estimate background using video data. 
Deep Burst Denoising (DBD) \cite{godard2018deep} performs multi-view denoising based on burst images.
Each image is taken in a short exposure time and serves as a corrupted observation of the clean image.

Unlike the above-mentioned methods, our \textit{MeD} aims to use multiple static observations simultaneously to learn the latent representation of a clean scene that is shared by multiple discrete views.

% an self-supervised architecture aim to 

% In addition to these sequential data based methods, 
% some frameworks have been proposed for image restoration based on discrete settings. 
% \textit{Zhang et al.} \cite{zhang2009multiple} performs non-local patch grouping across all the noisy images taken from different viewpoints for extracting shared content feature. 

% \subsection{ Image Restoration on Hybrid Distortion}
% Recently, with the increasing demand on image restoration, the task-specific image  restoration on single distortion can no longer meet the needs of real-life applications.

% Then, some works propose model that can handle different types of distortion. RL-Restore \cite{yu2018crafting} provide a pre-trained toolbox and select an appropriate method to gradually restore hybrid-distorted image. The OWAN \cite{suganuma2019attention} selectively performs proper operations in terms of the specific conditions of input image. IPT\cite{chen2021pre} develop pre-trained transformers with multi-heads and multi-tails specific to each low-level vision period.

% \noindent\textbf{\textbf{Self-supervised feature disentanglement:}}
\paragraph{Self-supervised feature disentanglement:}
Another notable path of work has attempted to disentangle underlying invariant content from distorted images. For instance, \textit{UID-GAN} \cite{lu2019uid} utilises unpaired clean/blurred images to disentangle content and blurring effects in the feature space and yield improved deblurring performance.  Similarly, \textit{LIR} \cite{du2020learning} used unpaired input to isolate invariant content features through self-supervised inter-domain transformation. 

However, these methods are limited to synthetic noise and do not extend well in real-world scenarios due to their reliance on clean images. In contrast, our method is purely based on multiple noisy views of the same static scene and aims to disentangle the scene from corruption, without the need for clean image supervision.

\section{Methodology}
\label{sec:meth}
Our primary objective is to identify the commonalities among different views in the denoising process. To achieve this, we aim to discover the shared scene $z_x^k$ that is degradation-agnostic over various corrupted views $\{y_i^k\}_{k\in \mathbb N}$ via our proposed training schema, namely \textit{\textbf{M}ulti-view s\textbf{e}lf-supervised \textbf{D}isentanglement} (MeD). A graphic depiction of MeD is shown in Figure \ref{fig:arch}, composed of the representation learning process in the left panel, and four distinct reconstruction pathways in the right panel.

The detailed design of the proposed schema will be introduced in the following subsections. 
Section \ref{sec:mfp} explains the restoration of noise and scene. Section \ref{sec:cross} details the reconstruction of noisy input using a cross-feature combination. 
Section \ref{sec:bfm} elaborates on the reconstruction of the scene using mixed scene latent.

We will start our introduction by outlining three essential properties that a multi-view representation disentanglement technique should exhibit. 

 % \hspace{0.2cm}

% We start our experiments by removing synthetic additive
% white Gaussian noise (AWGN), and gradually drilled down
% to testing unseen noise levels or noise types, and finally, we
% evaluate the performance in real-world
% To achieve this, we employ a training process that involves duplicating the input image and introducing different types and levels of noise to create multiple views. 
%%%% By leveraging the shared scene representation across multiple views, we can effectively restore the clean image. 

% 

\paragraph{Pre-assumed properties:}
% \noindent\textbf{\textbf{Pre-assumed properties:}} 
Suppose the scene latent space and corruption latent space are symbolised by $\mathcal Z_x$ and $\mathcal U_\eta$, respectively. 
% Two constant variables $i, k\in \mathbb N$.
\begin{itemize}
\item[(1)] %\label{pro:1}
\hypertarget{pro:1}{Independence}: For any scene latent $z_x^k \in \mathcal Z_x$, it is expected to be independent of any corruption latent $u_{\eta}^{k, i}, \in \mathcal U_\eta$.
\item[(2)] %\label{pro:2}
\hypertarget{pro:2}{Consistency}: There exists one shared latent code $z_x^k \in \mathcal Z_x$ that is capable of representing the shared clean component of all instances in the set  $\{y^k_i\}$.  
\item[(3)]  %\label{pro:3}
\hypertarget{pro:3}{Composability}: Recovery of the corrupted view $y^k_i$ can be achieved using the feature pairs ${z_x^k, u_{\eta}^{k, i}}$, and the index of the recovered view is determined by the index of the corruption latent, which represents the unique component within that particular view.
\end{itemize}

% and will later explain how we specifically develope each part of MeD to satisfy these properties.

A key step of our method is to realise these pre-requisitions by determining 
how to implement the latent space assumption. As shown in the left panel of Figure \ref{fig:arch}, to infer our latent space assumption, MeD is comprised of two encoders and three decoders: A shared content latent \textit{encoder} $\G$ and its \textit{decoder} $\D$, an auxiliary noise latent \textit{encoder} $\E$ and its \textit{decoder} $\N$, and a cross disentanglement \textit{decoder} $\R$.

% In supervised learning, the samples $(y^k, x^k)$ are drawn from the joint distributions $P_{\mathcal Y, \mathcal X}(y^k, x^k)$, and the latent space learning is supervised with explicit domain boundary signals given input distribution. 
% While in self-supervised learning, the latent space is only learned by given corrupted view $y^k$ drawn from $P_\mathcal Y(y^k)$. 

%  under two random independent image degradation operation $\mathcal{T}_1$ and $\mathcal{T}_2$
\subsection{Main Forward Process}
\label{sec:mfp}
Given two corrupted views of the same image $x^k$, $y_1^k \triangleq \mathcal{T}_1(x^k)$ and $y_2^k \triangleq \mathcal{T}_2(x^k)$, the \textit{encoder} $\G$ mainly perform the scene feature space encoding that can be formulated as:
\begin{equation}
z_{x}^{k, 1} \triangleq \G(y_1^k),\  z_{x}^{k, 2} \triangleq \G(y_2^k),
\end{equation}
where $z_{x}^{k, 1}$ and $z_{x}^{k, 2}$ are the estimation of the scene feature corresponding to the inputs $y_1^k$ and $y_2^k$.

The process of clean image reconstruction is then completed by the $\D$:
\begin{equation}
    \hat x_1^k \triangleq \D(z_{x}^{k, 1}),\ \hat x_2^k \triangleq \mathit D_\psi^\mathcal {X} (z_{x}^{k, 2}).
\end{equation}

Similar to the process of estimating scene features, the estimation of distortion features by $\E$, followed by the reconstruction of noise with $\N$, can be described as follows:
\begin{equation}
\begin{split}
\label{equ:noise}
u_{n}^{k, 1} \triangleq \E(y_1^k),\  u_{n}^{k, 2}\triangleq \E(y_2^k),  \\
\hat \eta_1^k \triangleq \N (u_{n}^{k, 1}),\ 
\hat \eta_2^k \triangleq \N (u_{n}^{k, 2}).
\end{split}
\end{equation}
% We first introduce the concept of variation predictability, defined as follows:
 % Definition 1. - Here we also need to define what easy is in this context: a prediction is called

We adhere to the methodology introduced by N2N \cite{lehtinen2018noise2noise} to use noisy images as supervisory signals. The objective function of the aforementioned process can be simplified to:
\begin{equation}
\begin{split}
\underset{\theta, \psi}{\operatorname{argmin}}\,\mathcal L^{\mathcal X} &\triangleq ||\hat x_1^k - y_2^k||,  \\ 
\underset{\rho, \phi}{\operatorname{argmin}}\,\mathcal L^{\mathcal N} &\triangleq ||(y_1^k - \hat \eta_1^k) - y_2^k||. 
\end{split}
\end{equation}

The objective of $\hat x_2^k$ and $\hat \eta_2^k$ are the same as above, with only a subscript difference. It should be noted that, although our objective functions are similar to that of N2N, our goal is not simply to find and remove noise, but rather to capture the common features shared across multiple views.

\subsection{Cross Disentanglement}
\label{sec:cross}
% We argue that disentanglement can be natually described by the correlated variations between the latent codes

% That is to say the 
% we adopt a straightforward implementation by directly
% maximizing the mutual information between the varied latent dimension d and

For general latent codes $z_x^k$ to sufficiently represent scene information in
the image space, it is natural to assume that these codes exhibit a certain degree of freedom, allowing them to intersect with the noise space. Consequently, there is no guarantee of complete isolation between $z_x^k$ and $u_\eta^k$. 
To  meet the requirements of properties ({\hyperlink{pro:1}{1}}) and ({\hyperlink{pro:3}{3}}), we use an additional decoder $\R$ to reconstruct a corrupted view over a cross-feature combination, \eg $z_{x}^{k,1}$ from $y_1$ and $u_{n}^{k, 2}$ from $x_2$, which can be represented as:
\begin{equation}
\label{equ:cross}
\begin{split}
\hat y_1^k \triangleq \R (z_{x}^{k,2}, u_{n}^{k, 1}),\
\hat y_2^k \triangleq \R (z_{x}^{k,1}, u_{n}^{k, 2}).
\end{split}
\end{equation}

This realisation explicitly requires $z_{x}^{k, i}$ to represent the common part and $u_n^{k,j}$ to represent the unique part within the corrupted views. We then optimise $\{\theta, \rho, \delta\}$ from $\{\G, \E, \R\}$ using the following objective:
\begin{equation}
\begin{split}
\underset{\theta, \rho, \delta}{\operatorname{argmin}}\,\mathcal L^{\mathcal C} &\triangleq ||\hat y_1^k - y_1^k|| + ||\hat y_2^k - y_2^k||.  
\end{split}
\end{equation}

Generally, it is possible for there to be a trivial solution from $u_{n}^{k, i}$ to $y^k_i$ in Equation (\ref{equ:cross}) such as, when $u_n^{k, 1}$ is extracted from $y_1^k$ and used to reconstruct it as well. 
However, Equation (\ref{equ:noise}) explicitly requires  $u_n^{k, 1}$ to rebuild the noise, which prevents the collapse of $u_n^{k, 1}$ in expressing $y_1^k$.

%%%%%%%%%%%%%%%%%%%%%%%%%%%%%%%%%%%%%%%%%%%%%%%%%%%%%%%%%%%%%%%
% \input{sections/images/mbm.tex}
%%%%%%%%%%%%%%%%%%%%%%%%%%%%%%%%%%%%%%%%%%%%%%%%%%%%%%%%%%%%%%%

\subsection{Bernoulli Manifold Mixture}
\label{sec:bfm}
% \item[$\bullet$] The \textit{Bernoulli Manifold Mixture} (BMM) in Section \ref{sec:bfm}  serves as an instrument for extracting consistent and inter-replaceable scene features. 

The aforementioned latent constraint might appear to be restricted at first,  but in fact, it enables us to capture a large number of degrees of freedom in latent space implementation. For instance, it is possible to have multiple scene features that correspond to a single scene. 
However, in such cases, the mapping from input to scene features becomes ambiguous. 
To tackle this issue, we propose the use of the \textit{Bernoulli Manifold Mixture} (BMM) as a means of leveraging the shared structure within the scene latent. 
% by incorporating a mixture of features for reconstruction.

Given the assumption of property ({\hyperlink{pro:2}{2}}), the acquired scene features $z_x^{k,1}$ and $z_x^{k,2}$ are expected to be identical and interchangeable with one another, as they both refer to the same scene feature. 
BMM establishes a new explicit connection between the scene features of multi-views, which can be expressed in the equation as:
\begin{equation}
\label{Eq:zx}
\hat z_x^k  \triangleq \text{Mix}_p(z_x^{k,1},\, z_x^{k,2}),     % z_x^{k,1} \triangleq z_x^{k,2}
\end{equation}
where the $\hat z_x^k$ is an estimation of the true underlying scene feature. Let $b_f$  define a sample instance drawn from a Bernoulli distribution with probability $p\in(0, 1)$, the function $\text{Mix}_p(\cdot)$ described in Equation (\ref{Eq:zx}) denotes: 
\begin{equation}
\text{Mix}_p(\boldsymbol m, \boldsymbol n) \triangleq  b_f \odot \boldsymbol m +  (1-b_f) \odot \boldsymbol n. 
\end{equation}

% 有了这种新的联系，我们可以通过优化z_x的重建效果来间接的增强z1和z2之前的可互换性
By establishing this new connection (Equation \ref{Eq:zx}), we can enhance the interchangeability between $z_x^{k,1}$  and $z_x^{k,2}$ by optimising the reconstruction performance on $\hat z_x^k$.

% We are interested in the solution of the following problem, at least in some specific asymptotic, for any a and b defined on the same domain.
% regimes:

% One common strategy to implement this constraint is to apply a distance loss directly on $z_{x1}$ and $z_{x2}$,  \textit{e.g.}, $ \mathcal L = || z_{x1} - z_{x2} ||^2$. However, this loss can possibly undermine the quality of the features as it only focuses on the numerical value rather than the representativeness of the features themselves.

% To enforce this constraint without degrading the quality of the features, we propose the \textit{Bernoulli Manifold Mixture}

% Corollary
\noindent\textbf{Lemma 1.}\textit{ Assuming $z_x^{k,i} \sim N_{\mathbf{x}}(\boldsymbol{\mu}, \boldsymbol{\Sigma})$, where $N_{\mathbf{x}}$ denotes multivariate Gaussian distributions and then $\boldsymbol{\mu}$ and $\boldsymbol{\Sigma}$ is the mean and the covariance matrix. }

\noindent\textit{For a given function $\G(\cdot)$, assume $\forall k, i, m, n\in \mathbb N$, the following property holds:}
\begin{equation}
\label{eq:mix}
     \mathbb{E} \left[ \G(z_x^{k,i}) \right] \triangleq \mathbb{E} \left[ \G(\text{Mix}_p(z_x^{k,m}, z_x^{k,n})) \right].
\end{equation}

\noindent\textbf{Proof. } Assume $z_x^{k,m}, z_x^{k,n} \in \mathbb{R}^{\operatorname{dim}}$ are i.i.d., we may also factorise $b_f \in \mathbb{R}^{\operatorname{dim}}$. 
\noindent Write
\begin{equation}
\begin{split}
 \hat z_x^k \triangleq \text{Mix}_p(z_x^{k,m}, z_x^{k,n}) %\\
          % & \triangleq  b_f \odot z_x^{k,m} +  (1-b_f) \odot z_x^{k,n}
\end{split}
\end{equation}

\noindent so that we can have 
\begin{equation}
\begin{split}
 \hat z_x & \sim N_{\mathbf{x}}((b_f + 1- b_f)\boldsymbol{\mu}, (b_f^2 + (1- b_f)^2)\boldsymbol{\Sigma}) \\
          & \sim N_{\mathbf{x}}(\boldsymbol{\mu}, \boldsymbol{\Sigma})
\end{split}
\end{equation}
with the fact that Bernoulli sample $b_f^2$ = $b_f$, the mixed feature $\hat z_x$ is in the same representation distribution as $z_x^{k,i}$.

\hspace{0.2cm}

In MeD, we denote $\hat z_x$ = $\G(\hat z_x^k)$, and the objective for implementing Equation (\ref{eq:mix}) can be formulated as follows:
\begin{equation}
\begin{split}
\underset{\theta, \rho, \psi}{\operatorname{argmin}}\,\mathcal L^{\mathcal M} &\triangleq \lambda  ||\hat z_x - \text{Mix}_p(y^k_1,
\,y^k_2)||, %^2_2, 
\end{split}
\end{equation}
where the $\lambda$ is the weight parameter. 
Here, the target is using a mixed version of $y^k_1$ and $y^k_2$. 
The choice of this is driven by the intuition that the hybrid version would better align with the aforementioned blended features.

\section{Experiments}
To evaluate the effectiveness of our proposed method, we assess our method against several representative self-supervised denoising methods, including Noise2Noise (N2N) \cite{lehtinen2018noise2noise}, Noise2Self (N2S) \cite{batson2019noise2self} and Recorrupted2Recorrupted (R2R) \cite{pang2021recorrupted}, and the invariant feature learning method LIR \cite{du2020learning}. 
Moreover, we also evaluate our approach against two supervised baseline methods (Noise2Clean (N2C) \cite{liu2021Swin} and multi-frame method DBD \cite{godard2018deep}) to further validate its effectiveness.
Comparisons to more methods, including methods using only one noisy image, are presented in the supplementary Table \ref{tab:unseenmore}.

\hq{We start our experiments by denoising synthetic additive white Gaussian noise (AWGN) in Section \ref{exp:Gaussian}, and then move on to testing unseen noise levels and noise types in Section \ref{exp:unseen} and Section \ref{exp:np}, respectively. Furthermore, we evaluate the performance in real-world scenarios in Section \ref{exp:real}. In Section \ref{exp:lambda}, we expand our experiments to incorporate more views to study their impact on performance. Finally, in Section \ref{exp:more}, we apply our method to other tasks, \eg image super-resolution and inpainting, to demonstrate its generalisation ability.}

\subsection{Experimental Setups}
Noted that, the nature of feature disentanglement requires no leak from input to output, however, the global residual connection of the original DnCNN \cite{dncnn} cannot satisfy. Thus, we incorporate the \textit{Swin-Transformer} (Swin-T) \cite{liu2021Swin} instead of the traditionally used DnCNN in our experiments. 
Nevertheless, as Swin-T is not originally designed for image restoration, we make some modifications to enforce local dependence across the image.  Specifically, we add one Convolution Layer each before the patch embedding and after the patch unembedding of Swin-T, as inspired by SwinIR \cite{liang2021swinir}. The resulting modified network backbone is denoted as \textit{Swin-Tx}.

To ensure a fair comparison, we use the \textit{Swin-Tx} backbone for all methods in our study, except for DBD. As DBD did not release the code, we follow the instructions presented in the paper and make our best effort to re-implement it. However, we observe that the two-view DBD could not converge efficiently, which is consistent with the findings in the paper. Therefore, we limit our evaluation to the four-view DBD, denoted as DBD$_4$. Furthermore, we replace the U-Net backbone originally used in the LIR method with Swin-Tx to maintain consistency in our evaluation. This results in an average improvement of approximately 1 dB in PSNR for Gaussian denoising.

% Overall, our evaluation was performed on a variety of challenging image denoising tasks, and we compared the performance of our proposed methods against state-of-the-art methods in the field.

% first train \textit{Swin-Tx} under the setting of \textit{N2C} for identifying the hyper-parameters of best performance and then applied these parameters to all the methods

In all experiments, all methods were trained using only DIV2K \cite{Agustsson_2017_CVPR_Workshops} and the same optimisation parameters, except for LIR and $\text{DBD}_4$ which used manually selected parameters obtained through experiments. For more training and evaluation details including the choice of parameters, please refer to the supplementary Section \ref{sec:expdetails}. %Supplementary \ref{app:training}. 
Code is available at: \href{https://github.com/chqwer2/Multi-view-Self-supervised-Disentanglement-Denoising}{https://github.com/chqwer2/Multi-view-Self-supervised-Disentanglement-Denoising}.

\paragraph{Remark:} In tables, the best results are highlighted in \textbf{bold},
while the second best is \underline{underlined}.

% To enhance the feature invariance across the image, we make a slight modification to the original Swin-T architecture by including a Convolution Layer before the patch embedding and after the patch unembedding stages. We use SGD with a weight decay of 0.0001, a momentum of 0.9, and a mini-batch size of 128. For our DnCNN models, we train for 50 epochs with the learning rate decayed exponentially from $1e^{-1}$ to $1e^{-4}$ over the 50 epochs. All other experimental settings follow those used in \cite{dncnn}. To ensure that the improvement in performance is not solely due to the choice of backbone, we also train a model with DnCNN without residual connection, which is provided in Supplementary \ref{app:DnCNNx}.

% To compare the performance of our approach with existing methods, we select representative unsupervised methods, including \textit{BM3D}, \textit{DIP}, \textit{Noise2Noise (N2N)}, \textit{Noise2Self (N2S)}, and \textit{Recorrupted2Recorrupted (R2R)}, as well as supervised methods such as \textit{Swin-Tx} and \textit{DnCNNx}.

% For further comparison, we also conduct experiments on DnCNN without residual connection, denote as \textit{DnCNNx}, which is provided in Supplementary \ref{app:DnCNNx}.

% rained without clean imag

%%%%%%%%%%%%%%%%%%%%%%%%%%%%%%%%%%%%%%%%%%%%%
% Table 1
% \jb{

\begin{table*}
	\caption{Quantitative comparison of different methods on CBSD68 Dataset \cite{martin2001database} for  Synthetic Gaussian noise. The experiments were conducted on fixed and random variance, respectively. The best results are highlighted in \textbf{bold}, while the second best is \underline{underlined}.}
\label{tab:gaussian}
 % Complete table can be found in the supplementary material. }
 % ∗ denotes that results are obtained from the website
% of SIDD Benchmark}
        % \vspace{0.2cm}
	% \small
	\centering
	\resizebox{\linewidth}{!}{
    \begin{tabular}{l|c|cc|ccc|cc}
     \toprule
     
     % First Line  \multirow{}{}   \makecell[c]{ 
     Training  %\multirow{2}{*}{} 
     & Test  & %\multicolumn{2}{c|}{Single-image-based} & 
     \multicolumn{2}{c|}{Noisy/ Clean}& 
     \multicolumn{3}{c|}{Noisy/ Noisy}  &\multicolumn{2}{c}{Invariant Feature} 
      \\
    %  \midrule  N2V \cite{krull2019noise2void} &

     Schema & $\hat \sigma$ &  N2C \cite{liu2021Swin}  & DBD$_4$ \cite{godard2018deep} & N2N \cite{lehtinen2018noise2noise} & N2S \cite{batson2019noise2self} &  R2R \cite{pang2021recorrupted} & LIR \cite{du2020learning} & MeD (ours)  \\  
    
     \midrule
      \multirow{4}{*}{\makecell[c]{Gaussian \\$\sigma =25$}   }  
% sigma|    Swinx   |    N2N     |     N2S      |      R2R    |     LIR     |    MeDIA  |
& 15  & \underline{33.36/ 0.9020} & \textbf{33.57/ 0.9092} & 32.64/ 0.8805  & 32.77/ 0.8780  & 29.74/0.7865 & 31.06/ 0.8632 & 33.11/ 0.8880 \\ % & 33.59/ 0.9 \\  
% sigma|    Swinx   |    N2N     |     N2S      |      R2R    |     LIR     |    MeDIA  |
& 25  & 30.83/ 0.8494 & \textbf{31.31/ 0.8548} & 30.68/ 0.8334 & \underline{30.99}/ 0.8405& 30.45/0.8183 & 30.01/ 0.8024 & 30.57/ \underline{0.8496} \\ % & 31.10/ 0.86 \\ 
% sigma|    Swinx   |    N2N     |     N2S      |      R2R    |     LIR     |    MeDIA  |
& 50 & 24.76/ 0.5519 & \underline{25.12/ 0.5583} & 24.59/ 0.5385 & 22.13/ 0.3928 & 24.02/0.5133 & 21.97/ 0.3578 & \textbf{25.67/ 0.6026}  \\ %& 25.45/ 0.59\\ 
% sigma|    Swinx   |    N2N     |     N2S      |      R2R    |     LIR     |    MeDIA  |
& 75 & 20.75/ 0.3376 & \underline{21.09/ 0.3412}  & 20.60/ 0.3162 & 17.86/ 0.1998 & 19.10/0.2641 & 16.23/ 0.1689 & \textbf{23.09/ 0.4320}  \\ %& 22.3/ 0.44 \\ 

% ---------------------- Color ------------------------------
%      \midrule
%       \multirow{4}{*}{$\sigma = 50$}     
% % sigma|    Swinx   |    N2N     |     N2S      |      R2R    |     LIR     |    MeDIA  |
% & 15  & 00.00/ 0.00 & 00.00/ 0.00 & 00.00/ 0.00 & 00.00/ 0.00 & 00.00/ 0.00 & 00.00/ 0.00 & 00.00/ 0.00\\  
% % sigma|    Swinx   |    N2N     |     N2S      |      R2R    |     LIR     |    MeDIA  |
% & 25  & 00.00/ 0.00 & 00.00/ 0.00 & 00.00/ 0.00 & 00.00/ 0.00 & 00.00/ 0.00 & 00.00/ 0.00 & 00.00/ 0.00\\ 
% % sigma|    Swinx   |    N2N     |     N2S      |      R2R    |     LIR     |    MeDIA  |
% & 50 & 00.00/ 0.00 & 00.00/ 0.00 & 00.00/ 0.00 & 00.00/ 0.00 & 00.00/ 0.00 & 00.00/ 0.00 & 00.00/ 0.00\\ 
% % sigma|    Swinx   |    N2N     |     N2S      |      R2R    |     LIR     |    MeDIA  |
% & 75 & 00.00/ 0.00 & 00.00/ 0.00 & 00.00/ 0.00 & 00.00/ 0.00 & 00.00/ 0.00 & 00.00/ 0.00 & 00.00/ 0.00\\ 

% ---------------------- Color ------------------------------
   \midrule
   \multirow{4}{*}{\makecell[c]{Gaussian \\$\sigma \in [5, 50]$}   }  
      % \multirow{4}{*}{Gaussian $\sigma \in [5, 50]$}     
% sigma|    Swinx   |    N2N     |     N2S      |      R2R    |     LIR     |    MeDIA  |
%.48095039, 0.49720835, 0.41007981, 0.91060088, 0.86115285,
       % 0.36015759, 0.55445752, 0.24973997, 0.12432532, 0.16963502])

& 15  &  \underline{33.47/ 0.9027} & 33.12/ 0.8915 & 33.45/ 0.8945 & 31.28/ 0.8187  & 20.76/ 0.2508 & 30.85/ 0.8431 & \textbf{33.62/ 0.9026}  \\ %& 33.69/ 0.91 \\  
% sigma|    Swinx   |    N2N     |     N2S      |      R2R    |     LIR     |    MeDIA  |
& 25  & \underline{30.87/ 0.8538} & 30.64/ 0.8491 & 30.77/ 0.8423 & 29.65/ 0.7801  & 23.91/ 0.4552 & 28.92/ 0.8069 & \textbf{30.91/ 0.8573} \\ % & 31.28/0.87 \\ 
% sigma|    Swinx   |    N2N     |     N2S      |      R2R    |     LIR     |    MeDIA  |
& 50 & \underline{27.41/ 0.7361} & 27.13/ 0.7290 & 27.15/ 0.7219 & 27.00/ 0.7114  & 26.92/ 0.6911 & 25.13/ 0.6191 & \textbf{27.48/ 0.7530} \\ % & 27.63/0.76\\ 
% sigma|    Swinx   |    N2N     |     N2S      |      R2R    |     LIR     |    MeDIA  |
& 75 & \underline{25.05/ 0.6223} & 24.97/ 0.6205 & 24.80/ 0.5908 & 24.89/ 0.6023 & 23.83/ 0.5132 & 22.37/ 0.4212 & \textbf{25.40/ 0.6645}  \\ %& 25.61/0.68 \\ 

     \bottomrule
    \end{tabular}
    }
% \vspace{-0.1cm}
\end{table*}

%%%%%%%%%%%%%%%%%%%%%%%%%%%%%%%%%%%%%%%%%%%%%

%%%%%%%%%%%%%%%%%%%%%%%
% Table 2

\begin{table*}%[!htbp]
\caption{Quantitative result of generalisation performance experiment on CBSD68 \cite{martin2001database}. All methods use Gaussian $\sigma=25$ for pre-trained methods and then Gaussian $\sigma\in [5, 50]$ for fine-turning. The better result in each method is highlighted in \textit{italics}.} % 
\label{tab:unseen}
	% \small
	\centering
	\resizebox{\linewidth}{!}{
	% \setlength{\tabcolsep}{4pt}
	%\renewcommand{\arraystretch}{0.98}
 % \scalebox{1}{
    \begin{tabular}{l|cc|cc|cc|c}
     \toprule
     
    Fine-tuning Method & \multicolumn{2}{c|}{N2C \cite{liu2021Swin}}& \multicolumn{2}{c|}{N2N \cite{lehtinen2018noise2noise} } & 
    % \multicolumn{2}{c|}{N2S \cite{batson2019noise2self}}& 
    \multicolumn{2}{c|}{LIR \cite{du2020learning}} & MeD \\
    Pretraining Method & N2C & MeD & N2N & MeD & LIR & MeD & MeD \\
      \midrule
      Gaussian, $\hat \sigma \in [15, 75]$ & 29.20/     0.7797 & \underline{\textit{29.53/ 0.8081}} &29.04/   0.7642& \textit{29.21/     0.7890 } &26.42/ 
  0.6640  &\textit{27.25/   0.7036}& \textbf{29.60/   0.8101}\\
  
     % Gaussian, $\hat \sigma =15$ & 33.47/  0.9032 & \underline{\textbf{33.69/  0.9177}} & 33.45/  0.8923  & \underline{33.57/  0.9002} & 
     % % \underline{31.73/  0.86} & 31.09/  0.85 &  
     % 30.85/   0.8471 & \underline{31.27/   0.8685} & 33.69/  0.9066  \\
     
     % Gaussian, $\hat \sigma =25$ &30.87/  0.8512 & \underline{31.02/  0.8625} &  30.77/  0.8491 & \underline{30.93/  0.8655} & 
     % % \underline{30.02/  0.82} & 29.47/  0.81 & 
     % 28.92/   0.8082 & \underline{29.22 /   0.8113} & \textbf{31.28/  0.8772}  \\
     
     % Gaussian, $\hat \sigma =50$ & 27.41/  0.7417 & \underline{27.68/  0.7662} & 27.15/  0.7253  & \underline{27.26/  0.7549} & 
     % % \underline{27.11/  0.72} & 26.73/  0.69 & 
     % 24.53/   0.5957 & \underline{24.98/   0.6454} & \textbf{27.81/  0.7680}  \\
 
     % Gaussian, $\hat \sigma =75$ & 25.05/  0.6226 & \underline{25.72/  0.6860} & 24.80/  0.5902  & \underline{25.08/  0.6355} & 
     % % 24.78/  0.58 & \underline{24.96/  0.59} & 
     % 21.37/   0.4049 & \underline{23.52/   0.4894} & \textbf{25.61/  0.6865}  \\
     
     Local Var Gaussian  &  35.62/  0.9308 & \underline{\textit{35.85/  0.9439}} & 35.66/  0.9256 & \textit{35.73/  0.9310} &  
     % \underline{33.08/  0.89} & 32.21/  0.89  &
     29.26/   0.8170  & \textit{30.51/   0.8387} & \textbf{35.91/  0.9762} \\
     
     Poisson Noise  & 40.49/  0.9736 & \underline{\textit{42.80/  0.9776}} & 41.35/  0.9736  & \textit{42.27/  0.9813} & 
     % 35.26/  0.93 & \underline{36.04/  0.95} & 
     31.23/   0.8672 & \textit{33.47/   0.8932} & \textbf{45.05/  0.9826} \\
     
     Speckle, $ \hat v\in [25, 50]$ & 33.36/  0.9004 & \underline{\textit{33.40/  0.9044}} & 33.32/ \textit{0.8931}   &  \textit{33.33}/  0.8907 &  
     % 32.86/  0.89 & \underline{32.03/  0.88} & 
      28.28/  0.7713  & \textit{29.82/   0.8229}  & \textbf{33.48/  0.9115}  \\
     
     % Speckle, $ v=50$ & 31.50/  0.8781 & \underline{31.58/  0.8867} & 31.39/  0.8661  & \underline{31.39/  0.8712} & 
     % \underline{30.42/  0.84} & 29.88/  0.83 & 
     % 27.42/   0.7432 & \underline{28.63/   0.8025} & \textbf{31.47/  0.8756}  \\
     
     S\&P, $\hat r \in [0.3, 0.5]$ & 28.85/  0.8267 & \underline{\textit{30.73/  0.8372}} & 28.59/  0.8003  & \textit{29.45/  0.8255} & 
     % 27.90/  0.77 & \underline{27.95/  0.77} & 
     26.69/   0.7241 &  \textit{27.62/   0.7460} & \textbf{30.84/  0.9135} \\
     \midrule
     Average & 33.50/   0.8822  & \underline{\textit{34.46/ 0.8942}} &33.59/   0.8714 &\textit{34.00/   0.8835} &28.38/
  0.7687 &\textit{29.73/   0.8009} & \textbf{34.98/   0.9188}\\
     % 32.05/  0.8534  & \underline{32.72/   0.8660} &  32.05/   0.8380 & 
 % \underline{32.33/     0.8530} & 
 % \underline{30.35  0.81} &  30.04/  0.81& 
 % 27.71/  0.7363 &   \underline{28.91/    0.7653} &   \textbf{32.98/    0.8851}  \\
% \midrule
% n2c_finer_swinv2_layer2_sigma[5, 50]_res48_mixup0_sp_pretrain@25_more
% n2c_finer_swinv2_layer2_sigma[5, 50]_res48_mixup0_sp_pretrain@MeDIA_b0.1_25_more

% n2n_finer_swinv2_layer2_sigma[5, 50]_res48_mixup0_sp_pretrain@25_more

% r2r_finer_swinv2_layer2_sigma[5,50]_res48_mixup0_sp_pretrain@25

% media_finer_swinv2_layer2_sigma[5, 50]_res48_mixup1.0_fused0.1_sp_pretrain@25

% gaussian@25, speckle@25 & 30.15/  0.8292  & 30.62/  0.8553 & 29.70/  0.8114 & 30.27/  0.8310 &   27.72/  0.6752 & 28.83/  0.7125  & 30.96/  0.8601 \\

% gaussian@50, speckle@25 &27.26/  0.7329 & 27.89/  0.7645  & 26.99/  0.7130 & 27.31/  0.7382 &  25.72/  0.6803 &  27.02/  0.7189  &  28.39/  0.7840\\
% LVG, poission &  30.58/  0.9087  & 31.78/  0.9114 & 30.96/  0.9029 & 31.60/  0.9073 & 27.63/  0.7894 & 29.65/  0.8495 & 32.75/  0.9290  \\

% poission, speckle@25 & 30.46/  0.9009 & 30.98/  0.9127 & 29.67/  0.8996 & 31.08/  0.8949  & 26.42/  0.7786   & 29.36/  0.8772 & 31.68/  0.9214 
     \bottomrule
    \end{tabular}
    }
\end{table*}

%%%%%%%%%%%%%%%%%%%%%%%

% -------------------------QUANTITATIVE Table ------------------------
\subsection{AWGN Noise Removal}
\label{exp:Gaussian}
We first investigate the denoising generalisation of the methods using synthetic zero-mean additive white Gaussian noise (AWGN). The experiments are divided into two parts. The first segment employs fixed variance AWGN, whereas the second segment employs varied variance Gaussian for training in a separate manner. Table \ref{tab:gaussian} summarises the quantitative results evaluated on CBSD68 Dataset \cite{martin2001database} at variance levels of 15, 25, 50, and 75.

\noindent \textbf{Analysis:} 
 In the fixed variance setting, MeD performs inferior compared to the other methods on lower noise levels of 15 and 25. However, as the methods face more severe corruption, MeD outperforms all self-supervised and supervised methods, showing our greater advantage of handling severe noise. For instance, at $\sigma = 75$, MeD outperforms the second-best method (N2C) by around 2 dB. These results suggest that MeD has a remarkable ability to generalise to a range of unseen noise levels in Gaussian noise. 
 
% In the fixed variance setting, MeD exhibits slightly inferior performance $\sim-0.5$dB compared to N2N and N2S when evaluated on the same noise levels of training or smaller $\sigma$ values, specifically at 15 and 25. However, when it comes to severe corruption, MeD surpasses even the performance of supervised methods by $\sim$1dB and $\sim$3dB with $\sigma$ levels of 50 and 75, while other self-supervised methods still perform worse than N2C and DBD$_4$. The results of training on other fixed variances are shown in the Supplementary \ref{app:sigma}.

% Similar observations can also be viewed over the vary variance training setting results. While all the unsupervised methods are slightly worse than supervised benchmarks, our MeD surpasses them at all four noise levels.

In the context of random variance, it has been observed that MeD exhibits superior performance across all four noise levels compared to other methods, including supervised methods. 
These findings imply that MeD can benefit more from varying training noise than other methods. More experiments and details on AWGN noise removal can be found in the supplementary Table \ref{sup:gauss}. %Supplementary \ref{app:sigma}.

% -------------------------Visualize Image ------------------------
% Figure 3
\begin{figure*}[!t]\footnotesize
% \captionsetup{font=small}
\hspace{-0.20cm}
\newcommand\M{\includegraphics[width=0.106 \textwidth]}

% 
%  左下右上
\newcommand\Mlarge{\includegraphics[width=0.945 \textwidth, trim=0 20 0 0,clip]}  

\begin{tabular}{c@{\extracolsep{0em}}
c@{\extracolsep{0.05em}}
c@{\extracolsep{0.05em}}
c@{\extracolsep{0.05em}}
c@{\extracolsep{0.05em}}
c@{\extracolsep{0.05em}}
c@{\extracolsep{0.05em}}
c@{\extracolsep{0.05em}}c}

        \M{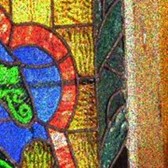}~
		&\M{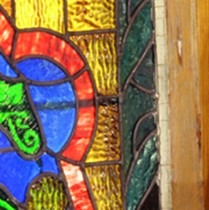}~
        &\M{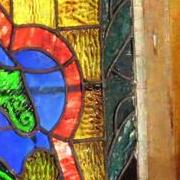}~
        &\M{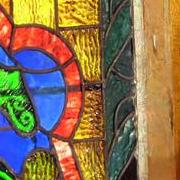}~
		&\M{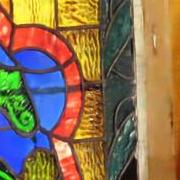}~
		% &\M{img/com/4/7.tif-Self2Self-100000.png}~
		&\M{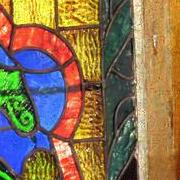}~
		&\M{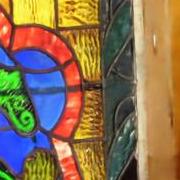}
        &\M{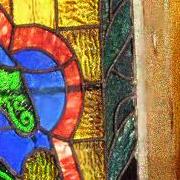}~
		&\M{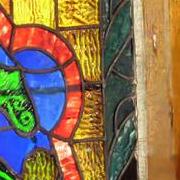}
  
             % \multicolumn{9}{c}{\includegraphics[width=1.0\linewidth, trim=\\ 310 0 140,clip]{sections/images/com.pdf}}
	\\ %~ 9 
		 McM-17 \cite{zhang2011color} & Reference & N2C~\cite{dncnn} & DBD~\cite{godard2018deep} &
		 N2N~\cite{lehtinen2018noise2noise} & N2S~\cite{batson2019noise2self} & R2R~\cite{pang2021recorrupted} & LIR~\cite{du2020learning} & MeD (Ours) \\
		 Speckle $\hat v=50$ &  PSNR/SSIM & 27.58/ 0.7712 & 28.10/ 0.7347 & 27.06/ 0.7452 & 26.99/ 0.7338 & 25.56/ 0.5529 & 23.29/ 0.5681 & \textbf{28.57/ 0.7722}
   \\
   
        \M{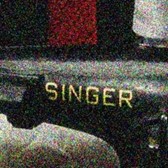}~
		&\M{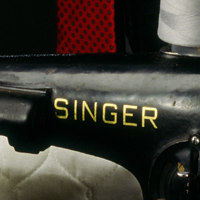}~
        &\M{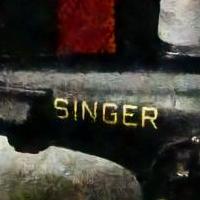}~
        &\M{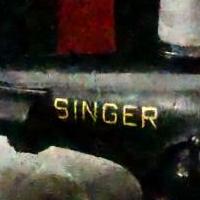}~
		&\M{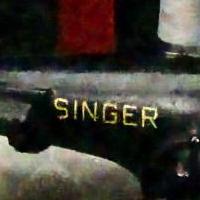}~
		% &\M{img/com1/1/mcm-1-75-Self2Self-100000.png}~
		&\M{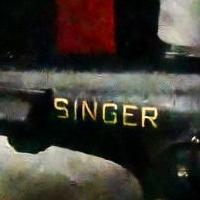}~&\M{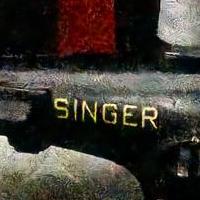}~
		&\M{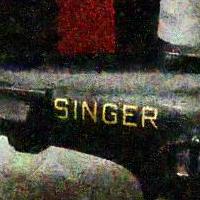}~
		
		&\M{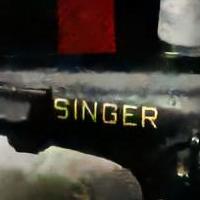}
            % \multicolumn{9}{c}{\includegraphics[width=1.0\linewidth, trim=3 430 0 10,clip]{sections/images/com.pdf}}
		\\ %~ 9 
		 McM-01~\cite{zhang2011color} &Reference  & N2C~\cite{dncnn} & DBD~\cite{godard2018deep} &
		 N2N~\cite{lehtinen2018noise2noise} &  N2S~\cite{batson2019noise2self} & R2R~\cite{pang2021recorrupted} & LIR~\cite{du2020learning} & MeD (Ours) \\
		Gaussian $\hat \sigma=75$ & PSNR/SSIM & 27.09/ 0.6561 &  26.77/ 0.4824  & 26.84/ 0.5177 & 26.64/ 0.4359 & 25.5/ 0.5451 & 23.24/ 0.2298 & \textbf{27.91/ 0.7651} \\

		\M{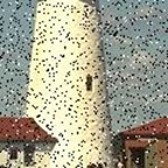}~
		&\M{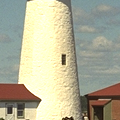}~
        &\M{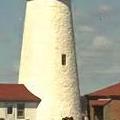}~
         &\M{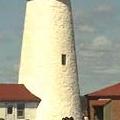}~
		&\M{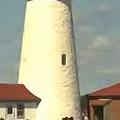}~
  &\M{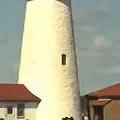}~
		&\M{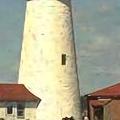}~
		&\M{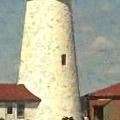}~
		&\M{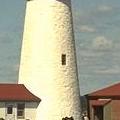}
            % \multicolumn{9}{c}{\includegraphics[width=1.0\linewidth, trim=3 170 0 0,clip]{sections/images/com.pdf}}
		\\ 
		Kodak-21~\cite{franzen1999kodak} & Reference  & N2C~\cite{dncnn} & DBD~\cite{godard2018deep} &
		 N2N~\cite{lehtinen2018noise2noise} &  N2S~\cite{batson2019noise2self} & R2R~\cite{pang2021recorrupted} & LIR~\cite{du2020learning} & MeD (Ours) \\
		S\&P $\hat r=0.3$ &  PSNR/SSIM  & 33.26/ 0.9224 & 34.25/ 0.9413 & 30.83/ 0.8857 & 31.07/ 0.8864 & 29.03/ 0.8377 & 27.40/ 0.6498 & \textbf{36.83}/ 0.9246

	\end{tabular}
    \vspace{-0.0cm}
	\caption{Qualitative denoising results on unseen noise types. All the methods are trained with Gaussian $\sigma=25$. The quantitative PSNR/SSIM results are provided underneath the respective images. Best viewed in colour (zoom-in for a better comparison). 
	}
	
	% \vspace{-0.1cm}
 \label{fig:visual}
\end{figure*}

% -------------------------Visualize Image ------------------------

\subsection{Generalisation on Unseen Noise Removal}
\label{exp:unseen}
% In the last column, we present the results of our MeD trained with Noise Pool (NP) Sampling, which shows a slightly better performance than when trained only with Gaussian noise. Further ablation study on NP sampling is presented in Sec. \ref{exp:np}.

In the previous subsection, we demonstrated the remarkable generalisation ability of our model in the case of Gaussian noise. Here, we aim to extend this investigation to other types of unseen noise and evaluate the denoising generalisation ability of our method. Specifically, we consider Poisson noise, Speckle noise, Local Variance Gaussian noise, and Salt-and-Pepper noise. 
For a more detailed synthetic process, please refer to the supplementary Section \ref{noise}.

% In Figures \ref{fig:visual}, we
% provide the qualitative comparisons of denoising unseen noise using model that only trained with Gaussian $\sigma=25$. It can be observed that our method yields competitive visual quality when compared to unsupervised  and supervised deep denoisers.

First, we demonstrate qualitative comparisons of denoising unseen noise types using models trained only with Gaussian $\sigma = 25$ in Figure \ref{fig:visual}.
 Next, in order to further verify the denoising generalisation ability of MeD, we employ its scene encoder and decoder as pre-trained models to be compared against other methods. It should be noted that the pre-training and fine-tuning methods employed in this study may differ, as shown in Table \ref{tab:unseen}. The pre-training of all test models was conducted on a Gaussian sigma value of 25, followed by fine-tuning with a Gaussian sigma range of 5 to 50. 
Since the training schema of N2S, R2R, and DBD$_4$ differs from MeD, we do not include them in this section. However, evaluations of these methods on unseen noise are still presented in Section \ref{exp:np} under different settings.

\definecolor{Gray}{rgb}{0.95,0.95,0.95}

\begin{table*}
\caption{Analysis of \textit{Noise Pool} on CBSD68 \cite{martin2001database}. All methods were trained using randomly drawn noise from the Noise Pool. }  
 % \vspace{-0.2cm} 
\small
\centering
% \resizebox{\linewidth}{!}{
\setlength{\tabcolsep}{4pt}
\resizebox{1\textwidth}{!}{
     \begin{tabular}{l|cc|ccc|cc}
     \toprule
   \multirow{2}{*}{Test Noise} & \multicolumn{2}{c|}{Noisy/ Clean}& 
     \multicolumn{3}{c|}{Noisy/ Noisy}  &\multicolumn{2}{c}{Invariant Feature} \\
    & N2C \cite{liu2021Swin}  & DBD$_4$ \cite{godard2018deep} & N2N \cite{lehtinen2018noise2noise} & N2S \cite{batson2019noise2self} & R2R \cite{pang2021recorrupted}  & LIR \cite{du2020learning} & MeD (ours) \\
    \midrule
    Gaussian, $\hat \sigma \in[15, 75]$ & \underline{29.24/  0.7754}  & 29.05/  0.7616  & 29.23/  0.7634  & 28.58/  0.7589   & 26.43/
  0.6639  & 26.67/  0.6866 & \textbf{29.61/    0.8178} \\
    
    % & 33.49/  0.8909 & 33.23/ 0.8822  & \underline{33.47}/  0.8289 & 31.88/  \underline{0.8719} & 30.15/  0.8669 & 30.72/  0.8932  & \textbf{33.70/  0.9219}\\
    
    %  Gaussian, $\hat \sigma =25$ & 30.88/  0.8419 & 30.53/  0.8086 & \underline{30.82/  0.8366} & 30.06/  0.8180 & 29.88/  0.8006    & 29.72/  0.7925 & \textbf{31.10/  0.8716} \\
     
    %  Gaussian, $\hat \sigma =50$ & 27.45/  0.7412 & 27.33/  0.7296 & \underline{27.39/  0.7452} & 27.26/  0.7254 & 24.92/  0.6160 & 25.09/  0.6552 & \textbf{27.84/  0.7740}  \\
 
    %  Gaussian, $\hat \sigma =75$ &25.15/  0.6274  & 25.10/  0.6259 &\underline{25.22/  0.6427}&25.12/  0.6204&20.76/  0.3722&21.16/  0.4055&\textbf{25.80/  0.7036} \\

     % 29.61/ 0.8178
     
     Local Var Gaussian (LVG)  &36.64/  \underline{0.9442}&36.18/  0.9307& \underline{36.65}/  0.9235 &33.24/  0.8858&34.70/  0.8779&31.61/  0.8627&\textbf{37.99/  0.9568}  \\
     
     Poisson Noise  &45.72/  0.9764 &44.23/  0.9606&45.64/  0.9799& \underline{46.31/  0.9808} &44.45/  0.9491&43.27/  0.9292& \textbf{48.10/  0.9916}  \\
     
     Speckle, $ \hat v\in [25, 50]$ &\underline{35.58}/  0.9417&35.24/  0.9385& 35.34/  0.9475 & 35.13/  \underline{0.9596} &34.20/  0.9078&33.98/  0.8810 & \textbf{37.21/  0.9715} \\

     S\&P, $\hat r \in [0.3, 0.5]$ &38.85/  0.9165 &37.10/  0.8884&\underline{38.89}/  0.9289&38.22/  \underline{0.9330}&36.17/  0.9087&33.43/  0.8202& \textbf{42.33/  0.9695} \\
     
     \rowcolor{Gray}
    Gaussian $\hat\sigma=25$ + Speckle $\hat v=25$ &  30.19/ 0.8279 & 29.24/ 0.8156 & \underline{30.32/ 0.8317} & 29.51/ 0.8050 & 28.78/ 0.7744 & 29.20/ 0.7871 & \textbf{31.92/ 0.8726} \\
\rowcolor{Gray}
Gaussian $\hat\sigma=50$ + Speckle $\hat v=25$ & \underline{27.30}/ 0.7251 & 26.55/ 0.7126 & 27.23/ \underline{0.7331} &26.91/ 0.7081 & 26.49/ 0.6935 & 26.19/ 0.6941 & \textbf{29.68/ 0.7928} \\
\rowcolor{Gray}
LVG + Poisson & \underline{31.78/ 0.9087} & 31.10/ 0.8842 & 31.60/ 0.7617  & 30.15/ 0.8086  & 28.52/ 0.7144 & 27.33/ 0.7234 & \textbf{34.29/ 0.9325}\\
\rowcolor{Gray}
Poisson + Speckle $\hat v=25$ & 31.39/ \underline{0.9069} & 30.86/ 0.8782 & \underline{31.52}/ 0.8935 &  30.58/ 0.9067 & 30.34/ 0.8897 & 29.93/ 0.8554 
& \textbf{33.04/ 0.9258} \\
     \midrule
     Average & \underline{34.08/  0.8803} & 33.28/  0.8634  & 34.05/  0.8626  & 33.18/  0.8607  & 32.23/
  0.8199  & 31.29/  0.8044  & \textbf{36.02/  0.9145}\\ \bottomrule
% n2c_finer_swinv2_layer2_sigma[5,50]_res48_mixup0.0_noise_pool_new
% n2n_finer_swinv2_layer2_sigma[5,50]_res48_mixup0_noise_pool_pretrain@25_new
% n2s_finer_swinv2_layer2_sigma[5,50]_res48_mixup0_noise_pool_pretrain@25_new  
    \end{tabular}
}
\label{tab:np}
\end{table*}

% ---------------------------- Noise Pool ---------------------------- %

\noindent \textbf{Analysis:} 
Qualitative results in Figure \ref{fig:visual} show that under Gaussian $\sigma = 25$ training settings, our method surpasses other methods in denoising unseen noise types. 
Additionally, Table \ref{tab:unseen} shows that the approaches using pre-trained MeD models outperform their self-transfer models for N2C, N2N, and LIR, with improvements of up to 2 dB in some cases. On average, the MeD pre-trained models show a performance gain of around 0.5 dB across all methods, highlighting the potential of MeD as a powerful pre-training method for image denoising.
It is noteworthy that the self-transfer MeD model exhibits the best denoising performance across all validation noise types, even outperforming the supervised method, N2C. This is particularly evident in Poisson noise, where MeD surpasses N2C by \textbf{$\sim$3~dB}. These results highlight the generalisation ability of our approach in handling unseen noise.

\subsection{Experiments on General Noise Pool}
\label{exp:np}
Here we further investigate the generalisation ability of our method by introducing our general \textit{Noise Pool}. The Noise Pool comprises the five aforementioned types of noise, each with a diverse range of noise levels. 
During training, we randomly sample from the noise pool to provide the model with noisy images. This novel approach simulates a realistic scenario where noise is unknown and can originate from various sources to some extent.

Specifically, we evaluated all methods using the random noise pool approach to train and test on combined or single noise types.  The results are summarised in Table \ref{tab:np}.

% ---------------------- Real Image --------------------------

\begin{figure*}
\begin{center}

\scalebox{0.96}{
\begin{tabular}[b]{c@{ } c@{ }  c@{ } c@{ } c@{ }}\hspace{-8mm}  % .326  
     &   
    \includegraphics[  width=.15\textwidth]{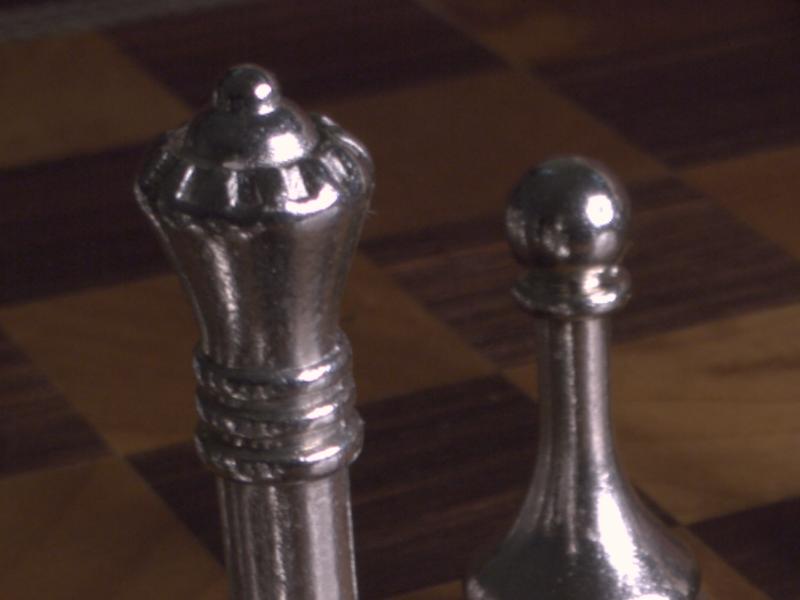}&
  	\includegraphics[width=.15\textwidth]{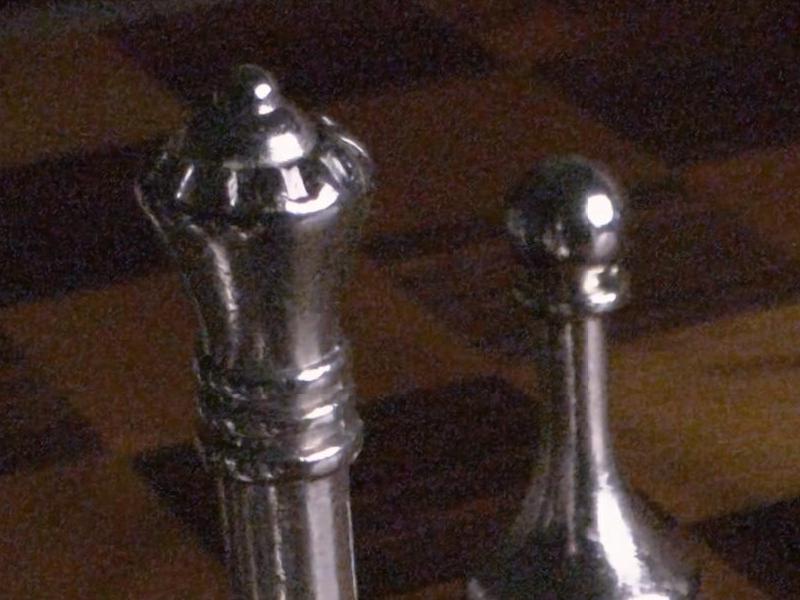}&   
    \includegraphics[width=.15\textwidth]{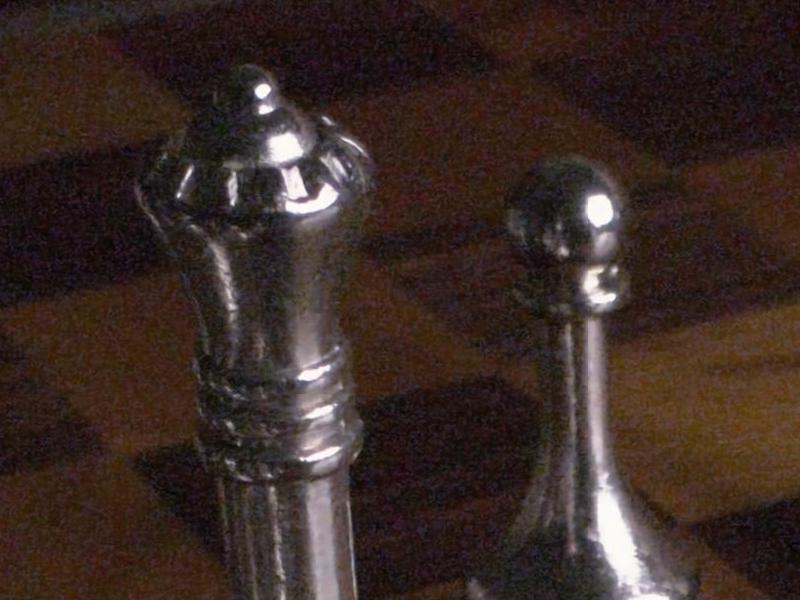}&
      \includegraphics[width=.15\textwidth]{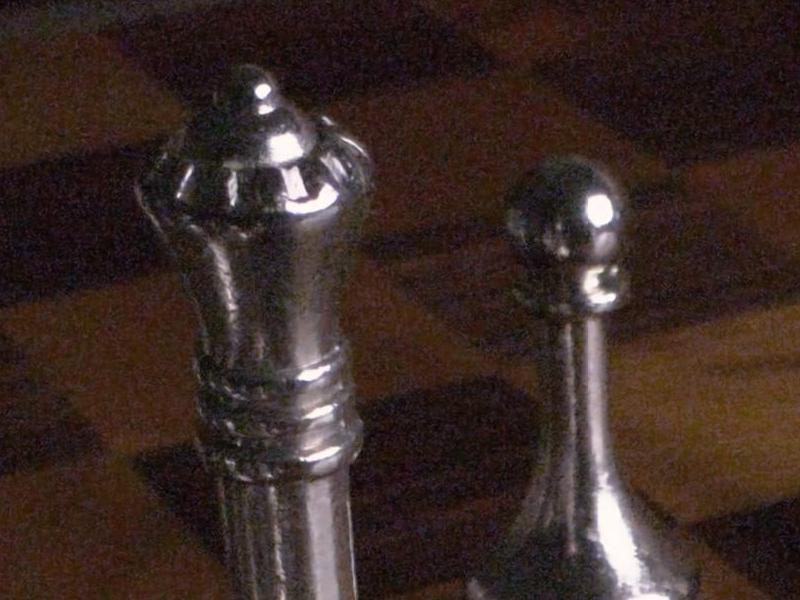}
\\

    &  
    \small~PSNR/ SSIM &\small~ 31.38/ 0.8912  & \small~30.19/ 0.7449  &  \small~29.60/ 0.7053  \\
    & \small~Ground Truth \cite{abdelhamed2018high} & \small~N2C \cite{liu2021Swin}  & \small~DBD$_4$ \cite{godard2018deep}   & \small~N2N \cite{lehtinen2018noise2noise} \\
    
  \multirow[t]{1}{*}{\includegraphics[width=.379\textwidth]{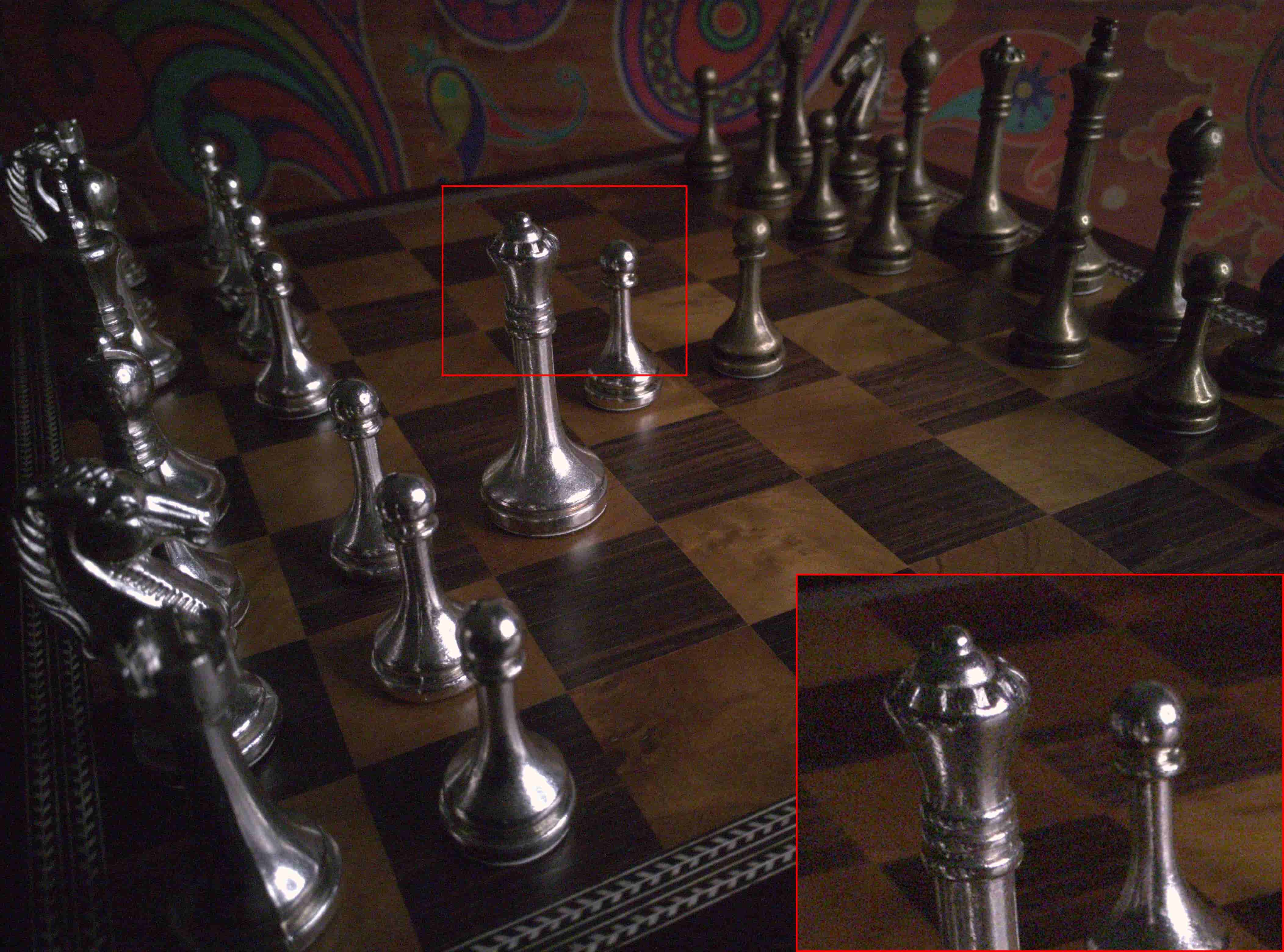}}   
  & \includegraphics[width=.15\textwidth]{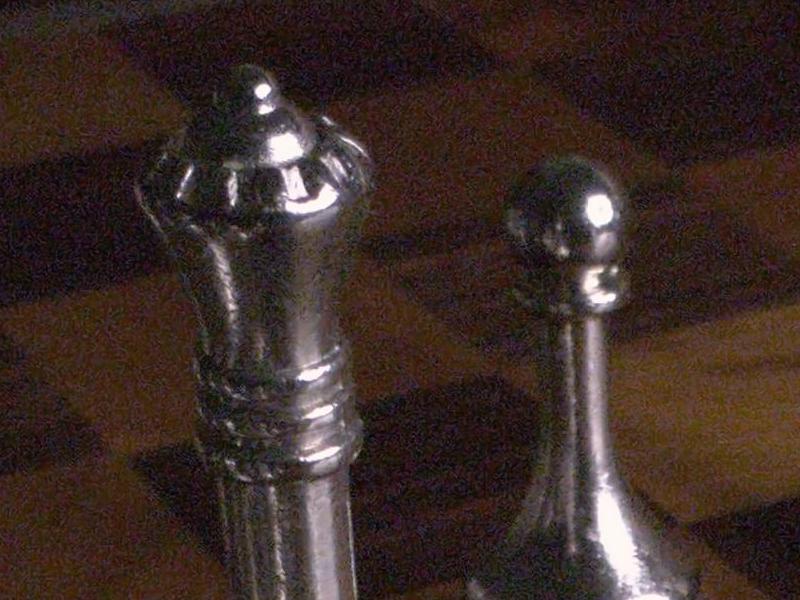}&
    \includegraphics[width=.15\textwidth] {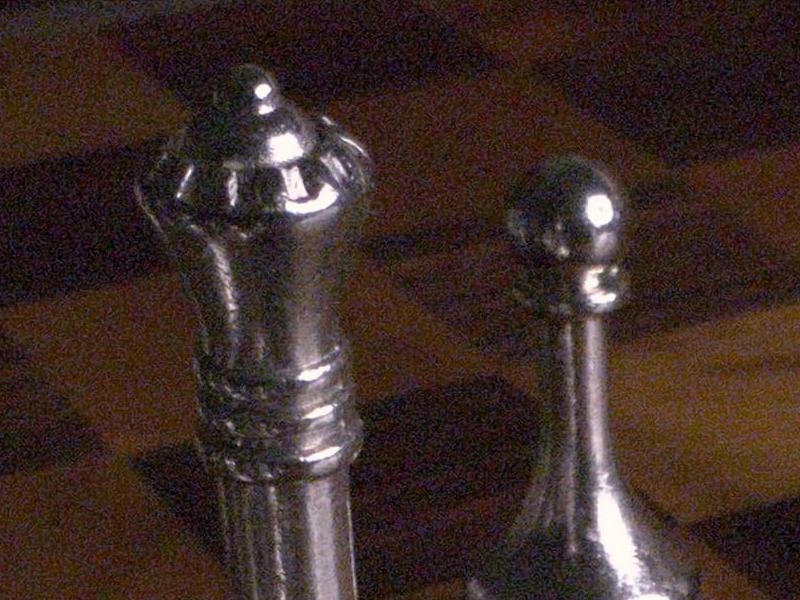}&
     \includegraphics[width=.15\textwidth]{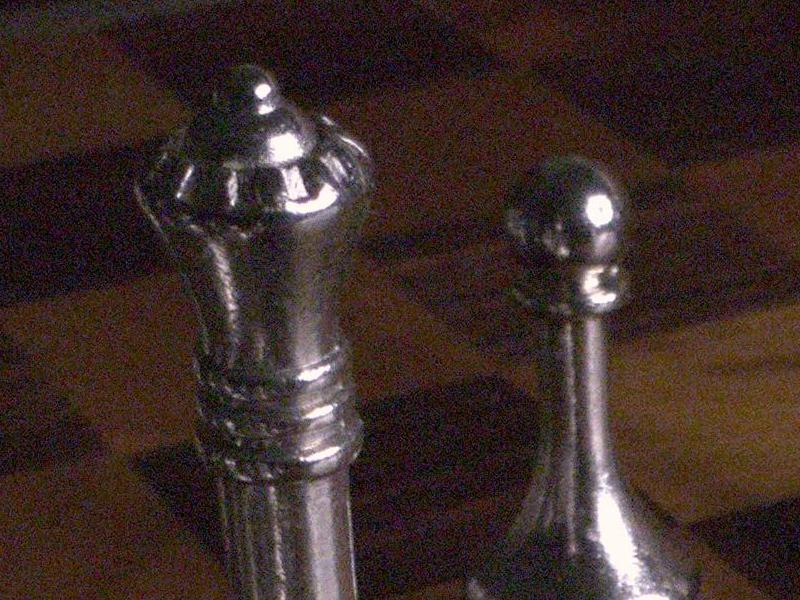}&
     \includegraphics[width=.15\textwidth]{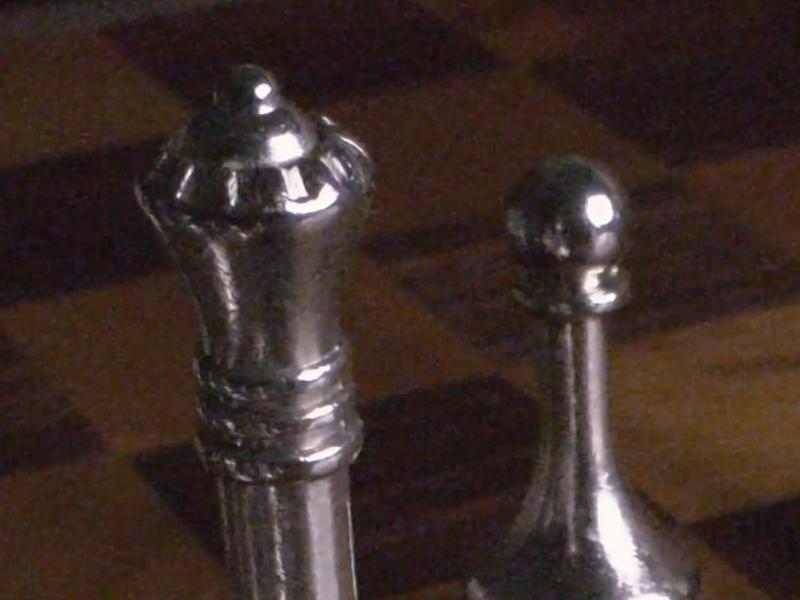}\\
     \small~23.68/ 0.2967& \small~28.58/ 0.6081
     & \small~26.58/ 0.4257 & \small~25.91/ 0.3795  & \small~\textbf{33.07/ 0.8849}\\
           \small~Noisy Image from SIDD \cite{abdelhamed2018high} (ISO 800)  & \small~N2S \cite{batson2019noise2self} & \small~ R2R \cite{pang2021recorrupted}  & \small~ LIR \cite{du2020learning}   & \small MeD (Ours)  \\
           
    &   
    \includegraphics[  width=.15\textwidth]{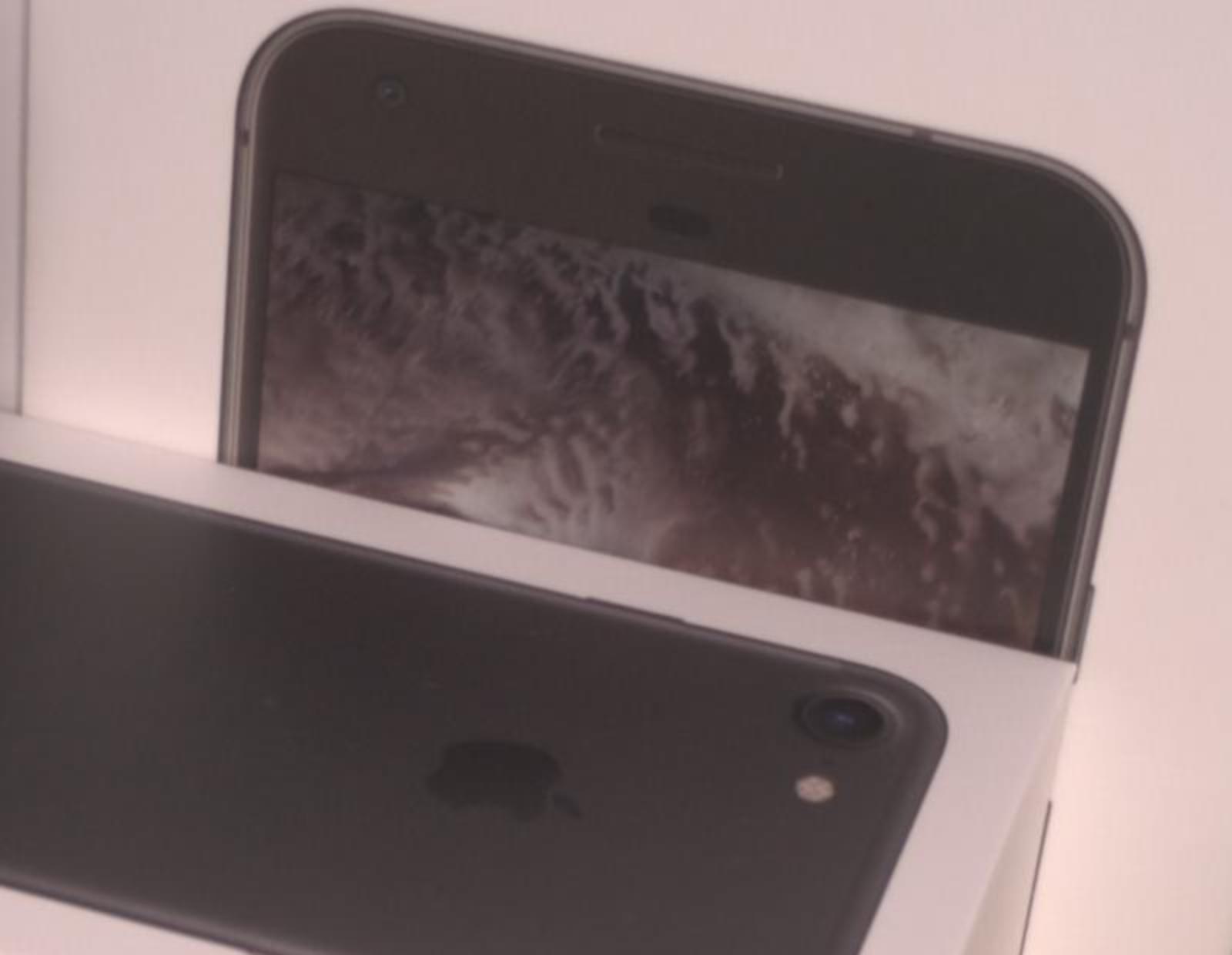}&
  	\includegraphics[width=.15\textwidth]{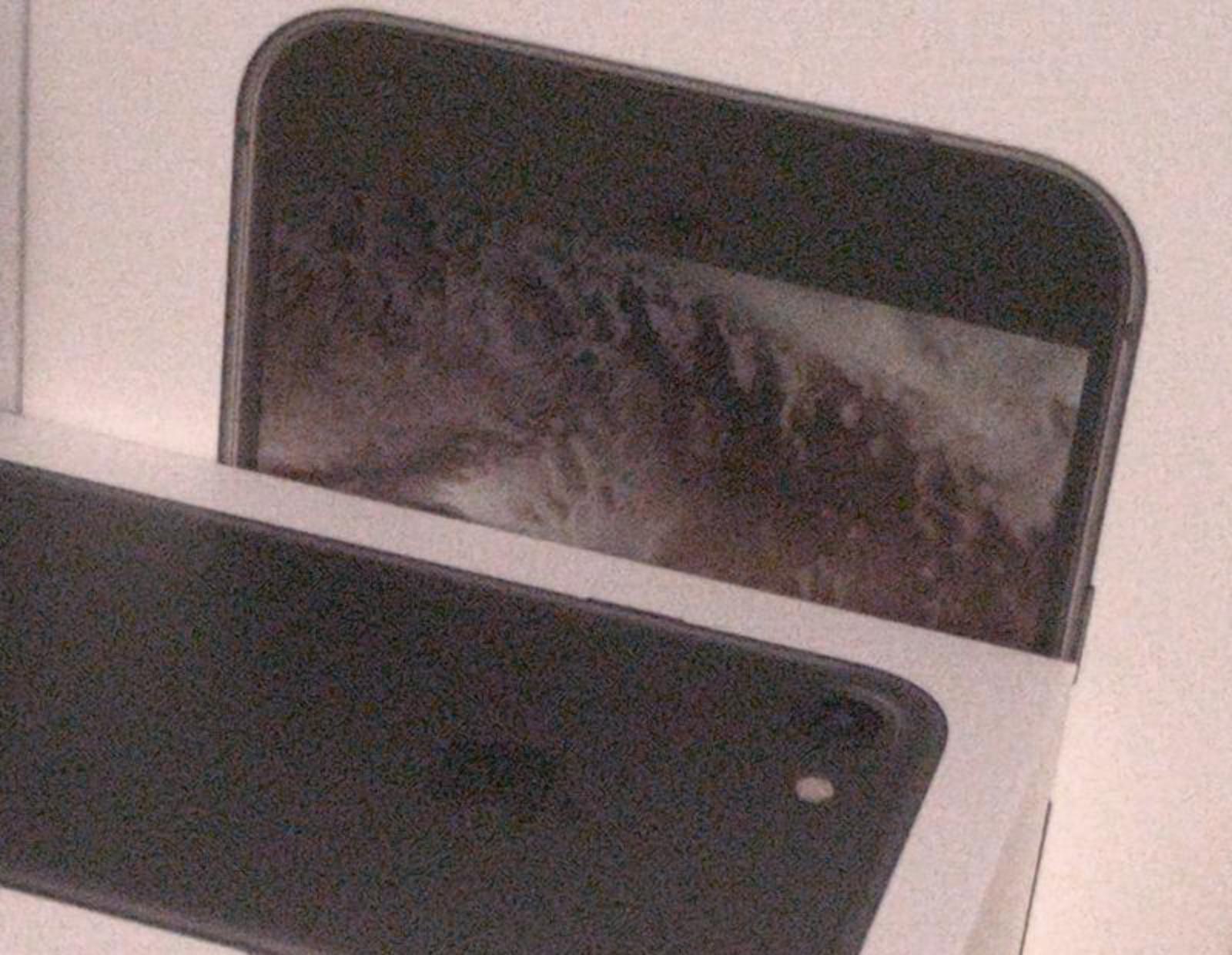}&   
    \includegraphics[width=.15\textwidth]{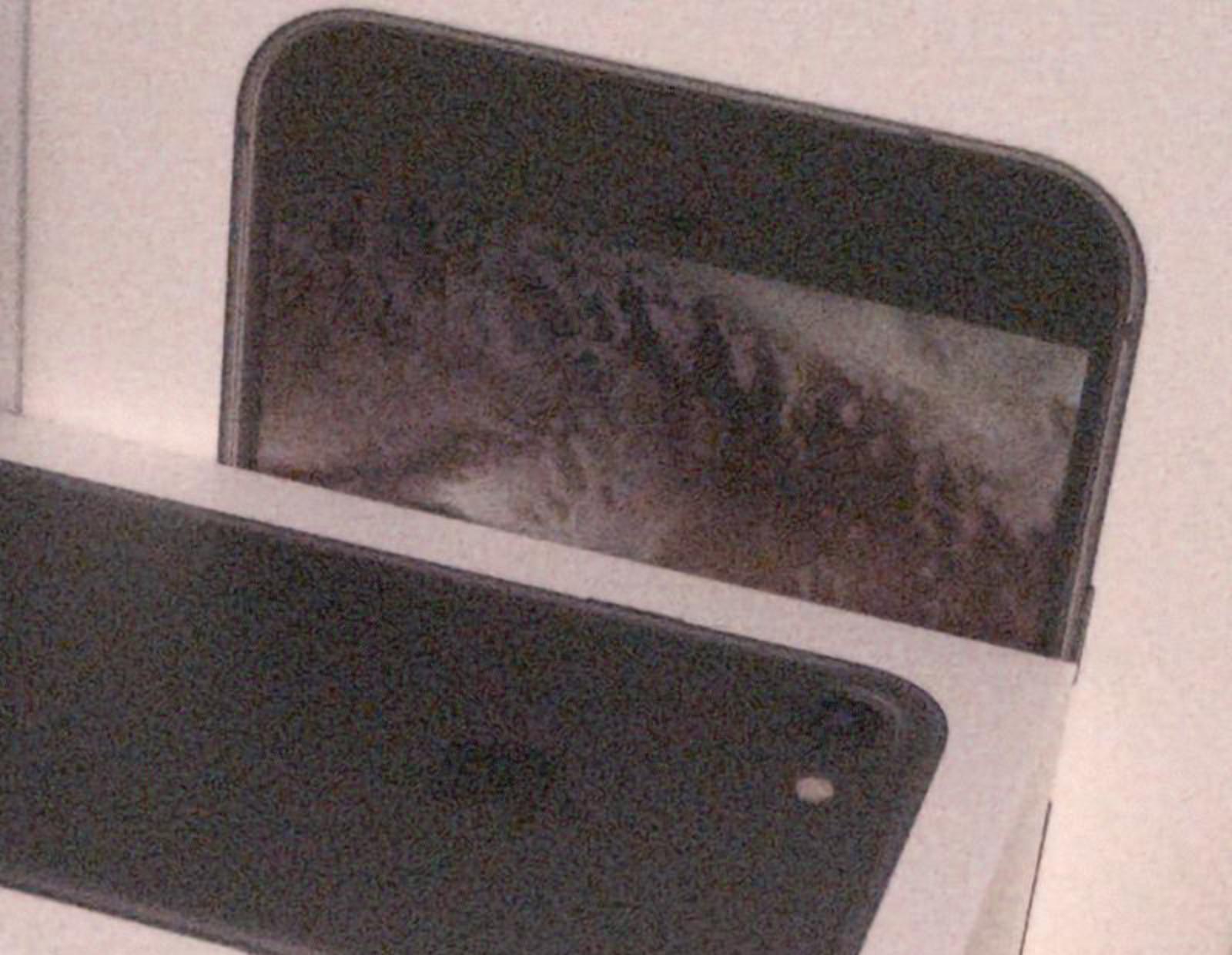}&
      \includegraphics[width=.15\textwidth]{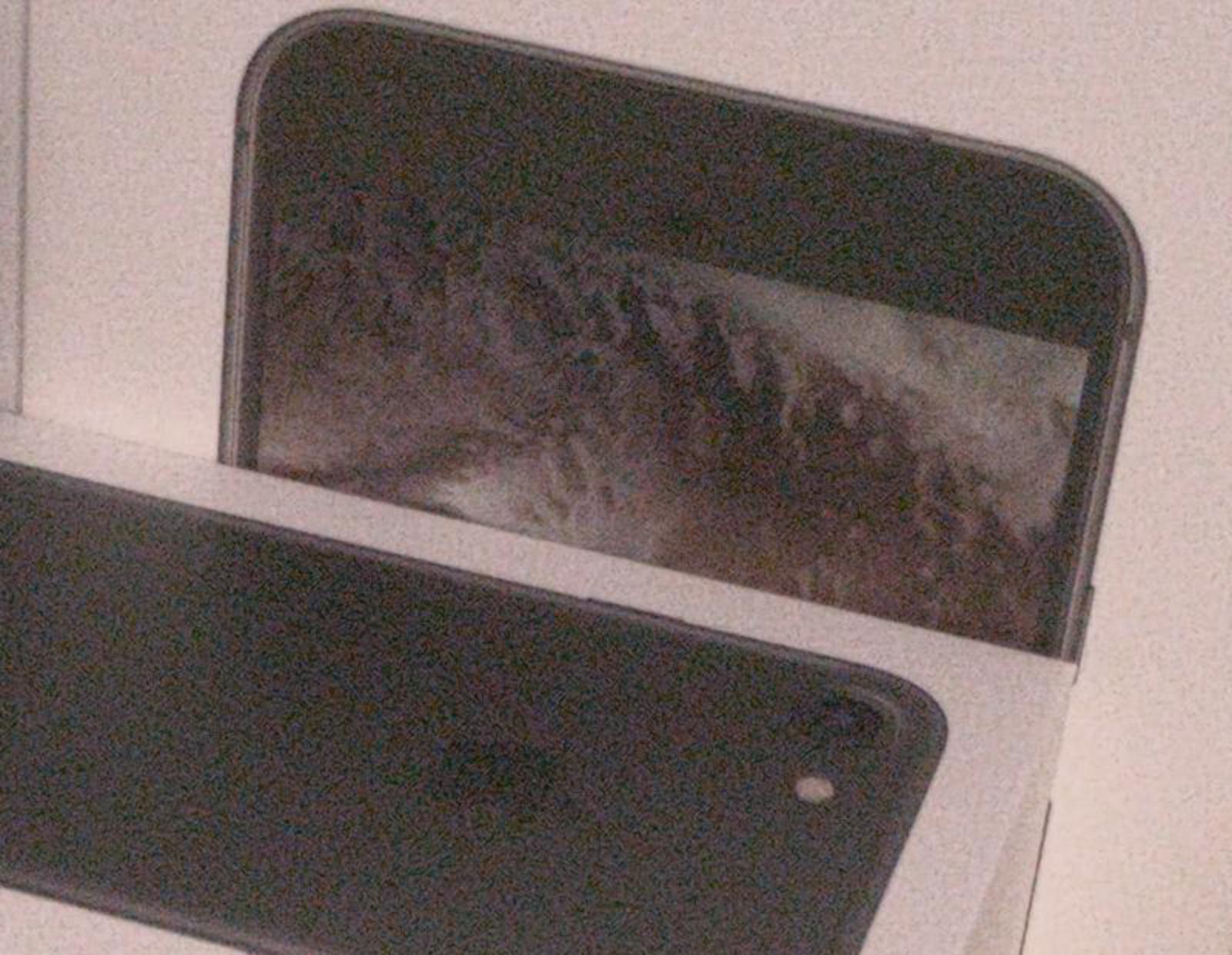}\\
      
    &  
    \small~PSNR/ SSIM &\small~ 28.36/ 0.5661  & \small~27.58/ 0.5406  &  \small~27.23/ 0.49  \\
    & \small~Reference & \small~N2C \cite{liu2021Swin}  & \small~DBD$_4$ \cite{godard2018deep}   & \small~N2N \cite{lehtinen2018noise2noise} \\
    
  \multirow[t]{1}{*}{\includegraphics[width=.379\textwidth]{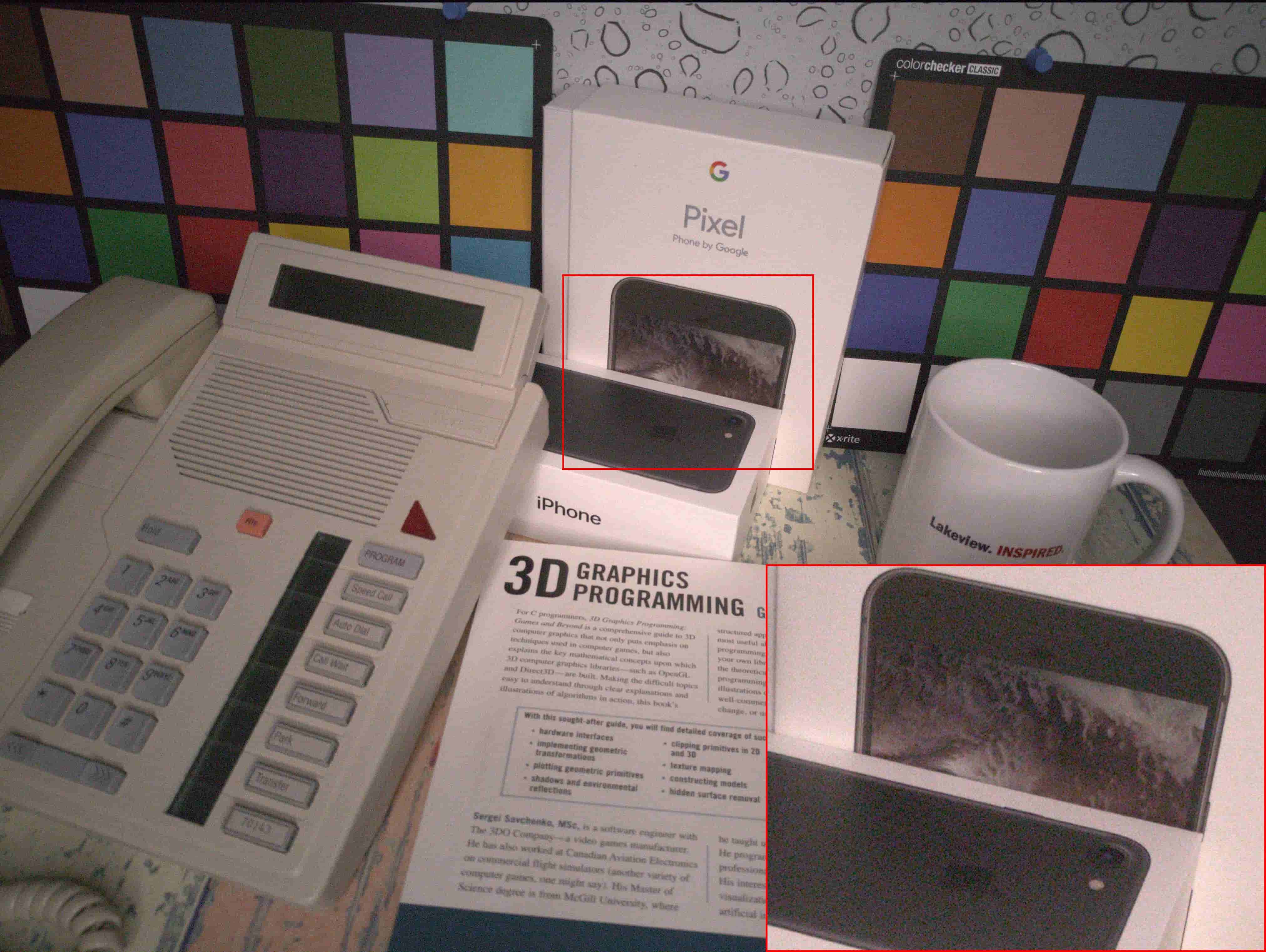}}   
  & \includegraphics[width=.15\textwidth]{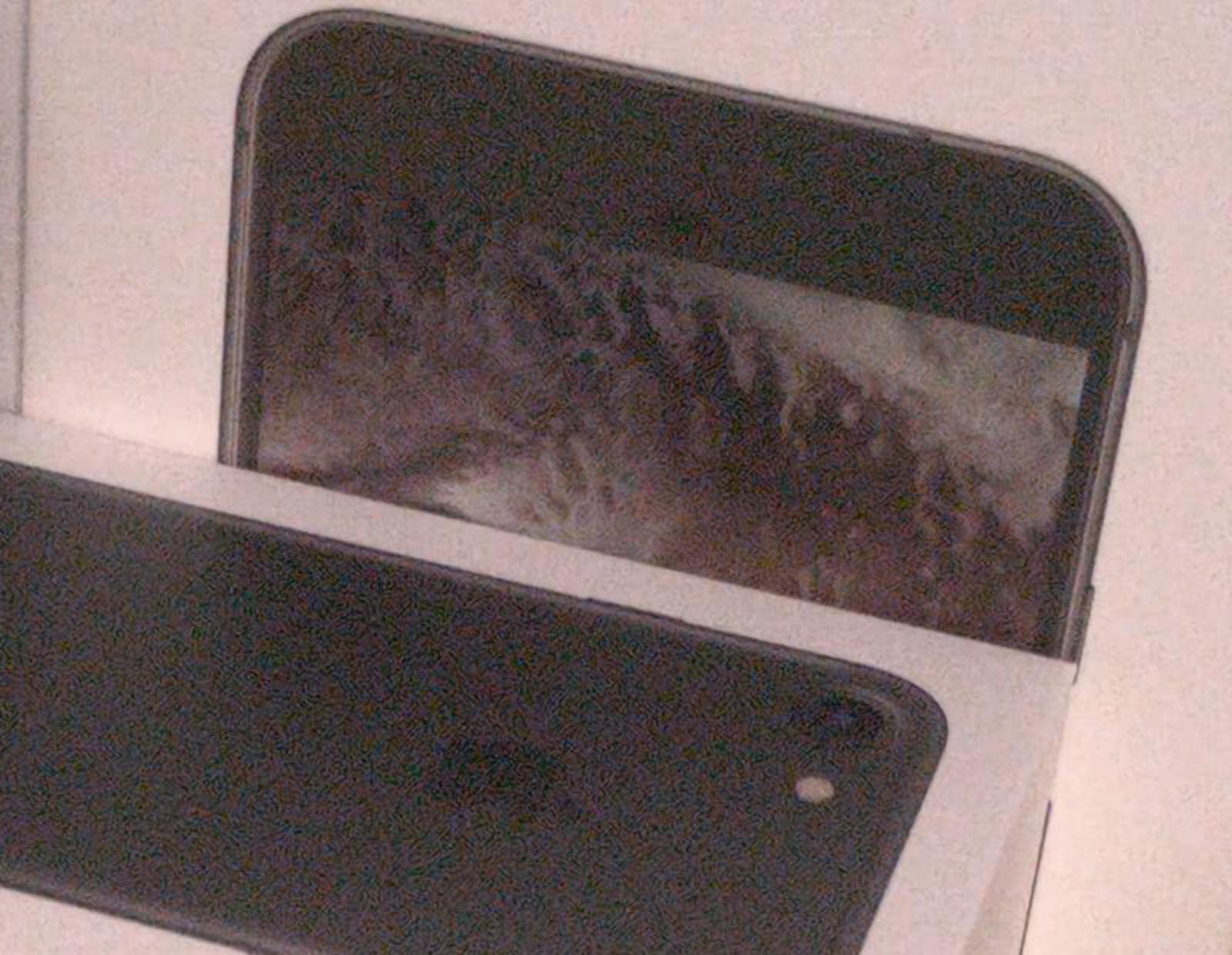}&
    \includegraphics[width=.15\textwidth]{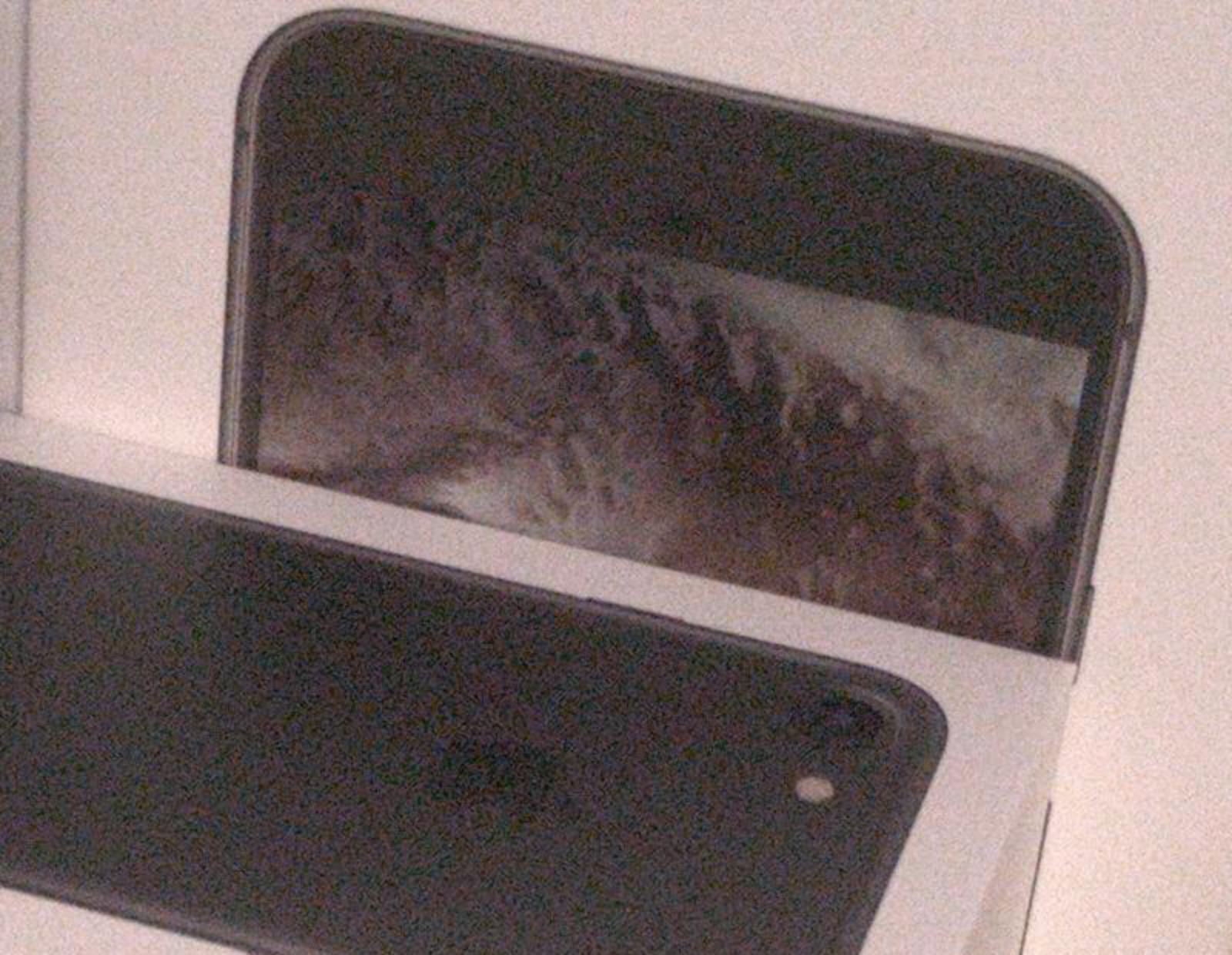}&
     \includegraphics[width=.15\textwidth]{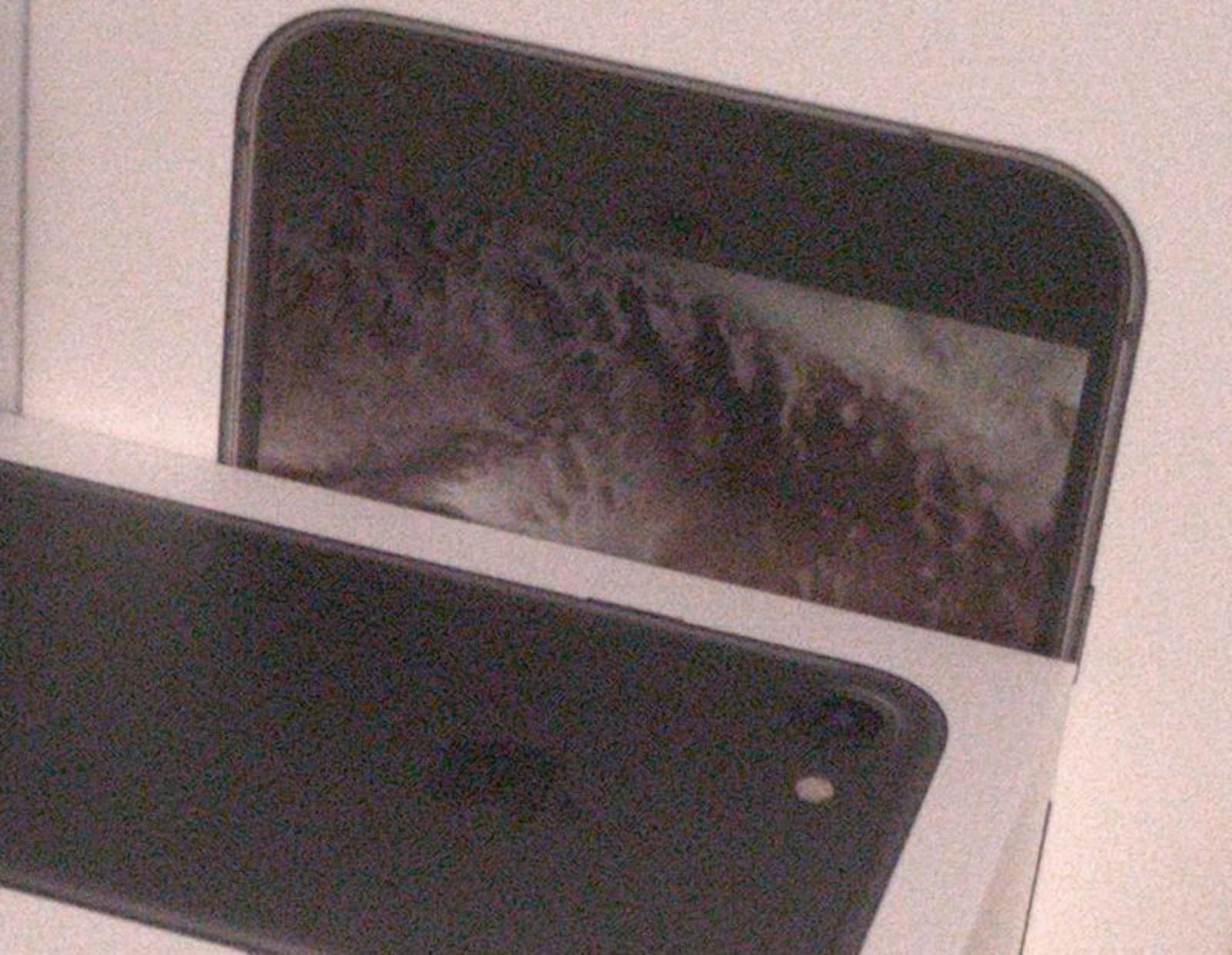}&
     \includegraphics[width=.15\textwidth]{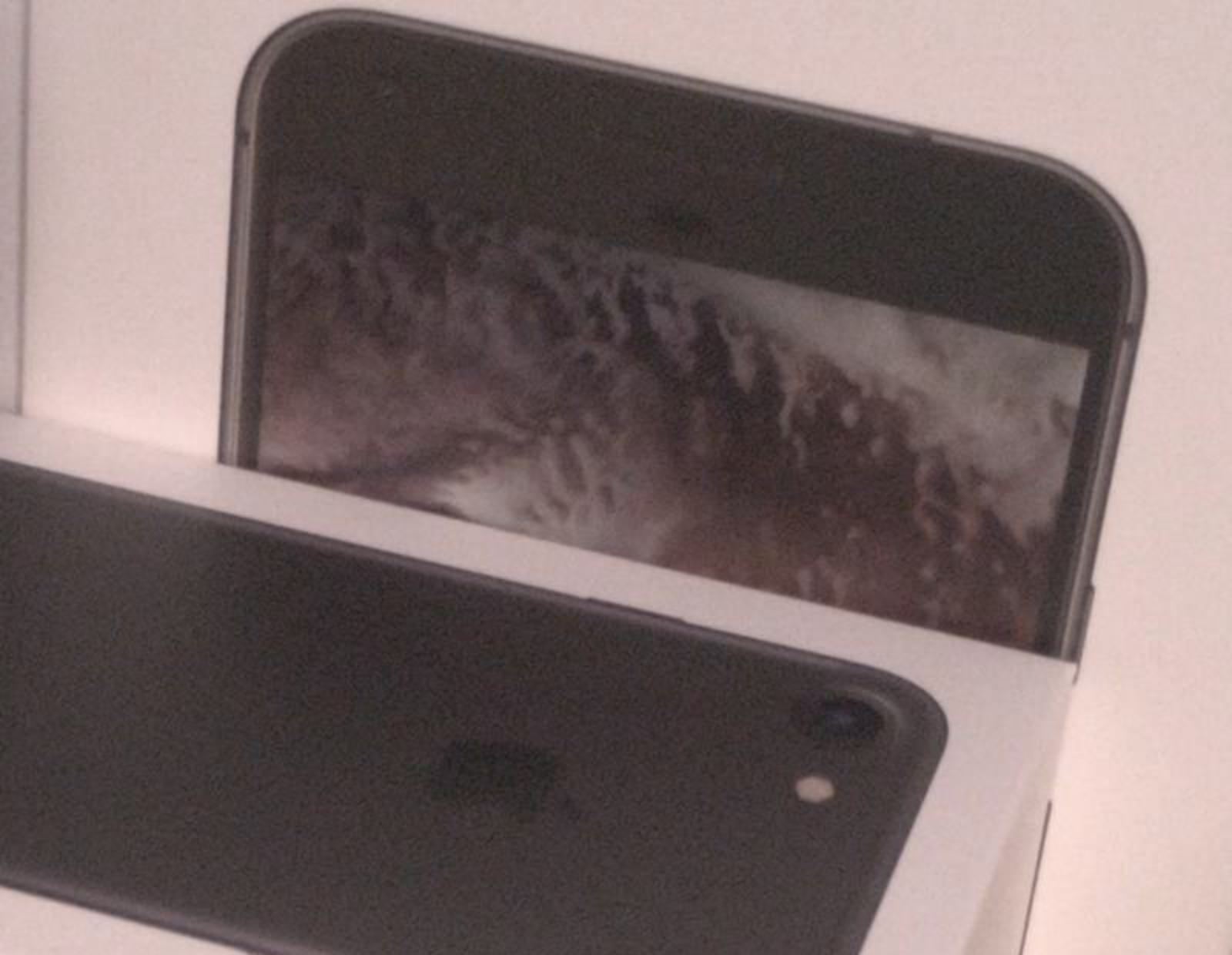}\\
     
     \small~25.33/ 0.3865& \small~27.58/ 0.5516
     & \small~26.77/ 0.4889 & 26.45/ 0.4752  & \small~\textbf{30.23/ 0.6703}\\
           \small~Noisy Image from SIDD  & \small~N2S \cite{batson2019noise2self} & \small~ R2R \cite{pang2021recorrupted}  & \small~ LIR \cite{du2020learning}   & \small  MeD (Ours)
\hspace{-3mm}
\end{tabular}
}

\end{center}
 \vspace{-2mm}
\caption{\textbf{Real Noise Removal Example of SIDD \cite{abdelhamed2018high}}. All the methods are trained with \textit{Noise Pool} on the DIV2K~\cite{agustsson2017ntire} dataset. It can be seen that the proposed MeD can remove much real noise even without training with real-noise distribution (zoom in for a better comparison). 
}
\label{fig:real}
\end{figure*}

% 
% ---------------------- Real Image --------------------------
\noindent \textbf{Analysis:} 
% Compared with results in Table \ref{tab:unseen} and the results presented in Table \ref{tab:np} demonstrate that training on the noise pool can improve the performance of all methods to a certain extent, but our method enjoys most benefit from 
% For instance, our MeD approach achieves an impressive \textbf{11.49 dB} improvement on Salt-and-Pepper noise with the use of the noise pool.
In Table \ref{tab:np}, our MeD approach outperforms all other methods significantly on all the test noise types. For example, when a test noise containing a combination of Gaussian noise with $\hat \sigma=50$ and Speckle noise with $\hat v=25$ is used, other methods exhibit an approximate performance of $\sim $27 dB. However, MeD achieves significantly better results with a performance of 29.68 dB. And on average, MeD exhibits a performance that is approximately 2 dB better than other methods. Our findings show that utilising a comprehensive noise pool for training purposes can effectively improve the generalisation capability. Furthermore, the remarkable denoising generalisation ability of our MeD approach, in comparison to other methods, is particularly advantageous for real-world applications.

% different types of noise, 
% all methods perform worse, but our MeD approach still handles them better than the other approaches. Specifically, our approach outperforms the second-best method by about 2 dB on average.

% In Table \ref{tab:np}, all the methods are the random noise pool and test on single or combined noise. By comparing the Table \ref{tab:np} and Table \ref{tab:unseen}, we can conclude that our method enjoy the most benefit from increasing the number of input noise distributions. For example, N2C increase from 33.36dB to 35.85dB and 35.62dB to 36.64dB in Speckle noise and Local Variance Gaussian noise by using Noise Pool. And overall, MeD lead to $\sim$2 dB improvement in Average performance compared to N2C and N2N, and more with other methods.

% TODOTODOTODOTODO

% As we can see that using the noise pool can boost the performance of the tested methods in Gaussian noise removal compared to only training with random Gaussian noise. However, our method not only gives the best performance on Gaussian noise but also on other types of noise. For example, under the same training conditions,  

% Overall, our results indicate that utilizing a noise pool with diverse noise distributions can significantly improve the denoising performance of deep learning models, and our proposed MeD method can particularly benefit from such training inputs

\subsection{Real Noise Removal}
\label{exp:real}
In our previous experiments, we demonstrated the exceptional denoising performance of our MeD approach on synthetic noises. However, real-world noise is often more complex and challenging than synthetic noise. In this subsection, we aim to evaluate the generalisation performance of our approach on real-world noise by testing it on the SIDD \cite{abdelhamed2018high}, CC \cite{nam2016holistic} and PolyU~\cite{xu2018real} datasets.
To assess the denoising performance on real-world noise, we use the same pre-trained models as in Section \ref{exp:np}. The representative qualitative results on the SIDD dataset in the standard RGB colour space are presented in Figure \ref{fig:real}. 

\begin{table}
	\caption{ Quantitative result obtained from the application of various methods trained on a general Noise Pool to real noise datasets.}
	\label{tab:real}
	% \small
	\centering
 \resizebox{\columnwidth}{!}{
    \begin{tabular}{l|cccc}
     \toprule
   Method & PolyU \cite{xu2018real} & SIDD \cite{abdelhamed2018high} & CC \cite{nam2016holistic} & Average \\
    \midrule
    
     N2C \cite{liu2021Swin}& 35.89/  0.9652 & 30.37/  0.6028  &   \underline{37.89}/ 0.9408 & 34.72/ 0.8363 \\
     DBD$_4$  \cite{godard2018deep}& 35.69/  0.9571 & 30.23/ 0.6173 & 37.74/  0.9357 & 34.55/ 0.8367 \\
     N2N \cite{lehtinen2018noise2noise}&  36.22/  0.9679	&  \underline{32.82/ 0.7297} & 37.39/  0.9570 & \underline{35.48/ 0.8849}  \\
     N2S \cite{batson2019noise2self} & \underline{36.41/  0.9721}  &	30.98/ 0.6018 & 37.58/  \underline{0.9622}	& 34.99/ 0.8454 \\
     R2R  \cite{pang2021recorrupted}& 34.58/  0.8890 &	29.64/ 0.5708   &  35.35/  0.8478 & 33.19/ 0.7692 \\
     LIR \cite{du2020learning}&  34.81/  0.7278  & 28.76/  0.5296  & 35.50/  0.8403 & 33.02/ 0.6992 \\
     \midrule
     MeD (ours) & \textbf{38.65/ 0.9855} & \textbf{35.81/ 0.8278} & \textbf{40.08/  0.9745 }     & \textbf{38.18/ 0.9293}
 
     \\ \bottomrule
    \end{tabular}%
    }
% \hspace{-0.7cm}
\end{table}
% \vspace{-0.5}

\noindent \textbf{Analysis:} 
As shown in Table~\ref{tab:real}, our approach significantly outperforms all other methods across all three datasets, with a performance improvement of \textbf{2-3 dB} over the second-best approach, and also consistently outperforms its supervised counterparts (\ie N2C and DBD$_4$) by over \textbf{3 dB}. These results suggest the effectiveness and generalisability of the proposed approach in real-world denoising scenarios. 

\hq{Our approach achieves remarkable performance on real-world noise without even being trained on more expensive real-world data. For a more complete study, we also conduct experiments on model training with real-world data (more details please refer to the supplementary Table~\ref{tab:realreal}), showing superior performance and even better generalisation ability to data out of the training data distribution.
}

In Figure \ref{fig:real}, the presence of noise persists even after applying denoising techniques, yet ours demonstrates the most authentic outcomes compared to others. For instance, while the noise particles remain prominent in the N2C results, they are absent in our results. Overall, the results indicate that the MeD approach is well-suited for real-world denoising tasks, providing a robust and reliable solution for improving image quality in challenging environments.

\subsection{Expand to More Views}
\label{exp:lambda}

\hq{
% In our previous experiments, we used a two-view method for a fair comparison. However, it's important to note that our proposed method can be expanded to incorporate more views. To investigate the impact of utilising different numbers of views, we conducted controlled experiments using 2, 3, and 4 views. The performance results under different synthesised noise conditions are presented in Table \ref{tab:more views}.
Although we only showcase two views for the experiments above, our method can be easily expanded to multiple views. To investigate the impact of the numbers of views, here we further conduct a study comparing 2, 3, and 4 views, in Table \ref{tab:more views}.
}

\begin{table}
    \caption{Multiple views ($\geq2$) study, with average PSNR/SSIM.}
    \label{tab:more views}
    % \vspace{0.5mm}
    % \small
    \centering
    \resizebox{\columnwidth}{!}{
    \begin{tabular}{c|ccccc}
        \toprule
        \#Views & Gaussian & LVG & Poission & Speckle & S\&P  \\
        \midrule
        2 & 29.61/ 0.8178 &   37.99/ 0.9568 & 48.10/ 0.9916 & 37.21/ 0.9715 & 42.33/ 0.9695\\
        3  & 29.68/ 0.8197 &  38.05/ 0.9577 &  48.23/ 0.9920 &  37.40/ 0.9733 &  42.45/ 0.9703\\
        4 & 29.70/ 0.8204 & 38.08/ 0.9580 & 48.31/ 0.9921 & 37.47/ 0.9740 & 42.49/ 0.9709 
        \\ \bottomrule
    \end{tabular}%
    }
    \label{tab:view}
\end{table}

% \noindent \textbf{Analysis: }
\hq{
The results indicate that increasing the number of views consistently improves the performance across different noise types. For example, when dealing with Speckle noise, the 4-view model achieves a 0.26 dB higher PSNR than the 2-view model. 
% These performance gains validate that incorporating additional views enhances the learning of robust scene representations. 
However, it is worth noting that employing $n$ views requires $n!$ cross-computations within each view pair during training, resulting in a significant increase in computational cost (\eg from 2-view to 4-view leads to a $10\times$ training time increase in our experiment). 
% In our experiments, when transitioning from 2 views to 4 views on the same hardware, the training time increased approximately tenfold.
}

\subsection{More Application Exploration}
\label{exp:more}
% In addition to denoising tasks, 
% the MeD model has also demonstrated strong performance in 
Here we investigate the potential of the proposed MeD for other more general image restoration tasks, such as image super-resolution and inpainting. 
% \noindent \textbf{Remark: }
In this study, we generalise the previously defined \textit{degradation (noise)} to a residual image between a clean image and a corrupted image. Moreover, we expand the definition from \textit{Noise Pool} to a more general one -- \textit{Corruption Pool} that contains not only noise but also general corruption.  

% \hspace{0.2cm}

 \begin{table}
	\caption{Average PSNR/SSIM of super-resolution results on Set5 \cite{bevilacqua2012low}. Learning-based methods are trained with \textit{Corruption Pool}. }
	\label{tab:sr}
 % \vspace{0.2cm}
	% \small
	\centering
	% \resizebox{\linewidth}{!}{
	% \setlength{\tabcolsep}{4pt}
	%\renewcommand{\arraystretch}{0.98}
 \resizebox{\columnwidth}{!}{
    \begin{tabular}{l|cccc}
     \toprule
   Scale & Bicubic &  RCAN \cite{zhang2018image} &  DASR \cite{wang2021unsupervised} & MeD (ours) \\
    \midrule
$\times 2$ & 33.63/ 0.9285 & 36.12/ 0.9339  & \underline{36.98/ 0.9471} & \textbf{37.12/ 0.9527}\\
$\times 3$ & 30.37/ 0.8652  & \underline{34.15/ 0.9286} & 34.11/ 0.9187 &  \textbf{34.92/ 0.9294} \\
$\times 4$ & 28.35/ 0.8084 & \underline{31.94/ 0.8871} & 31.54/ 0.8736  & \textbf{32.50/ 0.8956} \\

 \bottomrule
    \end{tabular}
}
% \hspace{-0.3cm}
\end{table}
% \vspace{-0.5}

\paragraph{Super resolution.} 
% \noindent \textbf{Super resolution.} 
Image super-resolution aims to enlarge the resolution of a low-resolution image. We train our method on the DIV2K dataset \cite{Agustsson_2017_CVPR_Workshops}, where we randomly choose different downscale methods from a \textit{Corruption Pool} that consists of random Gaussian noise and four types of down-scaling (bicubic, lanczos, bilinear, and hamming). We benchmark our method against the supervised method RCAN \cite{zhang2018image} that aims for high PSNR and the recent unsupervised methods DASR \cite{wang2021unsupervised} that are specialised for super-resolution. We conduct our evaluation on the Set5 dataset \cite{bevilacqua2012low} with scaling factors of 2, 3, and 4. The results in Table \ref{tab:sr} show the effectiveness of our method over both supervised and unsupervised approaches.

\paragraph{Inpainting.} 
% \noindent \textbf{Inpainting.} 
\hq{
We also apply our method to the image inpainting task, which fills in missing pixels. }

\hq{
We choose two single-image deep learning methods -- Self2Self (S2S) \cite{quan2020self2self} and DIP \cite{ulyanov2018deep}, for comparison. Our MeD is trained with \textit{Corruption Pool} containing noises, down-scaling, and inpainting mask operations altogether. }

\hq{
To compare our method (MeD) with other state-of-the-art methods, we conduct experiments on the Set 11 dataset \cite{ulyanov2018deep} with three different pixel dropping ratios: 50\%, 70\%, and 90\%. The results are shown in Table \ref{tab:im}, suggesting the effectiveness of MeD again in the image inpainting task.
}

\begin{table}
	\caption{Average PSNR/SSIM of inpainting results on Set11 \cite{ulyanov2018deep}. S2S and DIP are trained and tested on the same single image. MeD  is trained with \textit{Corruption Pool}.}
 \vspace{0.2cm}
	\small
	\centering
	% \resizebox{\linewidth}{!}{
	\setlength{\tabcolsep}{4pt}
 \resizebox{0.45\textwidth}{!}{
    \begin{tabular}{c|ccc}
     \toprule
   Dropping Ratio & DIP \cite{ulyanov2018deep} &  S2S \cite{quan2020self2self} &  MeD (Ours)\\
    \midrule
50\% & 33.45/ 0.9217 & {34.91/ 0.9479} & \textbf{36.24/ 0.9617} \\
70\% & 28.53/ 0.8501 & {30.94/ 0.8845} & \textbf{31.05/ 0.9161} \\
90\% & 24.39/ 0.7360 & {25.97/ 0.7933} & \textbf{26.01/ 0.8052} \\

 \bottomrule
    \end{tabular}%
    }
	% }
	\label{tab:im}
\end{table}
% \vspace{-0.5}

\section{Conclusion}
In this paper, we have presented a new self-supervised learning method (MeD) for image denoising that disentangles scene and noise features in a constraint feature space. Our approach has demonstrated exceptional denoising performance in both synthetic and real-world noise scenarios, with particularly significant performance on real-world noise. MeD can handle complex noise with better performance than other state-of-the-art methods, as validated by consistent performance gain across various datasets and noise types. Our approach has decent generalisation ability, requiring only noisy images for training and efficiently denoising real-world noise without seeing any clean ground truth data. This opens up new possibilities for training deep models without the need for costly labelled data. Furthermore, our model can be easily adapted to other low-level image restoration tasks. We hope this could provide a new baseline for future research in image disentanglement and the extension to other image processing tasks.
% In this work, we have presented a novel self-supervised learning method, MeD, for image denoising that disentangles the scene and noise features in a constraint feature space. Our approach uses only noisy images for training and does not require clean ground truth data, making it applicable to a wide range of real-world scenarios where such data is often not available. By leveraging the idea of disentanglement, MeD is able to perform well in real-world scenarios with complex noise distributions, outperforming several state-of-the-art self-supervised and supervised learning methods, including DnCNN, under certain circumstances. Our method opens up new possibilities for training deep models without the need for costly labelled data and can be extended to other domains.

\subsection*{Acknowledgement}
The computations described in this research were performed using the Baskerville Tier 2 HPC service\footnote{https://www.baskerville.ac.uk/}. Baskerville was funded by the EPSRC and UKRI through the World Class Labs scheme (EP/T022221/1) and the Digital Research Infrastructure programme (EP/W032244/1) and is operated by Advanced Research Computing at the University of Birmingham.

{\small
\bibliographystyle{ieee_fullname}
\bibliography{egbib}

\begin{thebibliography}{10}\itemsep=-1pt

\bibitem{abdelhamed2018high}
Abdelrahman Abdelhamed, Stephen Lin, and Michael~S Brown.
\newblock A high-quality denoising dataset for smartphone cameras.
\newblock In {\em Proceedings of the IEEE Conference on Computer Vision and
  Pattern Recognition}, pages 1692--1700, 2018.

\bibitem{SIDD_2018_CVPR}
Abdelrahman Abdelhamed, Stephen Lin, and Michael~S. Brown.
\newblock A high-quality denoising dataset for smartphone cameras.
\newblock In {\em IEEE Conference on Computer Vision and Pattern Recognition
  (CVPR)}, June 2018.

\bibitem{agustsson2017ntire}
Eirikur Agustsson and Radu Timofte.
\newblock Ntire 2017 challenge on single image super-resolution: Dataset and
  study.
\newblock In {\em Proceedings of the IEEE conference on computer vision and
  pattern recognition workshops}, pages 126--135, 2017.

\bibitem{Agustsson_2017_CVPR_Workshops}
Eirikur Agustsson and Radu Timofte.
\newblock Ntire 2017 challenge on single image super-resolution: Dataset and
  study.
\newblock In {\em The IEEE Conference on Computer Vision and Pattern
  Recognition (CVPR) Workshops}, July 2017.

\bibitem{batson2019noise2self}
Joshua Batson and Loic Royer.
\newblock Noise2self: Blind denoising by self-supervision.
\newblock In {\em International Conference on Machine Learning}, pages
  524--533. PMLR, 2019.

\bibitem{bereziat2011solving}
Dominique B{\'e}r{\'e}ziat and Isabelle Herlin.
\newblock Solving ill-posed image processing problems using data assimilation.
\newblock {\em Numerical Algorithms}, 56(2):219--252, 2011.

\bibitem{bevilacqua2012low}
Marco Bevilacqua, Aline Roumy, Christine Guillemot, and Marie~Line
  Alberi-Morel.
\newblock Low-complexity single-image super-resolution based on nonnegative
  neighbor embedding.
\newblock 2012.

\bibitem{chen2022simple}
Liangyu Chen, Xiaojie Chu, Xiangyu Zhang, and Jian Sun.
\newblock Simple baselines for image restoration.
\newblock In {\em European Conference on Computer Vision}, pages 17--33.
  Springer, 2022.

\bibitem{dong2015compression}
Chao Dong, Yubin Deng, Chen~Change Loy, and Xiaoou Tang.
\newblock Compression artifacts reduction by a deep convolutional network.
\newblock In {\em Proceedings of the IEEE International Conference on Computer
  Vision}, pages 576--584, 2015.

\bibitem{dong2014learning}
Chao Dong, Chen~Change Loy, Kaiming He, and Xiaoou Tang.
\newblock Learning a deep convolutional network for image super-resolution.
\newblock In {\em European conference on computer vision}, pages 184--199.
  Springer, 2014.

\bibitem{du2020learning}
Wenchao Du, Hu Chen, and Hongyu Yang.
\newblock Learning invariant representation for unsupervised image restoration.
\newblock In {\em Proceedings of the IEEE/CVF conference on computer vision and
  pattern recognition}, pages 14483--14492, 2020.

\bibitem{franzen1999kodak}
Rich Franzen.
\newblock Kodak lossless true color image suite.
\newblock {\em source: http://r0k. us/graphics/kodak}, 4(2), 1999.

\bibitem{godard2018deep}
Cl{\'e}ment Godard, Kevin Matzen, and Matt Uyttendaele.
\newblock Deep burst denoising.
\newblock In {\em Proceedings of the European conference on computer vision},
  pages 538--554, 2018.

\bibitem{guo2016building}
Jun Guo and Hongyang Chao.
\newblock Building dual-domain representations for compression artifacts
  reduction.
\newblock In {\em European Conference on Computer Vision}, pages 628--644.
  Springer, 2016.

\bibitem{krull2019noise2void}
Alexander Krull, Tim-Oliver Buchholz, and Florian Jug.
\newblock Noise2void-learning denoising from single noisy images.
\newblock In {\em Proceedings of the IEEE/CVF conference on computer vision and
  pattern recognition}, pages 2129--2137, 2019.

\bibitem{kulkarni2016reconnet}
Kuldeep Kulkarni, Suhas Lohit, Pavan Turaga, Ronan Kerviche, and Amit Ashok.
\newblock Reconnet: Non-iterative reconstruction of images from compressively
  sensed measurements.
\newblock In {\em Proceedings of the IEEE conference on computer vision and
  pattern recognition}, pages 449--458, 2016.

\bibitem{kupyn2019deblurgan}
Orest Kupyn, Tetiana Martyniuk, Junru Wu, and Zhangyang Wang.
\newblock Deblurgan-v2: Deblurring (orders-of-magnitude) faster and better.
\newblock In {\em Proceedings of the IEEE/CVF International Conference on
  Computer Vision}, pages 8878--8887, 2019.

\bibitem{NEURIPS2019_2119b8d4}
Samuli Laine, Tero Karras, Jaakko Lehtinen, and Timo Aila.
\newblock High-quality self-supervised deep image denoising.
\newblock In H. Wallach, H. Larochelle, A. Beygelzimer, F. d\textquotesingle
  Alch\'{e}-Buc, E. Fox, and R. Garnett, editors, {\em Advances in Neural
  Information Processing Systems}, volume~32. Curran Associates, Inc., 2019.

\bibitem{lehtinen2018noise2noise}
Jaakko Lehtinen, Jacob Munkberg, Jon Hasselgren, Samuli Laine, Tero Karras,
  Miika Aittala, and Timo Aila.
\newblock Noise2noise: Learning image restoration without clean data.
\newblock {\em arXiv preprint arXiv:1803.04189}, 2018.

\bibitem{li2020recurrent}
Jingyuan Li, Ning Wang, Lefei Zhang, Bo Du, and Dacheng Tao.
\newblock Recurrent feature reasoning for image inpainting.
\newblock In {\em Proceedings of the IEEE/CVF Conference on Computer Vision and
  Pattern Recognition}, pages 7760--7768, 2020.

\bibitem{liang2021swinir}
Jingyun Liang, Jiezhang Cao, Guolei Sun, Kai Zhang, Luc Van~Gool, and Radu
  Timofte.
\newblock Swinir: Image restoration using swin transformer.
\newblock In {\em Proceedings of the IEEE/CVF International Conference on
  Computer Vision}, pages 1833--1844, 2021.

\bibitem{lim2017enhanced}
Bee Lim, Sanghyun Son, Heewon Kim, Seungjun Nah, and Kyoung Mu~Lee.
\newblock Enhanced deep residual networks for single image super-resolution.
\newblock In {\em Proceedings of the IEEE conference on computer vision and
  pattern recognition workshops}, pages 136--144, 2017.

\bibitem{liu2020learning}
Yu-Lun Liu, Wei-Sheng Lai, Ming-Hsuan Yang, Yung-Yu Chuang, and Jia-Bin Huang.
\newblock Learning to see through obstructions.
\newblock In {\em Proceedings of the IEEE/CVF Conference on Computer Vision and
  Pattern Recognition}, pages 14215--14224, 2020.

\bibitem{liu2021Swin}
Ze Liu, Yutong Lin, Yue Cao, Han Hu, Yixuan Wei, Zheng Zhang, Stephen Lin, and
  Baining Guo.
\newblock Swin transformer: Hierarchical vision transformer using shifted
  windows.
\newblock In {\em Proceedings of the IEEE/CVF International Conference on
  Computer Vision}, 2021.

\bibitem{lu2019uid}
Boyu Lu, Jun-Cheng Chen, and Rama Chellappa.
\newblock Uid-gan: Unsupervised image deblurring via disentangled
  representations.
\newblock {\em IEEE Transactions on Biometrics, Behavior, and Identity
  Science}, 2(1):26--39, 2019.

\bibitem{martin2001database}
David Martin, Charless Fowlkes, Doron Tal, and Jitendra Malik.
\newblock A database of human segmented natural images and its application to
  evaluating segmentation algorithms and measuring ecological statistics.
\newblock In {\em Proceedings of IEEE International Conference on Computer
  Vision}, volume~2, pages 416--423. IEEE, 2001.

\bibitem{nam2016holistic}
Seonghyeon Nam, Youngbae Hwang, Yasuyuki Matsushita, and Seon~Joo Kim.
\newblock A holistic approach to cross-channel image noise modeling and its
  application to image denoising.
\newblock In {\em Proceedings of the IEEE conference on computer vision and
  pattern recognition}, pages 1683--1691, 2016.

\bibitem{neshatavar2022cvf}
Reyhaneh Neshatavar, Mohsen Yavartanoo, Sanghyun Son, and Kyoung~Mu Lee.
\newblock Cvf-sid: Cyclic multi-variate function for self-supervised image
  denoising by disentangling noise from image.
\newblock In {\em Proceedings of the IEEE/CVF Conference on Computer Vision and
  Pattern Recognition}, pages 17583--17591, 2022.

\bibitem{pang2021recorrupted}
Tongyao Pang, Huan Zheng, Yuhui Quan, and Hui Ji.
\newblock Recorrupted-to-recorrupted: unsupervised deep learning for image
  denoising.
\newblock In {\em Proceedings of the IEEE/CVF conference on computer vision and
  pattern recognition}, pages 2043--2052, 2021.

\bibitem{quan2020self2self}
Yuhui Quan, Mingqin Chen, Tongyao Pang, and Hui Ji.
\newblock Self2self with dropout: Learning self-supervised denoising from
  single image.
\newblock In {\em Proceedings of the IEEE/CVF conference on computer vision and
  pattern recognition}, pages 1890--1898, 2020.

\bibitem{tico2008multi}
Marius Tico.
\newblock Multi-frame image denoising and stabilization.
\newblock In {\em European Signal Processing Conference}, pages 1--4. IEEE,
  2008.

\bibitem{ulyanov2018deep}
Dmitry Ulyanov, Andrea Vedaldi, and Victor Lempitsky.
\newblock Deep image prior.
\newblock In {\em Proceedings of the IEEE/CVF conference on computer vision and
  pattern recognition}, pages 9446--9454, 2018.

\bibitem{wang2021unsupervised}
Longguang Wang, Yingqian Wang, Xiaoyu Dong, Qingyu Xu, Jungang Yang, Wei An,
  and Yulan Guo.
\newblock Unsupervised degradation representation learning for blind
  super-resolution.
\newblock In {\em Proceedings of the IEEE/CVF Conference on Computer Vision and
  Pattern Recognition}, pages 10581--10590, 2021.

\bibitem{wang2016d3}
Zhangyang Wang, Ding Liu, Shiyu Chang, Qing Ling, Yingzhen Yang, and Thomas~S
  Huang.
\newblock D3: Deep dual-domain based fast restoration of jpeg-compressed
  images.
\newblock In {\em Proceedings of the IEEE/CVF conference on computer vision and
  pattern recognition}, pages 2764--2772, 2016.

\bibitem{xu2020noisy}
Jun Xu, Yuan Huang, Ming-Ming Cheng, Li Liu, Fan Zhu, Zhou Xu, and Ling Shao.
\newblock Noisy-as-clean: Learning self-supervised denoising from corrupted
  image.
\newblock {\em IEEE Transactions on Image Processing}, 29:9316--9329, 2020.

\bibitem{xu2018real}
Jun Xu, Hui Li, Zhetong Liang, David Zhang, and Lei Zhang.
\newblock Real-world noisy image denoising: A new benchmark.
\newblock {\em arXiv preprint arXiv:1804.02603}, 2018.

\bibitem{linr}
Wentian Xu and Jianbo Jiao.
\newblock Revisiting implicit neural representations in low-level vision.
\newblock In {\em International Conference on Learning Representations
  Workshop}, 2023.

\bibitem{yu2019free}
Jiahui Yu, Zhe Lin, Jimei Yang, Xiaohui Shen, Xin Lu, and Thomas~S Huang.
\newblock Free-form image inpainting with gated convolution.
\newblock In {\em Proceedings of the IEEE/CVF International Conference on
  Computer Vision}, pages 4471--4480, 2019.

\bibitem{zamir2022restormer}
Syed~Waqas Zamir, Aditya Arora, Salman Khan, Munawar Hayat, Fahad~Shahbaz Khan,
  and Ming-Hsuan Yang.
\newblock Restormer: Efficient transformer for high-resolution image
  restoration.
\newblock In {\em Proceedings of the IEEE/CVF conference on computer vision and
  pattern recognition}, pages 5728--5739, 2022.

\bibitem{zhang2023mm}
Dan Zhang, Fangfang Zhou, Yuwen Jiang, and Zhengming Fu.
\newblock Mm-bsn: Self-supervised image denoising for real-world with
  multi-mask based on blind-spot network.
\newblock In {\em Proceedings of the IEEE/CVF Conference on Computer Vision and
  Pattern Recognition}, pages 4188--4197, 2023.

\bibitem{dncnn}
Kai Zhang, Wangmeng Zuo, Yunjin Chen, Deyu Meng, and Lei Zhang.
\newblock Beyond a gaussian denoiser: Residual learning of deep cnn for image
  denoising.
\newblock {\em IEEE transactions on image processing}, 26(7):3142--3155, 2017.

\bibitem{zhang2011color}
Lei Zhang, Xiaolin Wu, Antoni Buades, and Xin Li.
\newblock Color demosaicking by local directional interpolation and nonlocal
  adaptive thresholding.
\newblock {\em Journal of Electronic imaging}, 20(2):023016, 2011.

\bibitem{zhang2022idr}
Yi Zhang, Dasong Li, Ka~Lung Law, Xiaogang Wang, Hongwei Qin, and Hongsheng Li.
\newblock Idr: Self-supervised image denoising via iterative data refinement.
\newblock In {\em Proceedings of the IEEE/CVF Conference on Computer Vision and
  Pattern Recognition}, pages 2098--2107, 2022.

\bibitem{zhang2018image}
Yulun Zhang, Kunpeng Li, Kai Li, Lichen Wang, Bineng Zhong, and Yun Fu.
\newblock Image super-resolution using very deep residual channel attention
  networks.
\newblock In {\em Proceedings of the European conference on computer vision},
  pages 286--301, 2018.

\bibitem{zhou2020awgn}
Yuqian Zhou, Jianbo Jiao, Haibin Huang, Yang Wang, Jue Wang, Honghui Shi, and
  Thomas Huang.
\newblock When awgn-based denoiser meets real noises.
\newblock In {\em Proceedings of the AAAI Conference on Artificial
  Intelligence}, volume~34, pages 13074--13081, 2020.

\end{thebibliography}
}

\newpage
\clearpage
% \begin{appendices}

\appendix

\noindent {\Huge Supplementary}

\section{Introduction}
This document provides supplementary materials for the main paper. Specifically, 
Section~\ref{sec:expdetails} presents thorough details of the model training used in our experiments followed by an ablation study of parameters. 
The supplemental denoising quantitative evaluation is presented in Section \ref{supp eva}.
Section \ref{sec:apps} explores more low-level applications, including image super-resolution and inpainting, with comparison to the proposed MeD. Section~\ref{datasets} and Section~\ref{noise} provide further details regarding the datasets used in our research and the methods for synthesising noise and downscale corruption. 
Finally, we present more qualitative results, with comparison to other methods, in Section \ref{result}.

\section{Denoising Training Settings}
\label{sec:expdetails}
For methods that use \textit{Swin-Tx} model, \eg N2C~\cite{liu2021Swin}, MeD and N2N~\cite{lehtinen2018noise2noise}, we use the same set of hyperparameters for training. 
Prior to the formal experiment, we conducted some pilot experiments to test and select the final choice of hyperparameters on N2C. 
Following \cite{dncnn, liu2021Swin}, we use $48 \times 48$ random crops from DIV2K images. The training process is performed using a mini-batch size of 8 and undergoes a total of 500K iterations. We use Adam with $\beta_1$ = 0.9 and $\beta_2$ = 0.99 and learning rate of $10^{-4}$, which decays every 100K with decay ratio 0.5. 

Since the model is not the focus of this work, we use a simple 2-layer \textit{Swin-Tx} for a fair comparison with other non-Transformer models, and is less likely to overfit the synthesised training noise distribution. 

\begin{figure*}[t]
\newcommand\M{\includegraphics[width=0.22 \textwidth]} 
\hspace{-0.4cm}

\begin{center}

	\begin{tabular}{c@{\extracolsep{0.2em}}c@{\extracolsep{0.2em}}c@{\extracolsep{0.2em}}c}
        %  左下右上
      \includegraphics[width=0.486 \textwidth,     trim=0 23 0 0,clip]{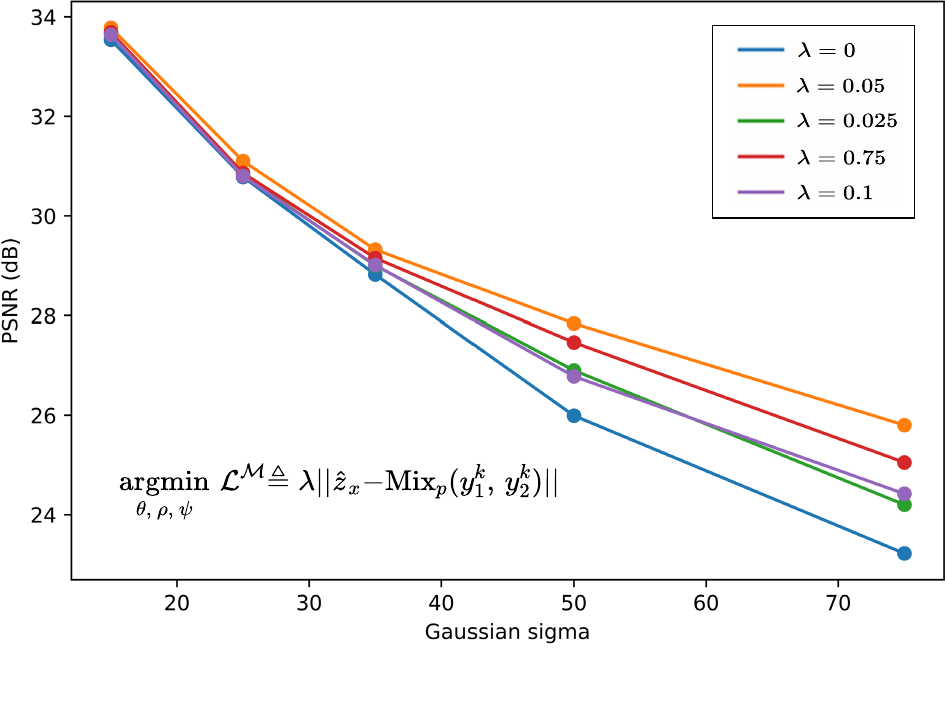}  \qquad   &
        \includegraphics[width=0.505 \textwidth, trim=0 0 0 0,clip]{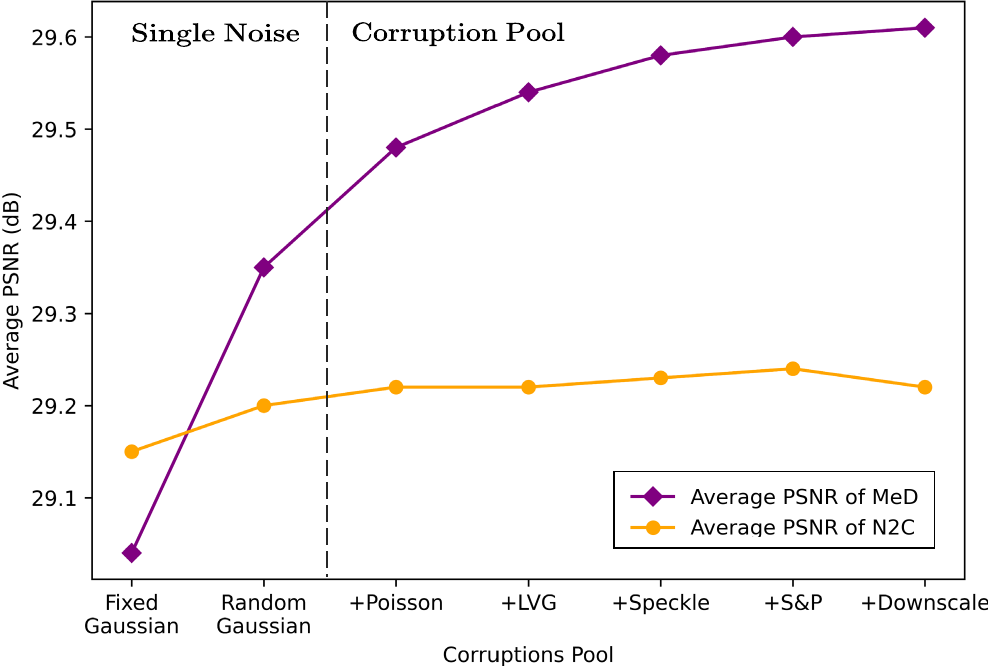}    \\
 		(a) Analysis on $\lambda$ for Bernoulli Manifold Mixture ~ \qquad   & (b) Analysis on the training Corruptions Pool \\

 	\end{tabular}

		\vskip 0.25cm
		\caption{ \textbf{Ablation experiments.} All models are tested with Gaussian noise removal on the CBSD68 dataset, unless otherwise specified. (a) MeD with Bernoulli Manifold Mixture Loss achieves the best performance at  $\lambda=0.05$ (the orange/top curve). 
        (b) The performance of N2C and MeD is assessed while varying trained input corruptions. The findings suggest that MeD benefits much better from a diverse training corruption pool (the ``+" sign in the horizontal axis indicates the further inclusion of a corruption in the corruption pool.).
		}
		\label{fig:ablation}
	\end{center}
	% \vskip -0.6cm % 0.4
\end{figure*}

The influence of the size of the \textit{Corruption Pool} on the performance of MeD and N2C~\cite{liu2021Swin} is demonstrated in Figure \ref{fig:ablation} (b). The experiments are started from only fixed Gaussian noise, then train on Gaussian noise with random sigma values. 
Finally, we expand the corruption pool from only Gaussian noise to more noise types and even with different types of down-scale and inpainting mask operations. 

\vspace{-1pt}
\begin{table}[h]
	% \caption{ Performance of on synthetic(NP) with Varying Hyperparameters.}
	\caption{Analysis of hyper-parameters for the loss terms.}
 % \vspace{0.2cm}
	\small
	\centering
	% \resizebox{\linewidth}{!}{
	\setlength{\tabcolsep}{4pt}
 \resizebox{0.48\textwidth}{!}{
    \begin{tabular}{c|ccc}
     \toprule
   Loss Hyperparameters for & \multicolumn{3}{c}{Gaussian Noise}\\
    $[ \mathcal L^{\mathcal X} $,\ \ \ 
    $ \mathcal L^{\mathcal N} $,\ \ \ 
    $ \mathcal L^{\mathcal C} $,\ \ \ 
    $ \mathcal L^{\mathcal M} ]$ & $\hat\sigma=25$ & $\hat\sigma=50$ & $\hat\sigma=75$\\
    \midrule
    $[1.0,\ \ 0.5,\  \ 0.5,\  \ 0.025]$ & 31.18/  0.8839 & 27.68/  0.7765 &  25.74/  0.7172  \\
    $[0.5,\ \ 1.0,\ \  0.5,\  \ 0.025]$ & 31.04/  0.8813 & 27.87/  0.7788 & 25.86/  0.7190  \\
    $[0.5,\ \ 0.5,\  \ 1.0,\  \ 0.025]$ & 30.85/  0.8752 & 27.96/  0.7794 & 25.92/  0.7199  \\
    $[1.0,\ \ 1.0,\  \ 1.0,\ \  0.025]$ & \textbf{31.31/  0.8876} & \textbf{28.05/  0.7810} & \textbf{26.01/  0.7216}\\
    $[1.0,\ \ 1.0,\  \ 1.0,\ \  0.050]$ & 31.29/  0.8870 & 27.58/ 0.7659  &  24.29/ 0.6931 \\
    $[1.0,\ \ 1.0,\  \ 1.0,\ \  0.000]$ & 31.20/  0.8832  & 26.24/  0.7391 &  23.78/ 0.6517 \\
     \bottomrule
    \end{tabular}%
    }
    
% \hspace{-1cm}
\label{tab:hyper}
\end{table}
% \vspace{-2mm}

% ----------------- Figure

% er methods. More experiments and details on AWGN noise removal can be found in the supplementary m

\begin{table*}[th]
% \hspace{0.8cm}
	\caption{Supplementary quantitative comparison of different methods on CBSD68 dataset \cite{martin2001database} for synthetic Gaussian noise. The experiments were conducted on fixed and random variance respectively. The best results are highlighted in \textbf{bold}, while the second best is \underline{underlined}.}
	\small
	\centering
	% \resizebox{\linewidth}{!}{
	\setlength{\tabcolsep}{4pt}

    \begin{tabular}{l|c|cc|ccc|cc}
     \toprule
     
     Training  %\multirow{2}{*}{} 
     & Test  & %\multicolumn{2}{c|}{Single-image-based} & 
     \multicolumn{2}{c|}{Noisy/ Clean}& 
     \multicolumn{3}{c|}{Noisy/ Noisy}  &\multicolumn{2}{c}{Invariant Feature} 
      \\
    %  \midrule  N2V \cite{krull2019noise2void} &

     Schema & $\hat \sigma$ &  N2C \cite{liu2021Swin}  & DBD$_4$ \cite{godard2018deep} & N2N \cite{lehtinen2018noise2noise} & N2S \cite{batson2019noise2self} &  R2R \cite{pang2021recorrupted} & LIR \cite{du2020learning} & MeD   \\  
    
     \midrule
      \multirow{4}{*}{\makecell[c]{Gaussian \\$\sigma =15$}   }  
% sigma|    Swinx   |    N2N     |     N2S      |      R2R    |     LIR     |    MeDIA  |
& 15  & \underline{33.21}/ 0.9194 &  33.17/ \underline{0.9179}&  33.16/ 0.9175 & 32.88/ 0.9099& 31.19/ 0.8752 & 29.29/ 0.8118& \textbf{33.20/ 0.9181}\\  
% sigma|    Swinx   |    N2N     |     N2S      |      R2R    |     LIR     |    MeDIA  |
& 25  & \underline{28.04/ 0.7588} & 26.36/ 0.7473 & 26.56/ 0.7521 &26.27/ 0.7456& 23.51/ 0.5529& 21.80/ 0.4802&  \textbf{29.80/0.8401} \\ % & 31.10/ 0.86 \\ 
% sigma|    Swinx   |    N2N     |     N2S      |      R2R    |     LIR     |    MeDIA  |
& 50 & \underline{19.89/ 0.3755} & 19.86/ 0.3741&  19.68/ 0.3672& 16.13/ 0.1978 & 16.09/ 0.2080& 11.12/ 0.1217 &     \textbf{23.51/ 0.5529 }\\%& 25.45/ 0.59\\ 
% sigma|    Swinx   |    N2N     |     N2S      |      R2R    |     LIR     |    MeDIA  |
& 75 &\underline{17.16/ 0.2562} & 16.62/ 0.2314 & 14.59/ 0.1561 & 14.99/ 0.1642& 14.09/ 0.1280& 11.10/ 0.1042 & \textbf{20.47/ 0.3870}  \\ %& 22.3/ 0.44 \\ 

% ---------------------- Color ------------------------------
   \midrule

      \multirow{4}{*}{\makecell[c]{Gaussian \\$\sigma =50$}   }  
% sigma|    Swinx   |    N2N     |     N2S      |      R2R    |     LIR     |    MeDIA  |
& 15  &  30.69/ 0.8497   & 30.70/ 0.8478 & \underline{30.80/ 0.8619}  &29.87/ 0.8267 & 28.15/ 0.7872 & 29.44/ 0.8011 & \textbf{30.88/ 0.8799}  \\  
% sigma|    Swinx   |    N2N     |     N2S      |      R2R    |     LIR     |    MeDIA  |
& 25  &   \underline{29.81}/ 0.8140  & 29.63/ \underline{0.8182} & 29.54/ 0.8256  & 29.54/ 0.8256 &  28.89/ 0.8099 & 28.95/ 0.7967 &    \textbf{30.19/ 0.8218} \\ % & 31.10/ 0.86 \\ 
% sigma|    Swinx   |    N2N     |     N2S      |      R2R    |     LIR     |    MeDIA  |
& 50 & \underline{28.56}/ 0.7721 & 28.32/ \underline{0.7765} & 28.30/ 0.7490 &  28.19/ 0.7802 & 27.80/ 0.7547   &  28.02/ 0.7682 & \textbf{28.56/ 0.7835} \\ %& 25.45/ 0.59\\ 
% sigma|    Swinx   |    N2N     |     N2S      |      R2R    |     LIR     |    MeDIA  |
& 75 &  \underline{22.60}/ 0.5877 & 22.47/ \underline{0.6042} & 22.45/ 0.5759  & 21.69/ 0.5433& 21.42/ 0.5881 & 20.25/ 0.5368 &  \textbf{25.63/  0.7372} \\ %& 22.3/ 0.44 \\ 

     \bottomrule
    \end{tabular}
\label{sup:gauss}
\end{table*}

\subsection{Hyperparameter Analysis}

Prior to finalising the training procedure, we conducted experiments to analyse the impact of different hyperparameters associated with the loss terms in our model. Specifically, we tested varying the weighting factors $ \mathcal L^{\mathcal X} $, $ \mathcal L^{\mathcal N} $, $ \mathcal L^{\mathcal C} $ and $ \lambda$ for the \textit{Noise Reconstruction loss}, \textit{Scene Reconstruction loss}, \textit{Cross Compose loss}, and \textit{Mix Scene reconstruction loss}, respectively. The analysis is shown in Figure \ref{fig:ablation} (a) and Table \ref{tab:hyper}

Firstly, we conducted the experiment for analysing the value $\lambda$ in Figure \ref{fig:ablation} (a). The orange (top) curve represents the performance of the optimal choice $\lambda=0.05$.

$ \mathcal L^{\mathcal X} $, $ \mathcal L^{\mathcal N} $ and $ \mathcal L^{\mathcal C} $ are tested from 0.5 to 1, with the best results obtained at $ \mathcal L^{\mathcal X} = 1$, $ \mathcal L^{\mathcal N} = 1$ and $ \mathcal L^{\mathcal C} = 1$.

Based on these experiments, we selected hyperparameters of $\mathcal L^{\mathcal X} = 1$, $ \mathcal L^{\mathcal N} = 1$, $ \mathcal L^{\mathcal C} = 1$, and $\lambda=0.025$ for all denoising training in our work. This provides an optimal balance between the different objectives.

%%%%%%%%%% 03

\section{Additional Denoising Evaluation}
\label{supp eva}
In this section, we present additional quantitative evaluations of our denoising results, which were not included in the main paper. 
\label{AWGN} The results are in Table \ref{sup:gauss} with additive Gaussian Noise, supplement to Table 1 in the main paper.

\subsection{Generalisation on Unseen Noise Removal}
To evaluate the generalisation ability of trained models on unseen noise removal, we conducted an additional experiment with more methods using only a single noisy image in Table \ref{tab:unseenmore}. The results show that our method achieves comparable performance to state-of-the-art methods specially designed for Gaussian noise removal, and outperforms all compared methods on other noise types, while training only on the Gaussian noise. This further highlights the generalisation ability of our approach in handling unseen and unfamiliar noise distributions.

\begin{table*}[t]
\vspace{-2mm}
	\caption{Performance comparison of single-view approaches and Ours training on Gaussian noise and testing on various noise types. }
 % \vspace{0.2cm}
	\small
	\centering
	% \resizebox{\linewidth}{!}{
	% \setlength{\tabcolsep}{4pt}
	%\renewcommand{\arraystretch}{0.98}
 
 \resizebox{1\textwidth}{!}{
    \begin{tabular}{l|ccccc|c}
     \toprule
   Noise Type & DIP~\cite{ulyanov2018deep} & NAC~\cite{xu2020noisy} & S2S~\cite{quan2020self2self} & IDR~\cite{zhang2022idr}  & Restormer~\cite{zamir2022restormer} & MeD (Ours)\\
    \midrule
    
     Gaussian, $\hat \sigma \in[25, 75]$  & 25.62/ 0.7017 & 27.13/ 0.7391  & 27.71/ 0.7622 & {28.52/ 0.8061} &  \textbf{29.10/ 0.8250} & 28.45/ 0.8057 \\
     Speckle, $ \hat v\in [25, 50]$ & 30.14/ 0.8574  & 31.55/ 0.8859  & 31.83/ 0.8980 &28.62/ 0.8763 &   30.12/ 0.8557 & \textbf{33.48/ 0.9115}\\ 
     S\&P, $\hat r \in [0.3, 0.5]$  & 28.62/ 0.7957  &  29.89/ 0.8741 & 30.57/ 0.9053 & 27.26/ 0.7544 &  23.09/ 0.6381 & \textbf{30.84/ 0.9135}\\
     \midrule
     Average & 28.13/ 0.7849 & 29.52/ 0.8330 & 30.04/ 0.8552 & 28.13/ 0.8123 &  27.44/ 0.7729 & \textbf{30.92/ 0.8770} \\
     % SIDD &&&&&& have not enough time for doing it\\
 
     \bottomrule
    \end{tabular}%
    }
    
\hspace{-0.3cm}
	\label{tab:unseenmore}
\end{table*}
\vspace{-3mm}

 \definecolor{Gray}{rgb}{0.8,0.8,0.8}
\newcommand{\cmark}{\ding{51}}%
\newcommand{\xmark}{\ding{55}}%
\newcommand{\g}[1]{ \colorbox{Gray}{#1}}

        \begin{table*}[h]
            \centering
            \caption{Train and test both on real-world datasets (PSNR/SSIM).}
            \label{tab:realreal}
            % \setlength{\tabcolsep}{4pt}

            % \resizebox{\textwidth}{!}{
            \begin{tabular}{lc|cc|cc}
                \toprule
                Method & MAC (G) &  Supervised & Trained with & SIDD~\cite{SIDD_2018_CVPR} &  PolyU~\cite{xu2018real}   \\
                \midrule
                % \rowcolor{Gray}
                \textcolor{Gray}{Restormer~\cite{zamir2022restormer}} & \textcolor{Gray}{140.99} & \textcolor{Gray}{\cmark}  &    \textcolor{Gray}{Real (SIDD)}   &  \textcolor{Gray}{40.06/ 0.9601} & \textcolor{Gray}{36.38/ 0.9588}              \\
                % \rowcolor{Gray}
                \textcolor{Gray}{NAFNet~\cite{chen2022simple}} & \textcolor{Gray}{63.6} & \textcolor{Gray}{\cmark}     &   \textcolor{Gray}{Real (SIDD)}   & \textcolor{Gray}{\textbf{40.31}/ 0.9667} & \textcolor{Gray}{27.36/ 0.9225}               \\
                % MAXIM [\textit{Tu et al.} 2022] & &\cmark      &    SIDD  & 39.98/ 0.9591 & 35.79/ 0.9519               \\
                \midrule
                N2N~\cite{liu2021Swin}  & 26.18 & \xmark  &    Real (SIDD)      &  32.82/ 0.7297 & 36.22/ 0.9679           \\
                N2S~\cite{batson2019noise2self} & 26.18 & \xmark  &    Real (SIDD)      & 30.98/ 0.6018  & 36.41/ 0.9721           \\
                CVF-SID~\cite{neshatavar2022cvf} & 77.86 & \xmark & Real (SIDD) & 34.71/ 0.9179 &  33.00/ 0.8768 \\ 
                MM-BSN~\cite{zhang2023mm}  & 339.46 & \xmark & Real (SIDD) & 37.37/ 0.9362 & 35.40/ 0.9484 \\
                \midrule
                % MeD (Ours) & 26.18 & \xmark & Real (SIDD) &  & \\
                MeD (Ours) & 26.18 & \xmark & Synthetic (NP) & 35.81/ 0.8278 & 38.65/ 0.9855\\
                MeD (Ours) & 26.18 & \xmark & NP + SIDD & \textbf{37.52}/ \textbf{0.9434} & \textbf{38.91/ 0.9894} \\

                \bottomrule
            \end{tabular}%
            % }
        \end{table*}

 \subsection{Further Analysis on Real-world Generalisation}
 \label{sec:realreal}
In the main paper, we have shown that our method generalises well to real-world scenarios when trained only on synthetic data. Moreover, here we conduct experiments on training with a real-world dataset (SIDD without GT), and report the test results on SIDD and PolyU in Table \ref{tab:realreal}.

 \noindent \textbf{Analysis: }
Comparing Table \ref{tab:realreal} and Table 4 in the main paper, it shows that all methods have significant improvement in performance on SIDD after training on SIDD, but little improvement on PolyU.

Considering that collecting real data is expensive and sometimes infeasible compared to synthetic data and, as our following experiments show, generalising to new real datasets (real-to-real) is another issue (since the noise distributions are different), the model trained on synthetic noise data is more feasible and practical.

% 04

\begin{figure*}[htpb]
% \footnotesize

% \hspace{-0.40cm}

\centering
\hspace{0.1cm}

\begin{tabular}{c@{\extracolsep{0em}}
c@{\extracolsep{0.05em}}
c@{\extracolsep{0.05em}}c}
\centering
\large
		\includegraphics[width=0.241 \textwidth]{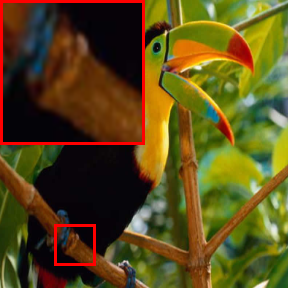}~
		&\includegraphics[width=0.241 \textwidth]{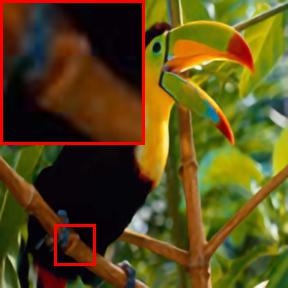}~  
        & \includegraphics[width=0.241 \textwidth]{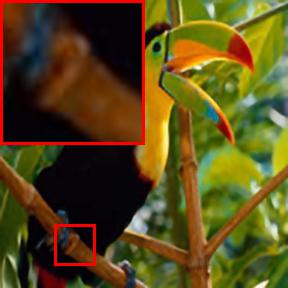}~
		&\includegraphics[width=0.241 \textwidth]{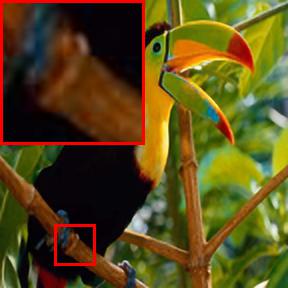}~   \\
     
	Set5 ``Bird" \cite{bevilacqua2012low}~  & RCAN \cite{zhang2018image}~ &  DASR \cite{wang2021unsupervised}~ & MeD (Ours)  \\
   
          PSNR/SSIM ~ & 34.89/ 0.9512 ~ & 34.42/ 0.9364~ & \textbf{36.66/ 0.9747}  \\

\end{tabular}

	\caption{Visual comparison of image super-resolution (×3) methods on Set5 ``Bird" \cite{bevilacqua2012low} images. }
	
	% \vspace{-0.1cm}
 \label{fig:sr1}
\end{figure*}

\begin{figure*}[!htpb]
% \footnotesize
% \captionsetup{font=small}
% \hspace{-0.20cm}
\newcommand\M{\includegraphics[width=0.241 \textwidth]}

\centering

\begin{tabular}{c@{\extracolsep{0em}}c@{\extracolsep{0em}}c@{\extracolsep{0em}}c}
\centering
\large

		\includegraphics[width=0.241 \textwidth]{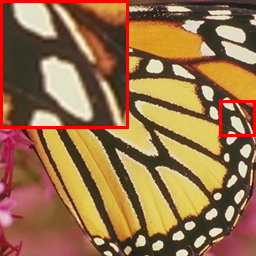}~
		&\includegraphics[width=0.241 \textwidth]{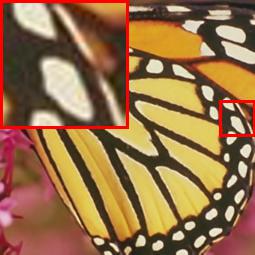}~  
       &\includegraphics[width=0.241 \textwidth]{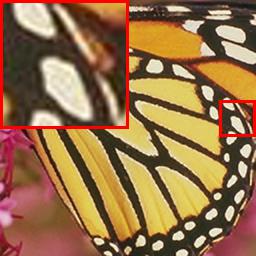}~
		&\includegraphics[width=0.241 \textwidth]{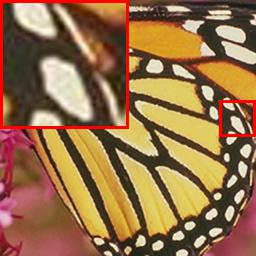}~   \\
         Set 5 ``Butterfly" \cite{bevilacqua2012low}  &  RCAN \cite{zhang2018image} &  DASR \cite{wang2021unsupervised} & MeD (Ours)   \\
         PSNR/SSIM & 30.91/ 0.9459 & 30.82/ 0.9527 & \textbf{31.12/ 0.9636}\\

	\end{tabular}
    \vspace{-0.0cm}
	\caption{Visual comparison of image super-resolution (×4) methods on Set5 ``Butterfly" \cite{bevilacqua2012low} images. 
	}
	
	% \vspace{-0.1cm}
 \label{fig:sr2}
\end{figure*}

%%%    !t
\begin{figure*}[!htpb]
% \footnotesize
% \captionsetup{font=small}
% \hspace{-0.20cm}
\newcommand\M{\includegraphics[width=0.241 \textwidth]}
\centering
\begin{tabular}{c@{\extracolsep{0em}}c@{\extracolsep{0em}}c@{\extracolsep{0em}}c}
\centering
% \Large

		\includegraphics[width=0.241 \textwidth]{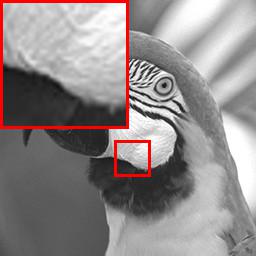}~
		&\includegraphics[width=0.241 \textwidth]{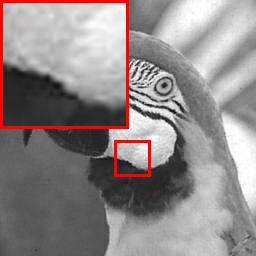}~  
        &\includegraphics[width=0.241 \textwidth]{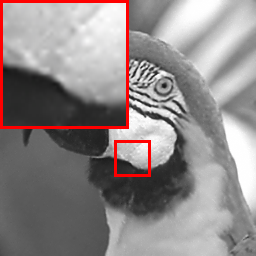}~
		&\includegraphics[width=0.241 \textwidth]{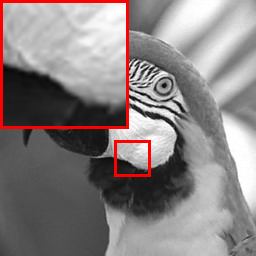}~   \\
         Set 11 ``Parrots" \cite{bevilacqua2012low}  &  DIP \cite{ulyanov2018deep} &  S2S \cite{quan2020self2self} & MeD (Ours)  \\
         PSNR/SSIM & 31.94/ 0.9479 & 33.91/ 0.9224 & \textbf{34.01/ 0.9507}\\

        \includegraphics[width=0.241 \textwidth]{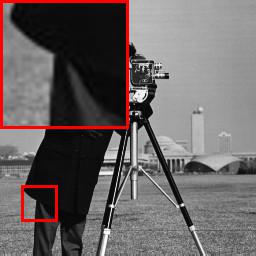}~
		&\includegraphics[width=0.241 \textwidth]{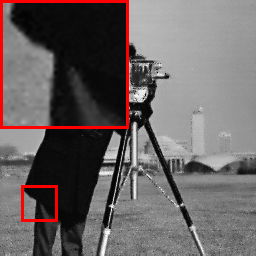}~  
        &\includegraphics[width=0.241 \textwidth]{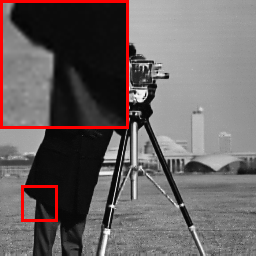}~
		&\includegraphics[width=0.241 \textwidth]{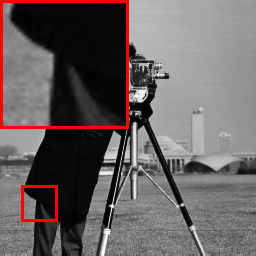}~   \\
  
         Set 11 ``Cameraman" \cite{bevilacqua2012low}  & DIP \cite{ulyanov2018deep} &  S2S \cite{quan2020self2self} & MeD (Ours)  \\
         PSNR/SSIM & 30.97/ 0.9778 & 33.37/ 0.9355 & \textbf{34.99/ 0.9478}\\

	\end{tabular}
    % \vspace{-0.0cm}
	\caption{Visual comparison of image Inpainting methods on Set11 \cite{kulkarni2016reconnet} images. 
	}
	
	% \vspace{-0.1cm}
 \label{fig:im}
\end{figure*}

\section{More Application Exploration}
\label{sec:apps}
\subsection{Experiment on Image Super-resolution}
\label{sec:sr}

Figure \ref{fig:sr1} and Figure \ref{fig:sr2} show the qualitative results of our method for ×3 and ×4 super-resolution on Set5 dataset \cite{bevilacqua2012low}, compared with RCAN \cite{zhang2018image} and DASR \cite{wang2021unsupervised}. It shows that our method achieves better performance than these methods, by using a corruption pool that contains both noise and down-scaling process.

% 05

% % \include{tables/im}
\subsection{Experiment on Image Inpainting}
\label{sec:in}
Evaluation is performed on Set11 \cite{kulkarni2016reconnet}. 
Please see Figure~\ref{fig:im} for two examples.  It can be seen although our method is not designed for inpainting, we can still achieve better performance than sate-of-the-art methods such as DIP~\cite{ulyanov2018deep} and S2S~\cite{quan2020self2self}.

% 06

\section{Datasets}
\label{datasets}
We used five different datasets to train and evaluate the denoising methods: DIV2K \cite{Agustsson_2017_CVPR_Workshops}, CBSD68 \cite{martin2001database}, SIDD \cite{SIDD_2018_CVPR}, CC \cite{nam2016holistic}, and PolyU \cite{xu2018real}.\\

 \noindent \textbf{DIV2K:}
DIVerse 2K resolution high-quality images \cite{Agustsson_2017_CVPR_Workshops} (DIV2K) contain 800 high-resolution images with a resolution of 2K or 4K.
To train our denoising method, we added different types and levels of noise to the DIV2K dataset.\\

 \noindent \textbf{CBSD68:}
 CBSD68 dataset \cite{martin2001database} contains 68 colourful images with various levels of synthesising noise. These images were obtained from a range of sources, including natural scenes and synthetic images. \\
 
 \noindent \textbf{SIDD:}
The Smartphone Image Denoising Dataset \cite{SIDD_2018_CVPR} (SIDD) is a large-scale real-world dataset containing ~ 24,000 images captured by smartphone cameras in ten scenes with varying lighting conditions. The ground truth images for the SIDD dataset are provided along with the noisy images in the dataset.\\

 \noindent \textbf{CC:}
Cross-Channel Image Noise Modeling \cite{nam2016holistic} (CC) is another real-world dataset which contains 11 static scenes captured by three different consumer cameras. For each scene, it contains one temporal image and the precomputed temporal mean and covariance matrix data.\\

 \noindent \textbf{PolyU:}
PolyU dataset \cite{xu2018real} is comprised of 40 different scenes captured by cameras. It contains the original image corrupted by realistic noise and the ground truth version which is obtained by averaging multiple exposures to remove the noise.\\
 
We also used Set5 dataset \cite{bevilacqua2012low} and Set11 \cite{kulkarni2016reconnet} to evaluate the super-resolution and image inpainting performances.\\

 \noindent \textbf{Set5 dataset:}
We use Set5 dataset \cite{bevilacqua2012low} for super-resolution task. 
The Set5 dataset \cite{bevilacqua2012low} consists of 5 high-quality images with different contents, including ``baby", ``bird", ``butterfly", ``head", and ``woman". Each of the images in the Set5 dataset has a magnifying factor of 2, 3, or 4, allowing us to evaluate the performance of our image super-resolution model across a range of magnification factors.\\

 \noindent \textbf{Set11 dataset:}
We compare our method (MeD) with DIP \cite{ulyanov2018deep} and S2S \cite{quan2020self2self}  on image inpainting tasks using the Set11 dataset \cite{kulkarni2016reconnet}, which contains 11 grayscale images. \\

% 07
\section{Synthesising Noisy Data and Downsampling Corruptions}
\label{noise}
 We utilise the Pillow library\footnote{https://pillow.readthedocs.io/en/stable/} in Python to synthesise noisy data and perform downsampling corruptions.
\subsection{Synthesising Noise}

To evaluate the performance of our proposed algorithm, we synthesised noisy images using several types of noise models, including Gaussian, Local Variance Gaussian, Poisson, Speckle, and Salt-and-Pepper.

\noindent \textbf{Remark:} The original pixel value at position $(i,j)$ in the image can be notated as $I(i, j)$ and the noisy pixel value can be notated as $I_{noisy}$. \\

 \noindent \textbf{Gaussian Noise:} 
Gaussian noise is a type of additive noise that is commonly found in digital images. 
It is modelled as a normal distribution with zero mean and a standard deviation $\sigma$. To synthesise Gaussian noise, we added Gaussian-distributed noise to the original image. Specifically, we added Gaussian noise with zero mean and standard deviation $\sigma$ to each pixel of the input image, where $\sigma$ was set to 10. The noisy pixel value $I_{noisy}(i,j)$ is given by:
\begin{equation}
    I_{noisy}(i,j) = I(i,j) + N(i,j),
\end{equation}
\noindent $N(i,j)$ is a random variable generated from a Gaussian distribution with zero mean and standard deviation $\sigma$.\\

 \noindent \textbf{Local Variance Gaussian Noise:}
Local Variance Gaussian noise is a variant of Gaussian noise that takes into account the local variance of the image. 
In this case, we added Gaussian noise with different standard deviations to different local regions of the input image to achieve more realistic noise patterns. Specifically, the standard deviation of Gaussian noise for each pixel was calculated based on the local variance of its neighbouring pixels. The noisy pixel value $I_{noisy}(i,j)$ is given by:
\begin{equation}
    I_{noisy}(i,j) = I(i,j) + N_{L}(i,j),
\end{equation}

\noindent where $N_{L}(i,j)$ is a random variable generated from a Gaussian distribution with zero mean and standard deviation $\sigma_L(i,j)$, which is calculated as:
\begin{equation}
    \sigma_L(i,j) = k * \sigma_{local}(i,j),
\end{equation}

\noindent where $\sigma_{local}(i,j)$ is the local variance of the image at pixel $(i, j)$, and $k$ is a scaling factor that determines the strength of the noise.\\

 \noindent \textbf{Poisson Noise:}
Poisson noise is a type of noise that arises from the random nature of photon arrival in digital images. It is modelled as a Poisson distribution with parameter $\lambda$. To synthesise Poisson noise, we first modelled the image as a Poisson process and then generated noisy pixels based on this model. Specifically, the noisy pixel value $I_{noisy}(i,j)$ is given by:
\begin{equation}
   \resizebox{0.87\columnwidth}{!}{ $I_{noisy}(i,j) = \min(255, \max(0, \text{Poisson}(\lambda(i,j)) + I(i,j)))$},
\end{equation}
where $I(i,j)$ is the original pixel value at position $(i,j)$, $\lambda(i,j)$ is the mean value of the Poisson distribution, and $\text{Poisson}(\lambda(i,j))$ is a random variable generated from a Poisson distribution with mean $\lambda(i,j)$.\\

 \noindent \textbf{Speckle Noise:}
Speckle noise is a type of multiplicative noise that is commonly found in ultrasound and radar images. It is modelled as a multiplicative noise with a uniform distribution between 0 and 1.
To synthesise speckle noise, we multiplied each pixel of the original image with a random value drawn from a uniform distribution between 0 and 1. Specifically, the noisy pixel value $I_{noisy}(i,j)$ is given by:
\begin{equation}
    I_{noisy}(i,j) = I(i,j) * U(0,1),
\end{equation}
\noindent where $U(0,1)$ is a random variable drawn from a uniform distribution between 0 and 1.\\

 \noindent \textbf{Salt-and-Pepper Noise:}
Salt-and-pepper noise is a type of impulse noise that occurs when some pixels in the image are replaced with the maximum or minimum pixel value. It is modelled as a random process that replaces a certain percentage of the pixels in the image with either the maximum or minimum pixel value. 
Specifically, the noisy pixel value $I_{noisy}(i,j)$ can be calculated as follows:
\begin{equation}
    I_{noisy}(i,j) = I(i,j) + S(i,j) - P(i,j),
\end{equation}

\noindent where $S(i,j)$ and $P(i,j)$ are random variables that model the presence of salt-and-pepper noise, respectively. They are defined as follows:
\begin{equation}
    S(i,j) = I_{max} * \text{Bernoulli}(p_s),
\end{equation}
\begin{equation}
    P(i,j) = I_{min} * \text{Bernoulli}(p_p),
\end{equation}
\noindent where $\text{Bernoulli}(p)$ is a random variable that takes the value 1 with probability p and the value 0 with probability $1-p$.

Note that $S(i,j)$ and $P(i,j)$ are only added to the pixel value $I(i,j)$ with the respective set probabilities $p_s$ and $p_p$. Therefore, the total percentage of pixels affected by salt and pepper noise is $p_s + p_p$.

\subsection{Downscale Corruption}
The down-scale corruption contains down-scale interpolation, including Bicubic, Lanczos, Bilinear and Hamming. We use OpenCV-Python for the down-scaling process.

% \onecolumn
\section{Additional Qualitative Results}
\label{result}
\begin{figure*}[h!]
    \centering  
    %  左下右上  , trim=10 0 150 200,clip
    \includegraphics[width=\textwidth ]{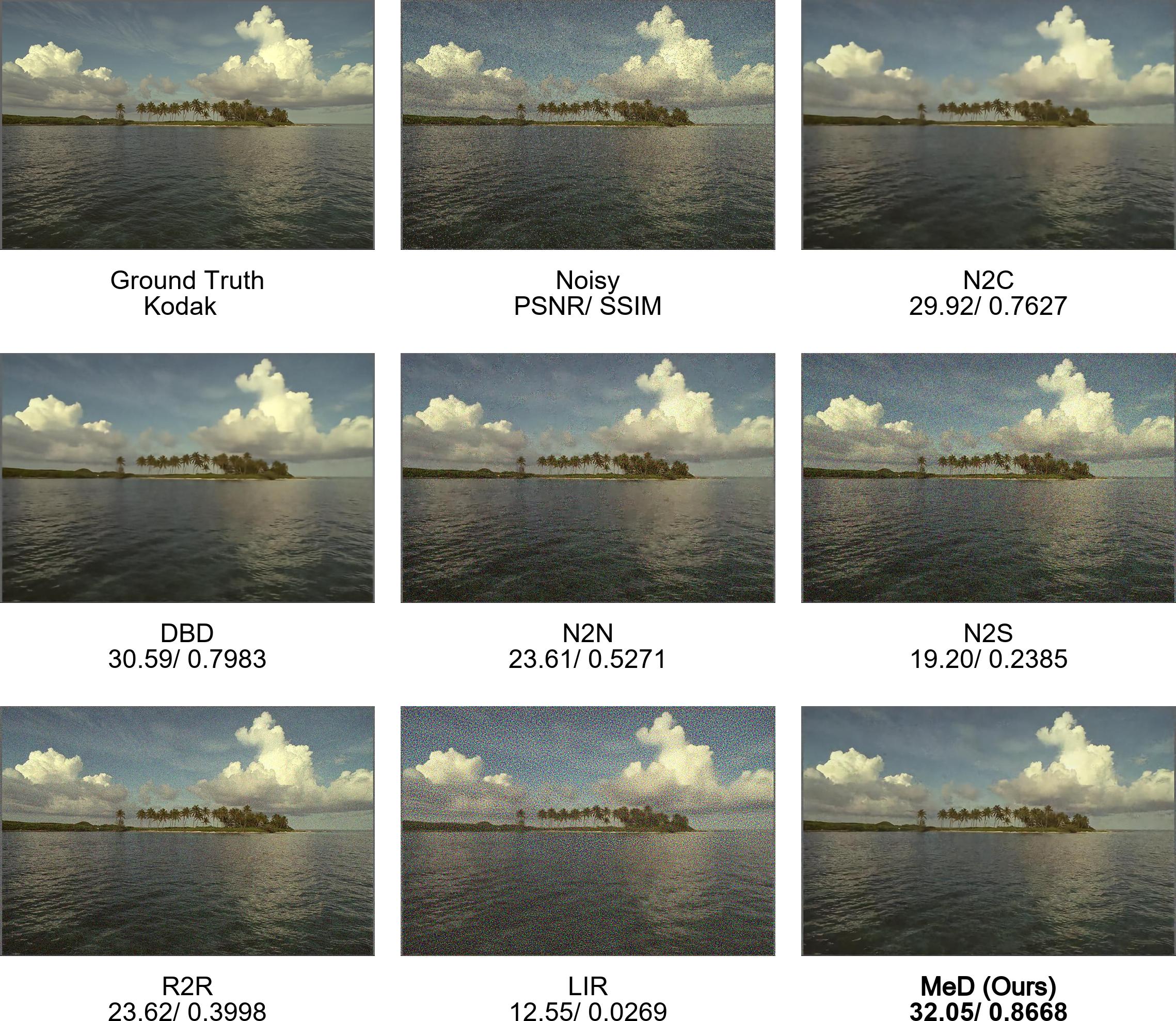}
	\caption{Visual comparison of image denoising methods on Kodak \cite{franzen1999kodak} images with Gaussian ($\sigma=25$) + local variance Gaussian noise.} 
 \label{fig:synthfirst}
\end{figure*}

\begin{figure*}[!hbtp]
    \centering
    \includegraphics[width=\textwidth]{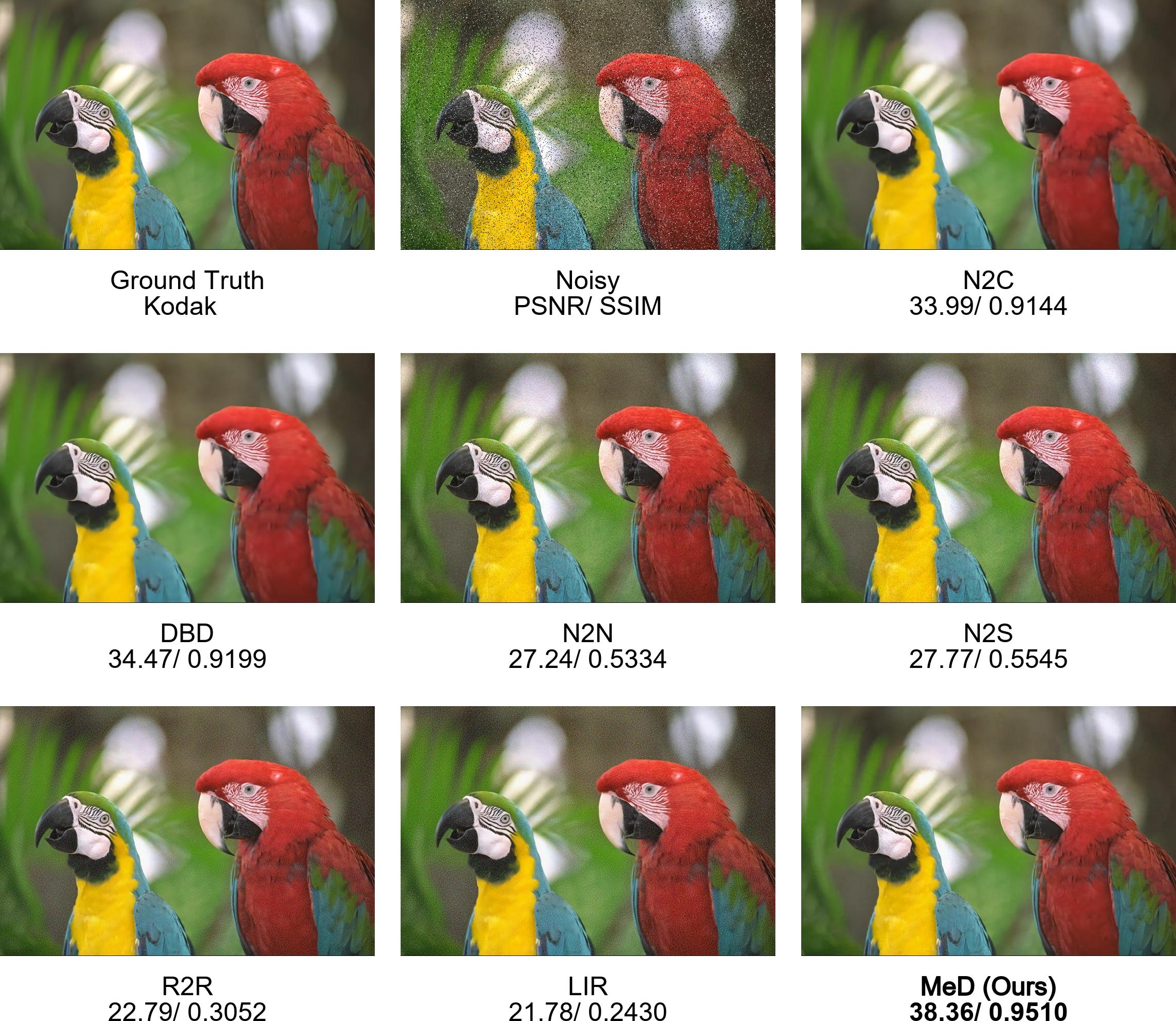}
	\caption{Visual comparison of image denoising methods on Kodak \cite{franzen1999kodak} images with local variance Gaussian + Poisson noise.} 
 \label{fig:synthlast}
\end{figure*}

\begin{figure*}[!hbtp]
    \centering
    \includegraphics[width=\textwidth]{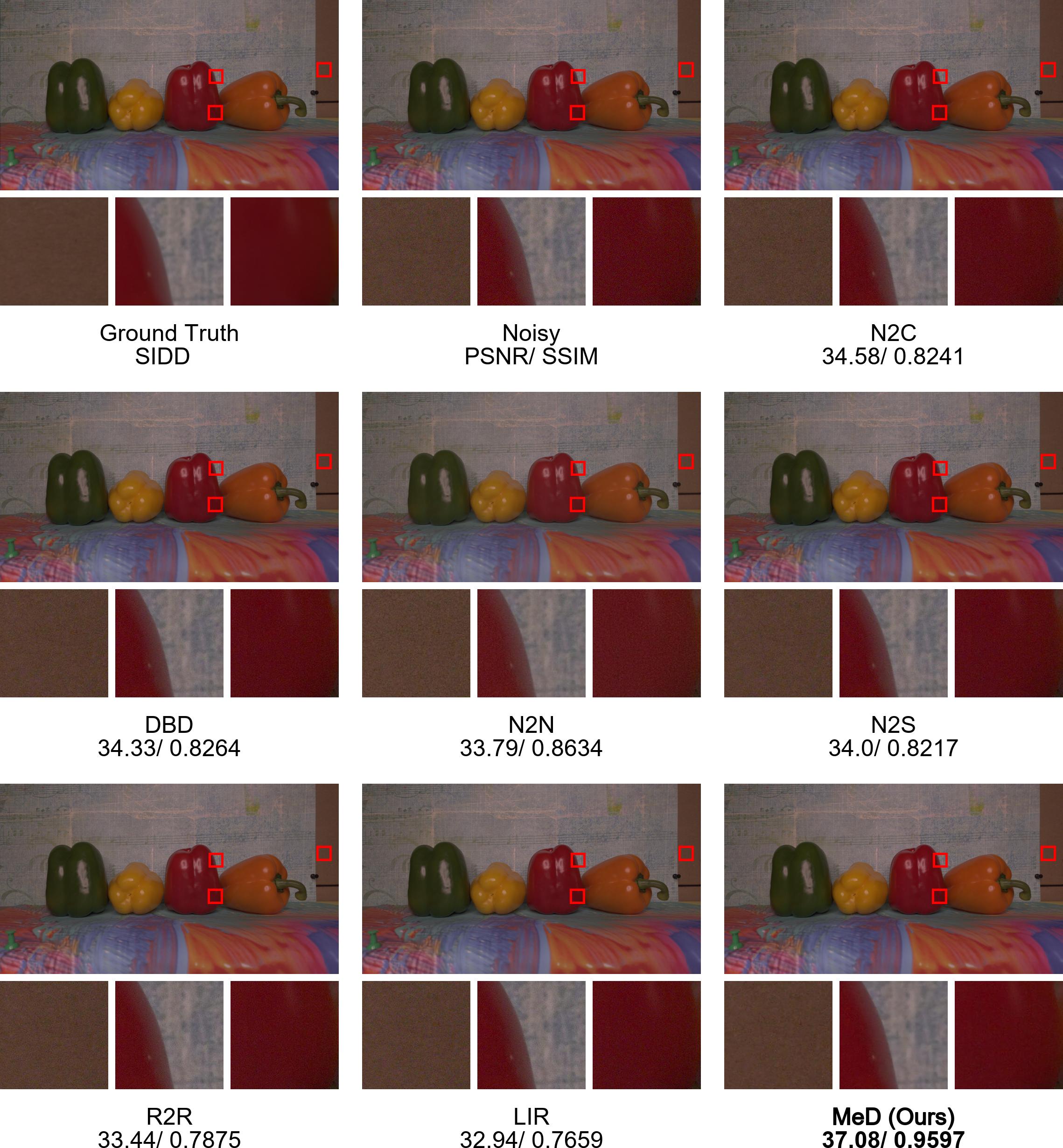}
	\caption{Visual comparison of image denoising methods on real noisy image dataset SIDD \cite{SIDD_2018_CVPR} example images with real noise.} 
 \label{fig:realfirst}
\end{figure*}

\begin{figure*}[!hbtp]
    \centering
    \includegraphics[width=\textwidth]{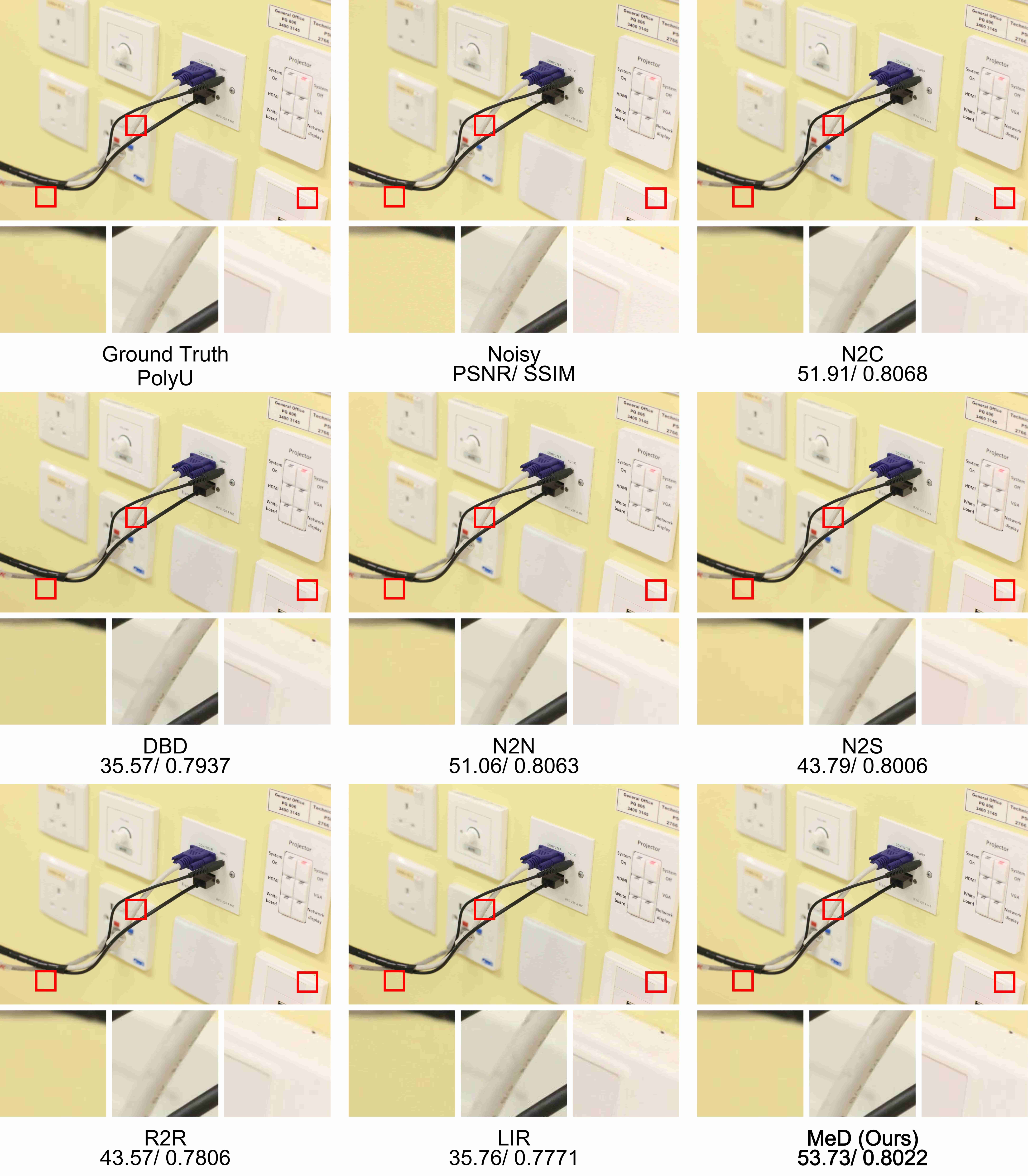}
	\caption{Visual comparison of image denoising methods on real noisy image dataset PolyU \cite{xu2018real} example images with real noise.} 

 \label{fig:reallast}
\end{figure*}

The following figures show the denoising comparison on both synthetic noise removal (Figure~\ref{fig:synthfirst} -- Figure~\ref{fig:synthlast}) and denoising real noise data (Figure~\ref{fig:realfirst} -- Figure~\ref{fig:reallast}).

%%%%%%%%%%%%% -- more supp images

\begin{figure*}[!hbtp]
    \centering
    \includegraphics[width=0.75\textwidth]{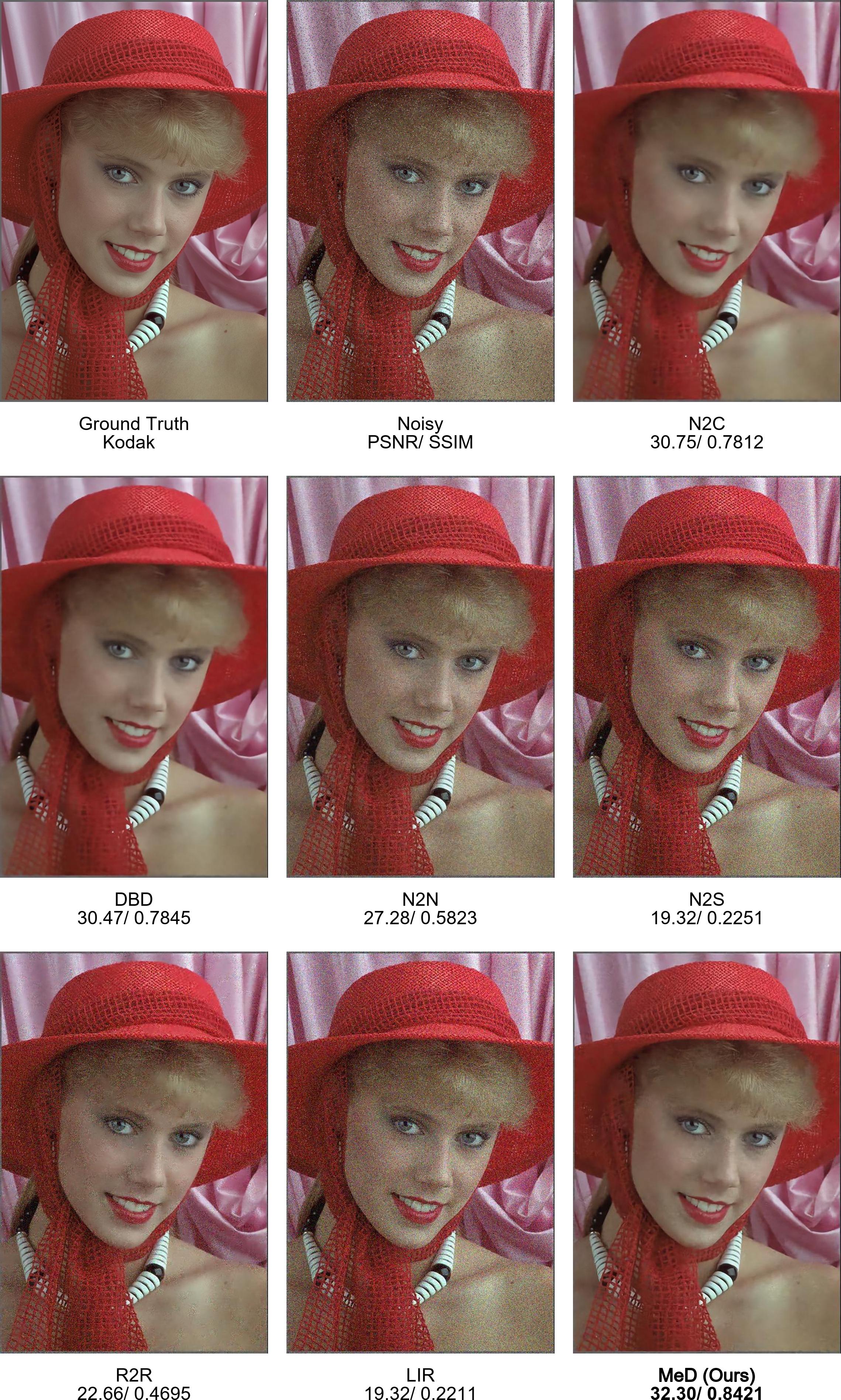}
	\caption{Visual comparison of image denoising methods on Kodak \cite{franzen1999kodak} images with Gaussian ($\sigma=25$) + local variance Gaussian noise.} 
\end{figure*}

\begin{figure*}[!hbtp]
    \centering
    \includegraphics[width=\textwidth]{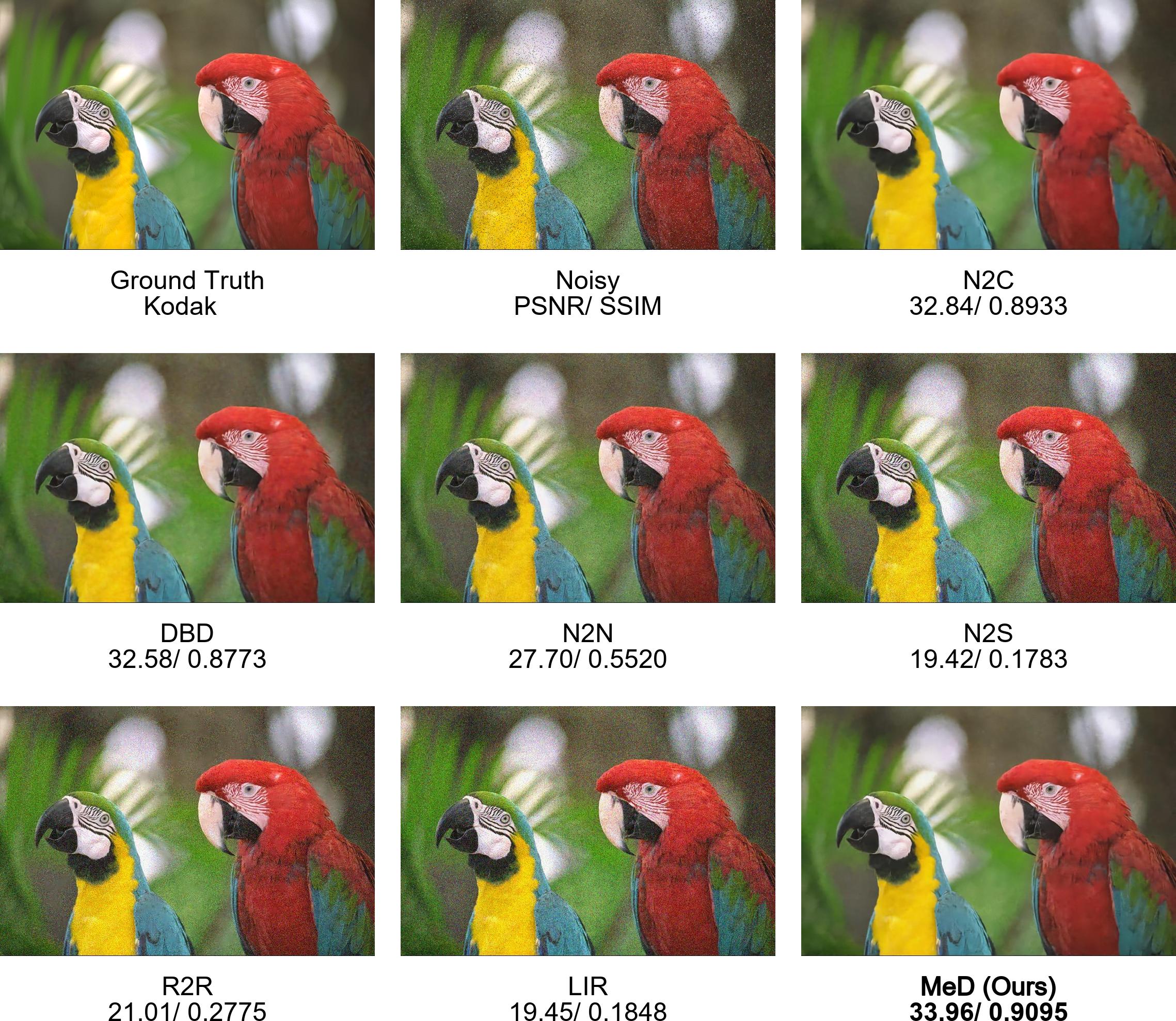}
	\caption{Visual comparison of image denoising methods on Kodak \cite{franzen1999kodak} images with Gaussian ($\sigma=25$) + local variance Gaussian noise.} 
\end{figure*}

\begin{figure*}[!hbtp]
    \centering
    \includegraphics[width=\textwidth]{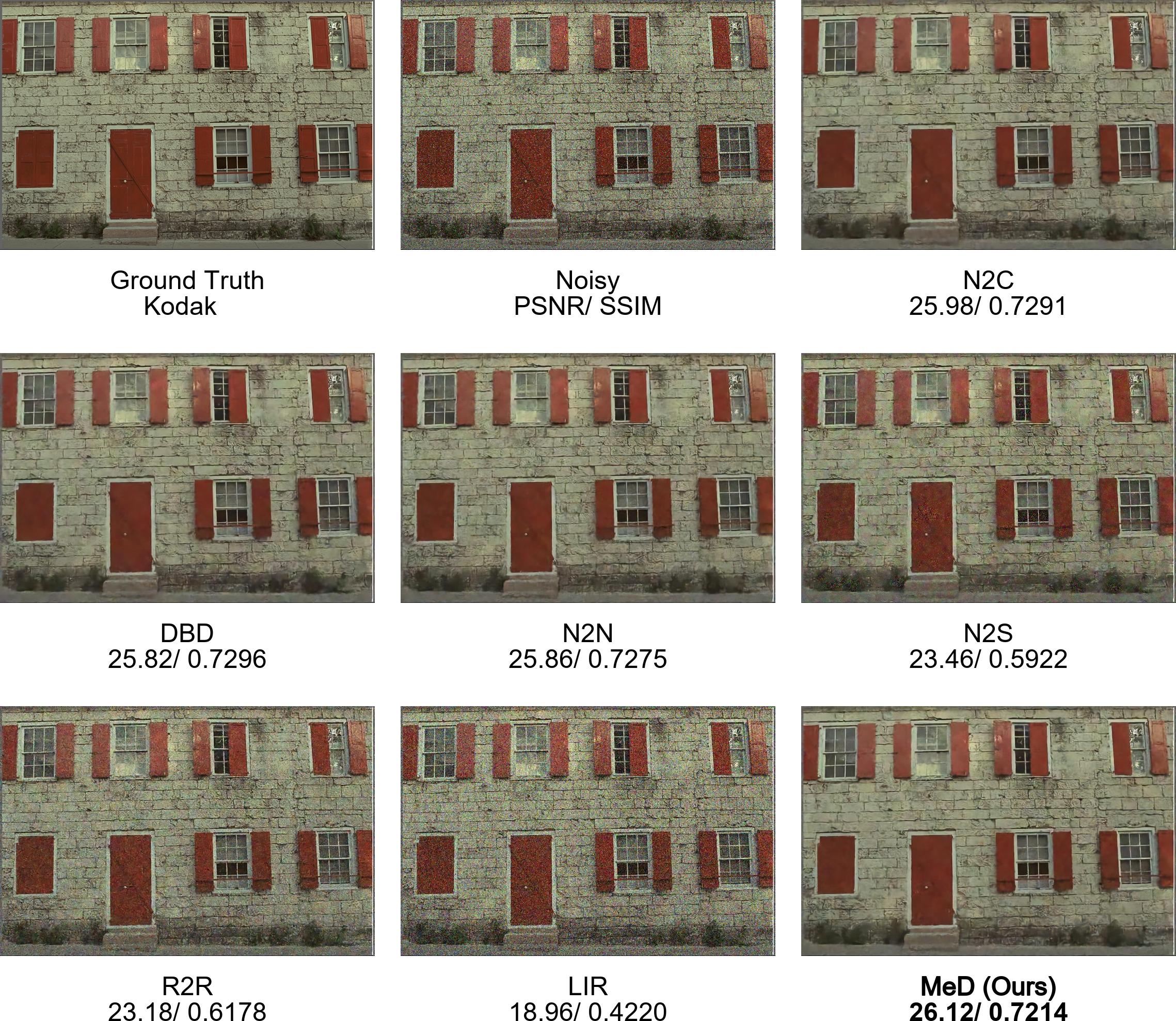}
	\caption{Visual comparison of image denoising methods on Kodak \cite{franzen1999kodak} images with Gaussian ($\sigma=50$) + local variance Gaussian noise.} 
\end{figure*}

\begin{figure*}[!hbtp]
    \centering
    \includegraphics[width=\textwidth]{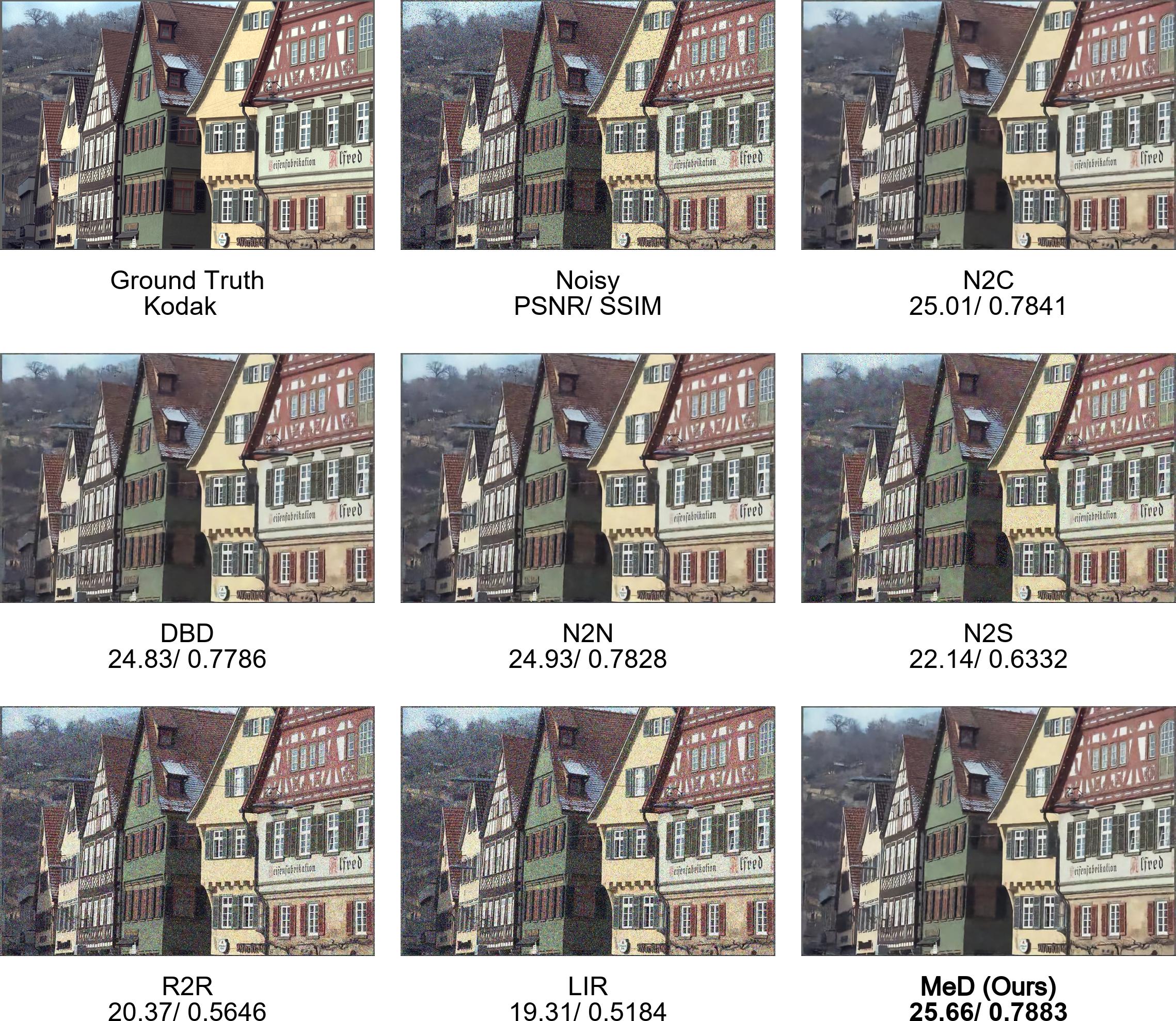}
	\caption{Visual comparison of image denoising methods on Kodak \cite{franzen1999kodak} images with Gaussian ($\sigma=50$) + local variance Gaussian noise.} 
\end{figure*}
\begin{figure*}[!hbtp]
    \centering
    \includegraphics[width=0.75\textwidth]{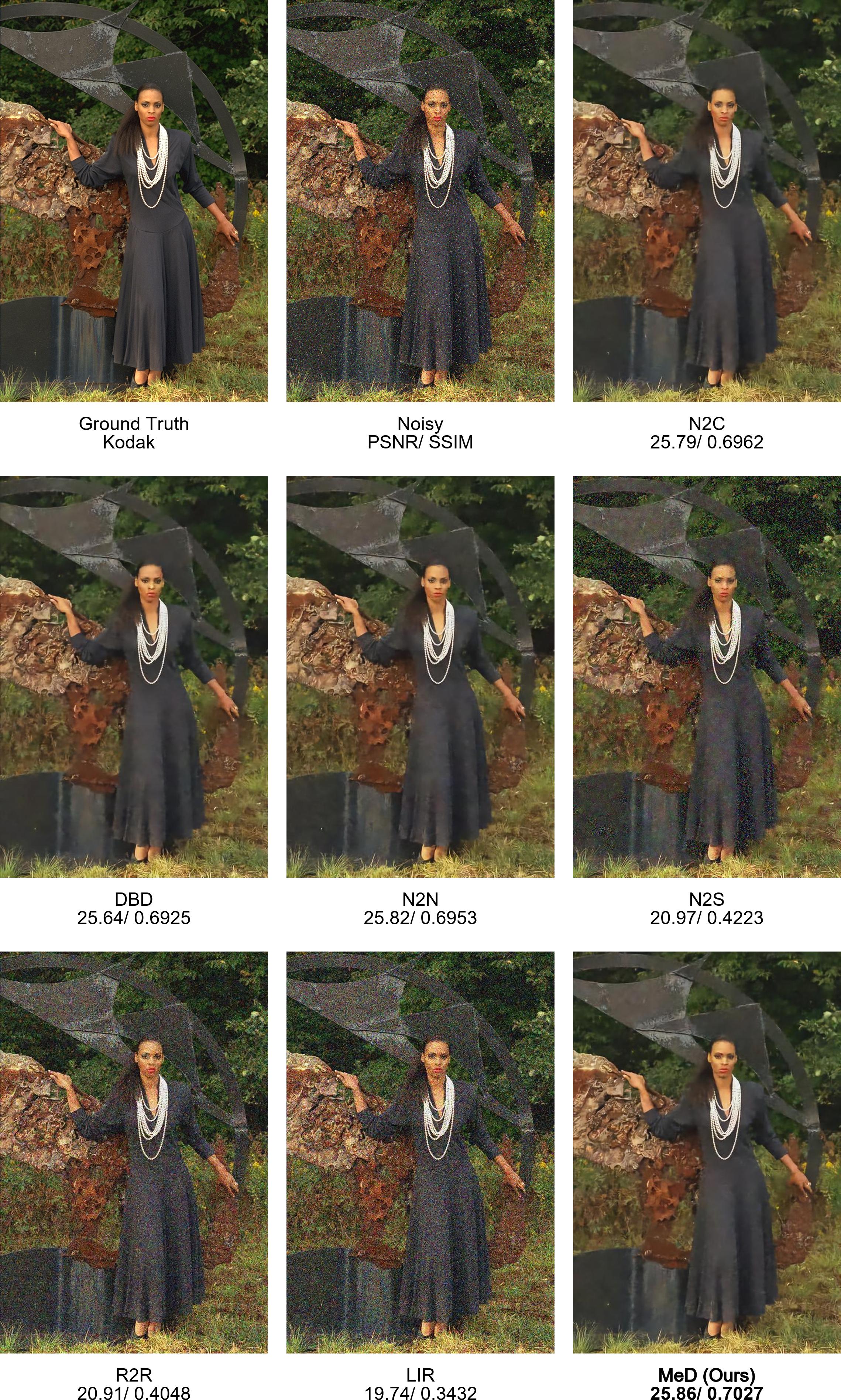}
	\caption{Visual comparison of image denoising methods on Kodak \cite{franzen1999kodak} images with Gaussian ($\sigma=50$) + local variance Gaussian noise.} 
\end{figure*}
\begin{figure*}[!hbtp]
    \centering
    \includegraphics[width=0.75\textwidth]{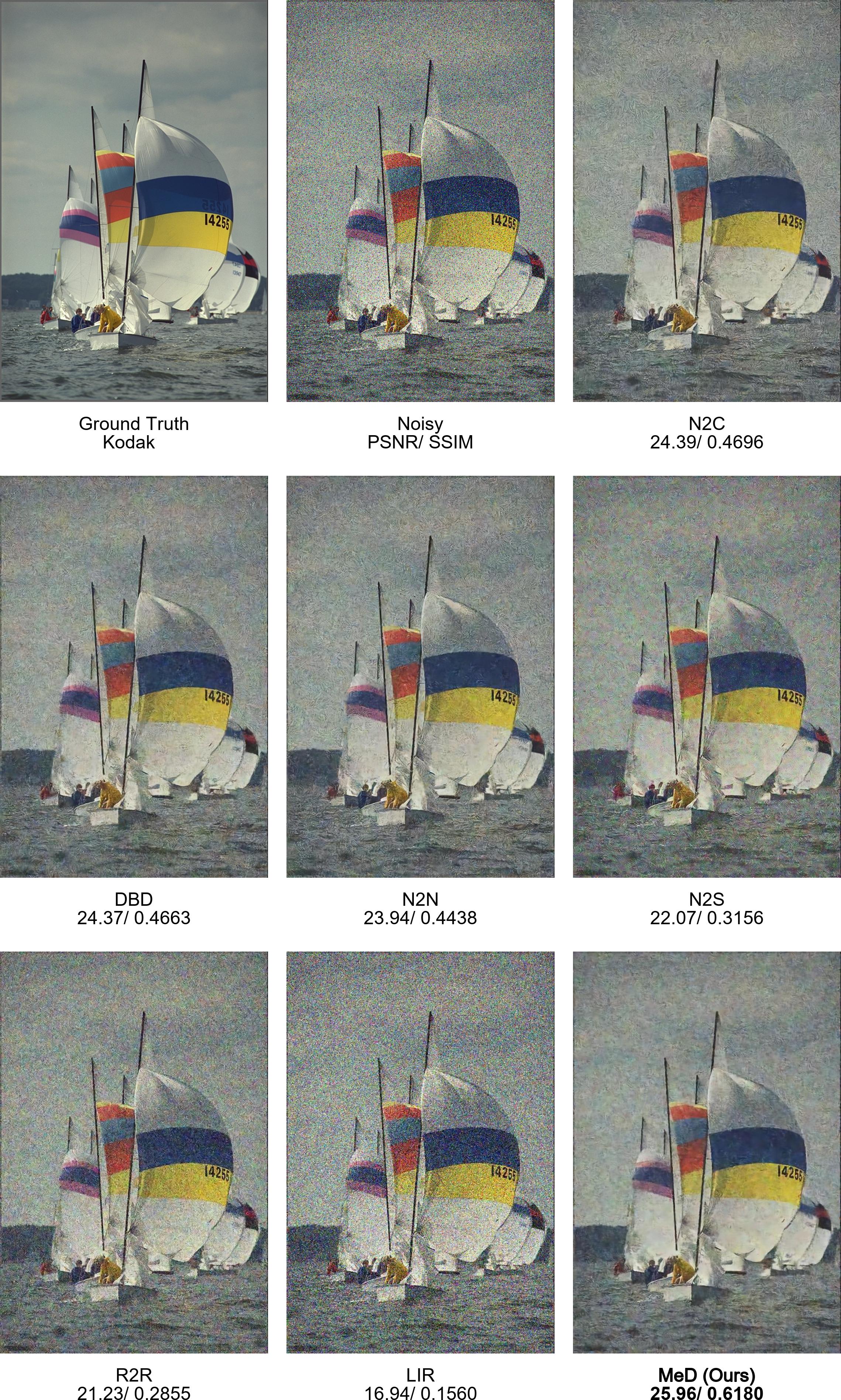}
	\caption{Visual comparison of image denoising methods on Kodak \cite{franzen1999kodak} images with Gaussian ($\sigma=75$) + local variance Gaussian noise.} 
\end{figure*}
\begin{figure*}[!hbtp]
    \centering
    \includegraphics[width=0.75\textwidth]{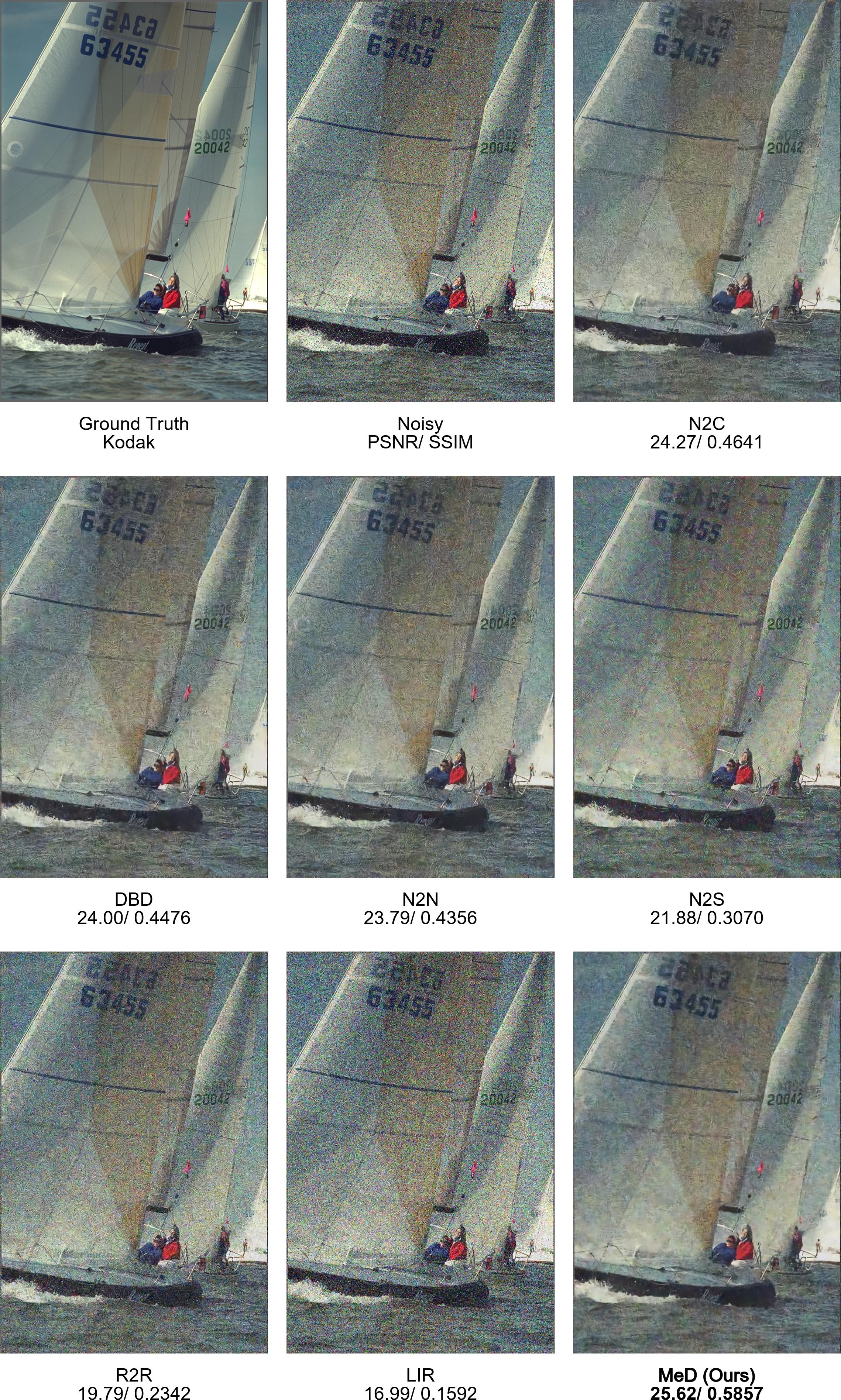}
	\caption{Visual comparison of image denoising methods on Kodak \cite{franzen1999kodak} images with Gaussian ($\sigma=75$) + local variance Gaussian noise.} 
\end{figure*}
\begin{figure*}[!hbtp]
    \centering
    \includegraphics[width=\textwidth]{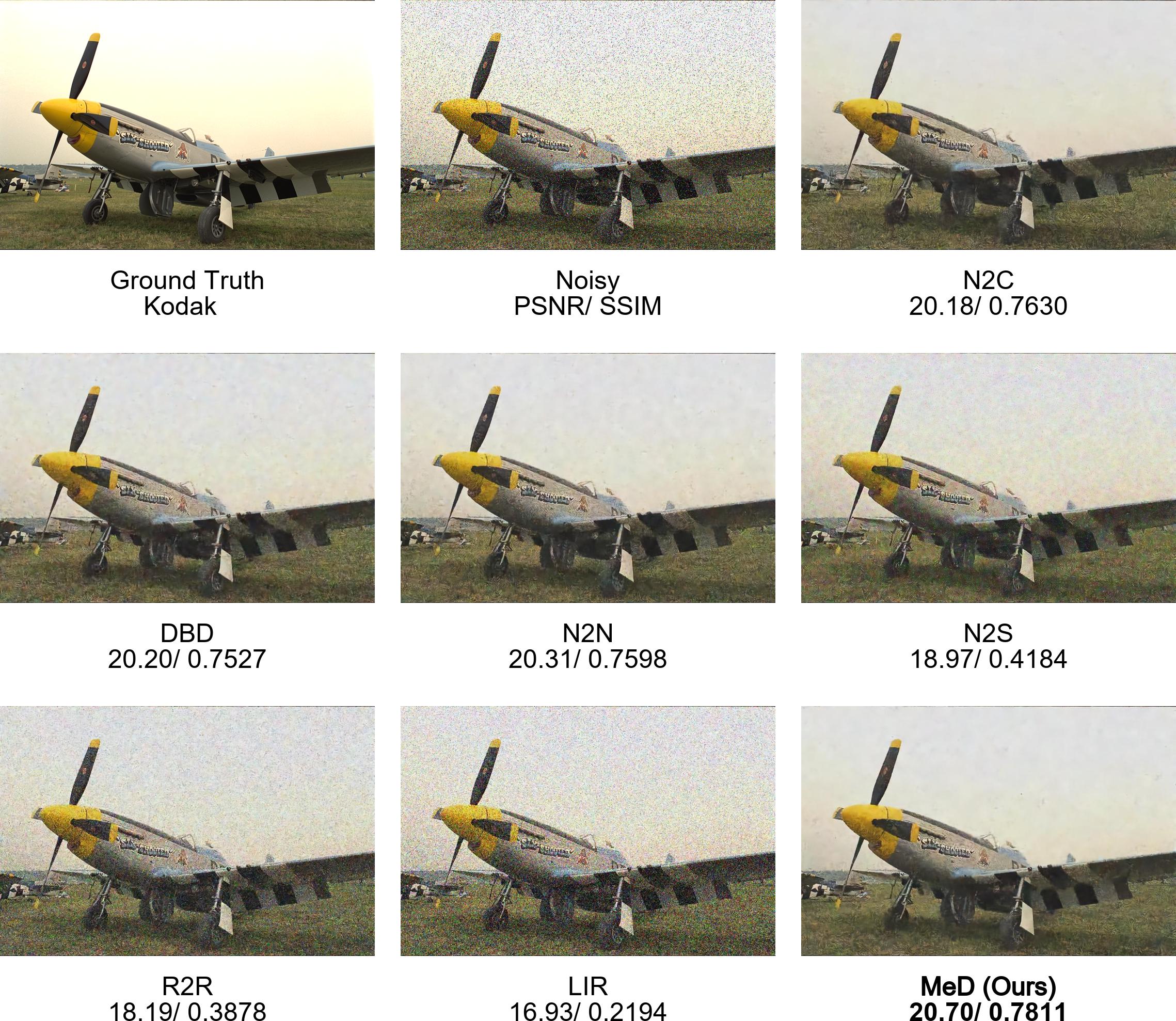}
	\caption{Visual comparison of image denoising methods on Kodak \cite{franzen1999kodak} images with Gaussian ($\sigma=75$) + local variance Gaussian noise.} 
\end{figure*}

% \begin{figure*}[!hbtp]
%     \centering  
%     %  左下右上  , trim=10 0 150 200,clip
%     \includegraphics[width=\textwidth ]{figures/real/real_50.jpg}
%     % \includegraphics[height=0.9\textheight ]{figures/real/real_50.jpg}
% 	\caption{Visual comparison of image denoising methods on real noisy image dataset SIDD \cite{SIDD_2018_CVPR} example images with real noise.} 
%  \label{fig:realfirst}
% \end{figure*}

% \begin{figure*}[!hbtp]
%     \centering
%     \includegraphics[width=\textwidth, trim=10 0 150 200,clip]{figures/real/real_85.jpg}
% \end{figure*}

\begin{figure*}[!hbtp]
    \centering
    \includegraphics[width=\textwidth]{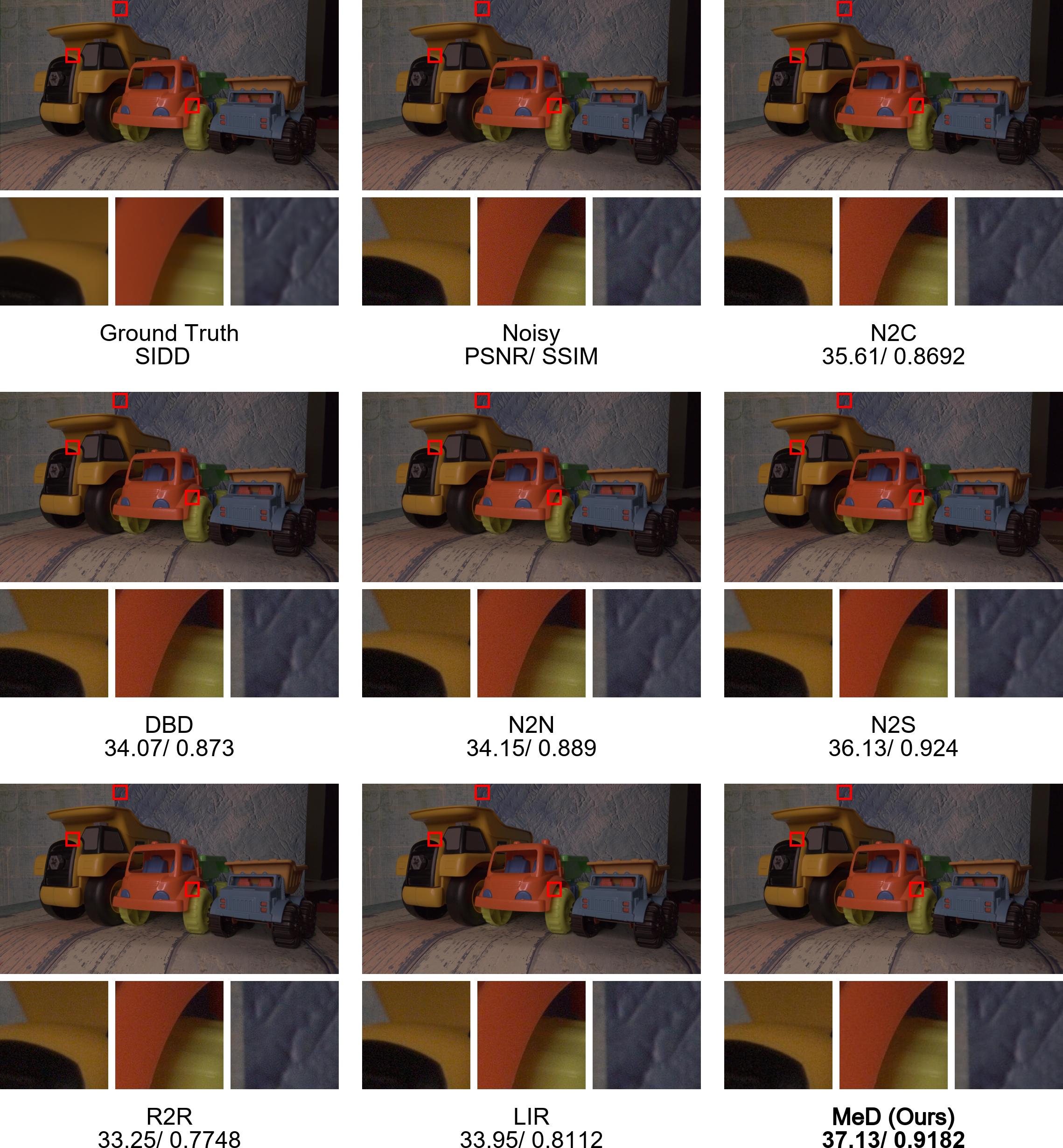}
	\caption{Visual comparison of image denoising methods on real noisy image dataset SIDD \cite{SIDD_2018_CVPR} example images with real noise.} 
\end{figure*}

\begin{figure*}[!hbtp]
    \centering
    \includegraphics[width=\textwidth]{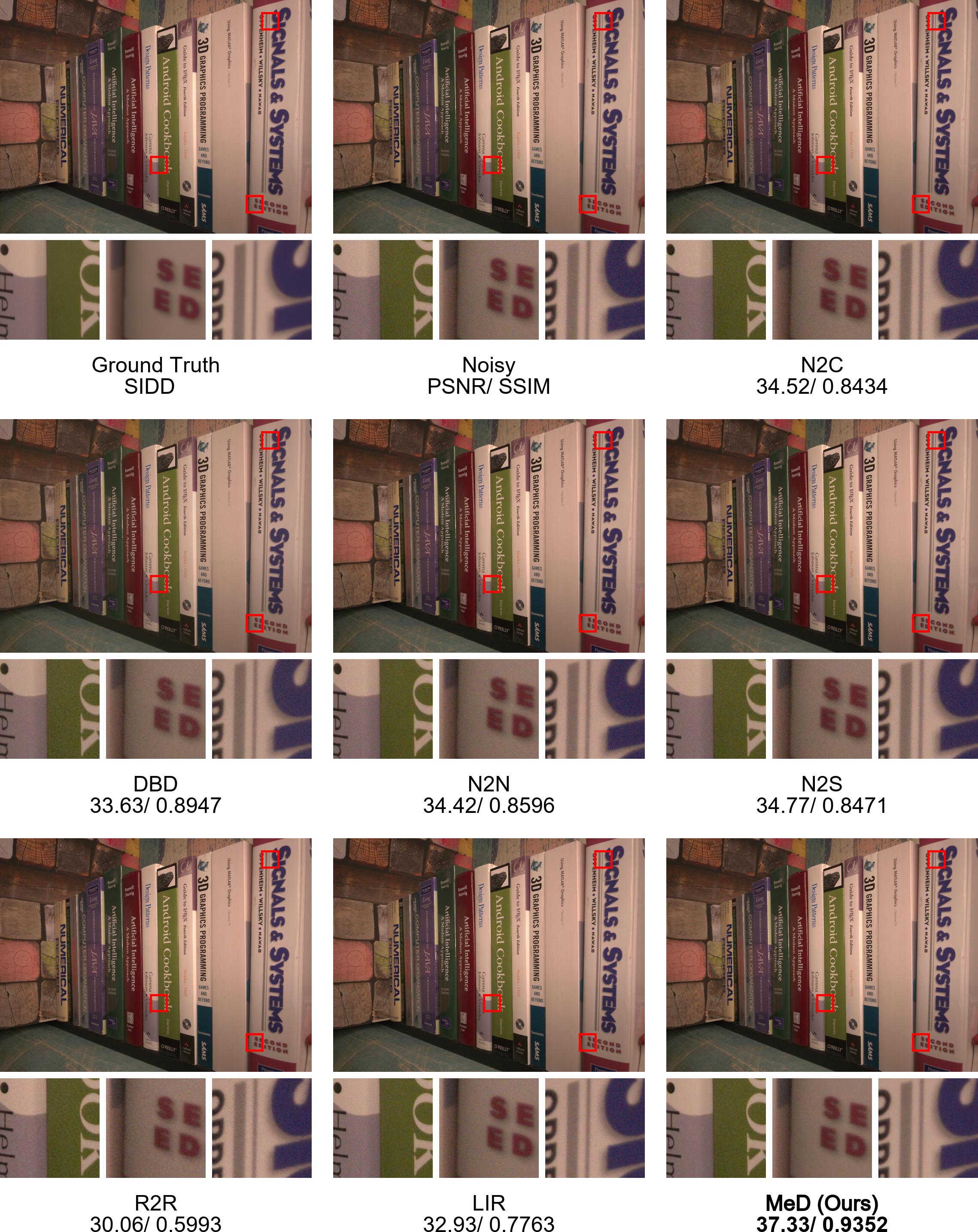}
	\caption{Visual comparison of image denoising methods on real noisy image dataset SIDD \cite{SIDD_2018_CVPR} example images with real noise.} 
\end{figure*}

\begin{figure*}[!hbtp]
    \centering
    \includegraphics[width=\textwidth]{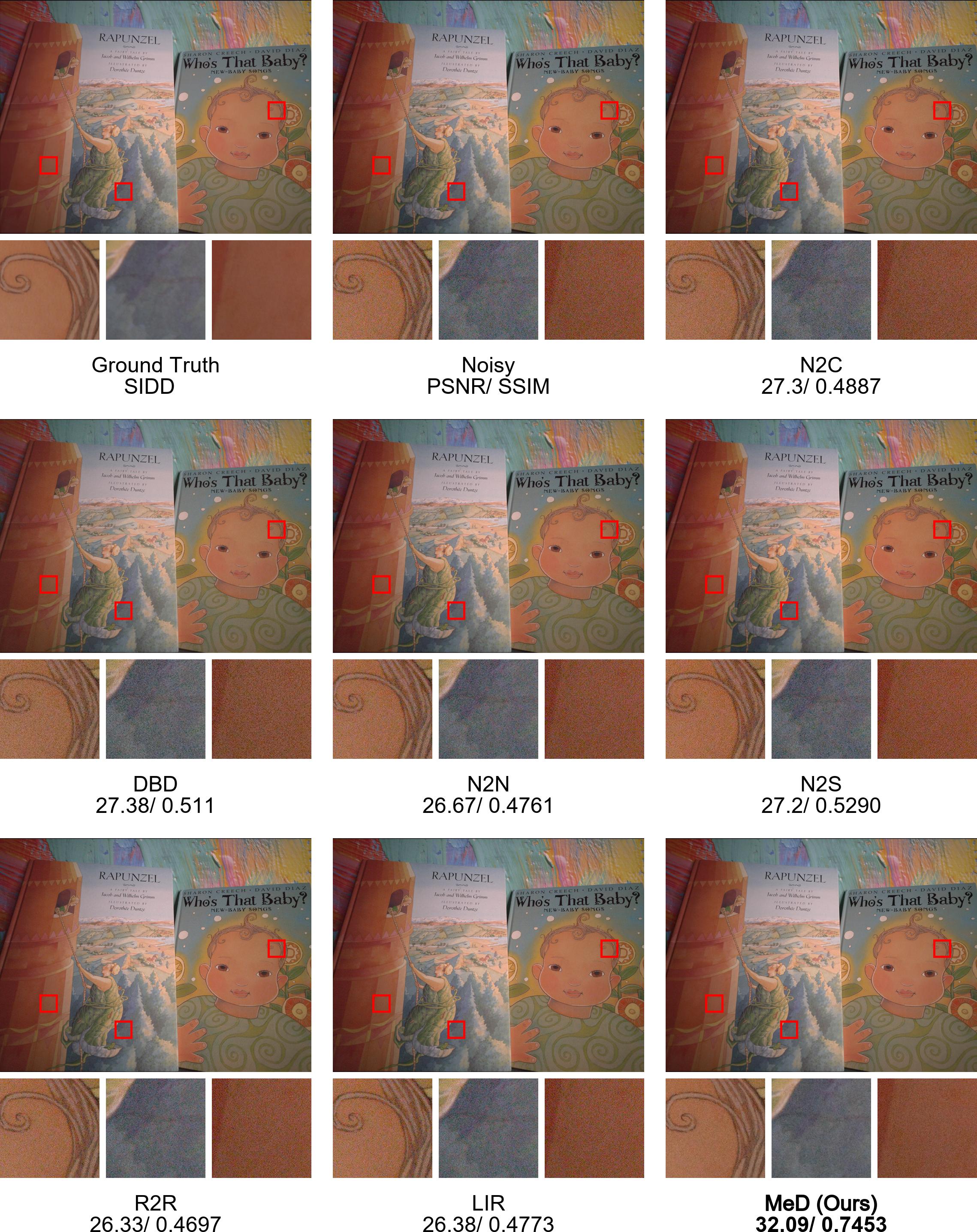}
	\caption{Visual comparison of image denoising methods on real noisy image dataset SIDD \cite{SIDD_2018_CVPR} example images with real noise.} 
 
\end{figure*}

\begin{figure*}[!hbtp]
    \centering
    \includegraphics[width=\textwidth]{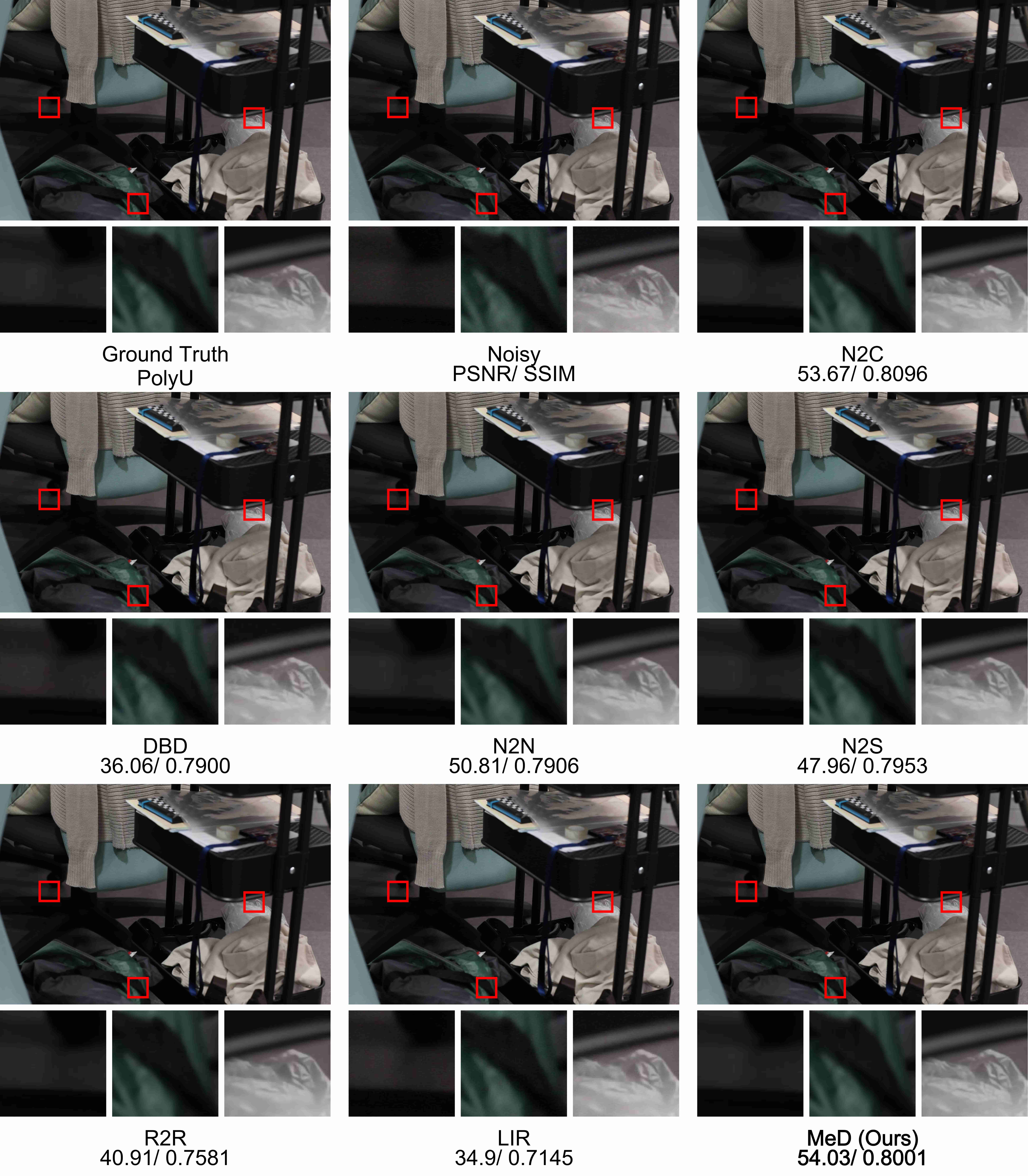}
	\caption{Visual comparison of image denoising methods on real noisy image dataset PolyU \cite{xu2018real} example images with real noise.}

\end{figure*} 

\begin{figure*}[!hbtp]
    \centering
    \includegraphics[width=\textwidth]{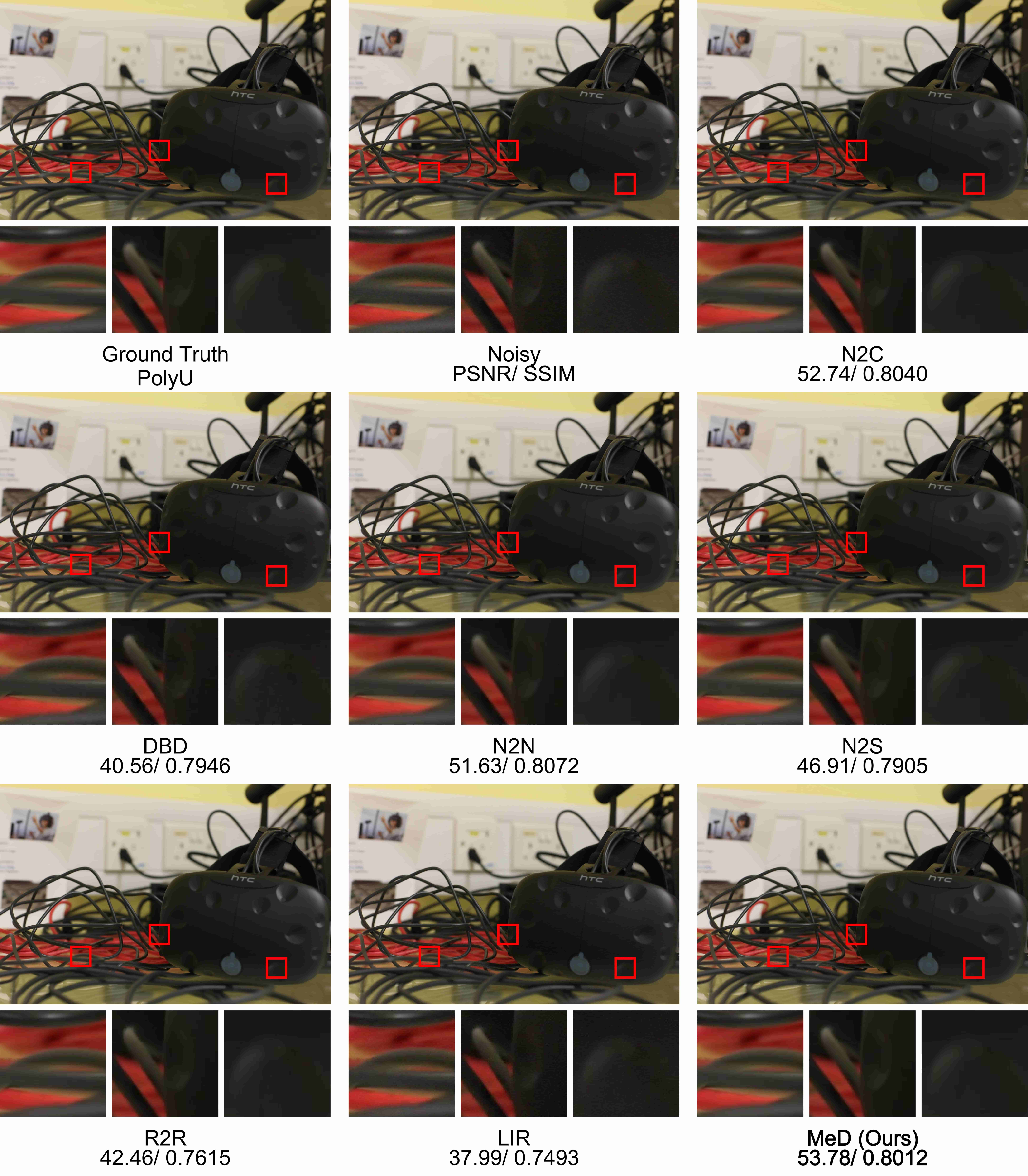}
	\caption{Visual comparison of image denoising methods on real noisy image dataset PolyU \cite{xu2018real} example images with real noise.}

\end{figure*} 

\begin{figure*}[!hbtp]
    \centering
    \includegraphics[width=\textwidth]{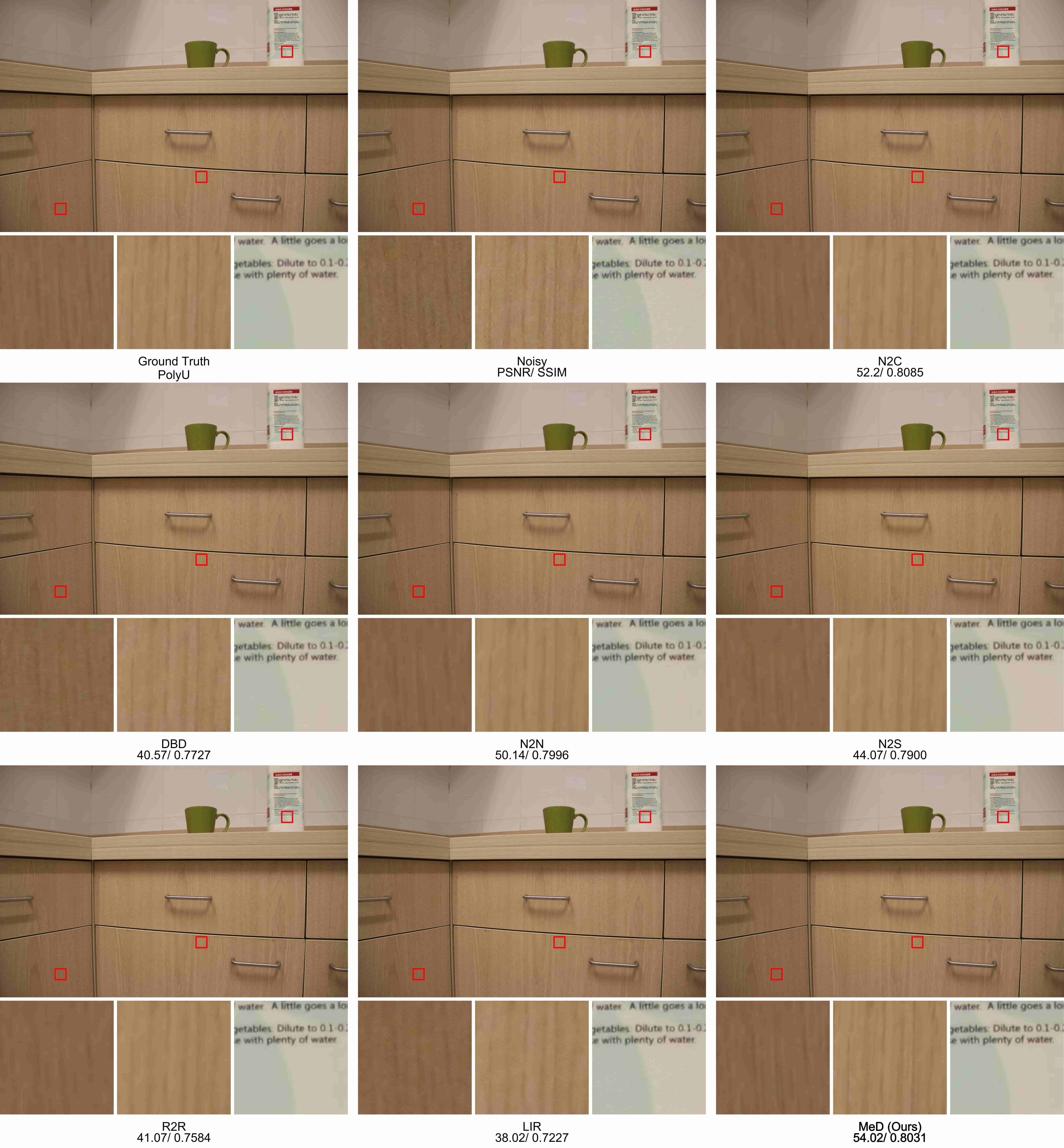}
	\caption{Visual comparison of image denoising methods on real noisy image dataset PolyU \cite{xu2018real} example images with real noise.}

\end{figure*} 

% \end{appendices}

% % {\small
% % 	\bibliographystyle{../ieee_fullname}
% % 	\bibliography{../main}
% % }

% \end{document}

\end{document}